\documentclass[nohyperref]{article}

\usepackage{microtype}
\usepackage{graphicx}
\usepackage{subfigure}
\usepackage{booktabs} %

\usepackage{hyperref}

\usepackage[accepted]{icml2022}

\usepackage{amsmath}
\usepackage{amssymb}
\usepackage{mathtools}
\usepackage{amsthm}

\usepackage[capitalize,noabbrev]{cleveref}

\theoremstyle{plain}
\newtheorem{theorem}{Theorem}[section]

\newtheorem{lemma}[theorem]{Lemma}

\theoremstyle{definition}

\newtheorem{assumption}[theorem]{Assumption}
\theoremstyle{remark}
\newtheorem{remark}[theorem]{Remark}

\usepackage[textsize=tiny]{todonotes}

\usepackage[utf8]{inputenc} %
\usepackage[T1]{fontenc}    %
\usepackage{titletoc}
\usepackage{hyperref}       %
\usepackage{url}            %
\usepackage{booktabs}       %
\usepackage{amsfonts}       %
\usepackage{nicefrac}       %
\usepackage{microtype}      %
\usepackage{xcolor}         %

\newcommand{\argmax}{\mathop{\rm argmax}\limits}
\newcommand{\argmin}{\mathop{\rm argmin}\limits}

\newcommand{\ep}{\hfill $\Box$}

\usepackage{bbm}
\usepackage{paralist}
\usepackage{enumitem}
\usepackage{xcolor}
\usepackage{mathtools}
\usepackage{url}

\usepackage{dsfont}

\renewcommand{\vec}{\boldsymbol}

\DeclareMathOperator{\EXP}{\mathbb{E}}

\renewcommand{\tilde}{\widetilde}

\renewcommand{\Pr}{\mathbb{P}}

\newcommand{\xpt}{\textnormal{xpt}}

\newcommand{\indicator}{\mathds{1}}

\usepackage{pdfrender}
\newcommand*{\boldcheckmark}{%
  \textpdfrender{
    TextRenderingMode=FillStroke,
    LineWidth=.5pt, %
  }{\checkmark}%
}

\renewcommand{\cite}{\citep}

\usepackage{tablefootnote}

\icmltitlerunning{Thresholded Lasso Bandit}

\begin{document}

\twocolumn[
\icmltitle{Thresholded Lasso Bandit}

\icmlsetsymbol{equal}{*}

\begin{icmlauthorlist}
\icmlauthor{Kaito Ariu}{kth,ca}
\icmlauthor{Kenshi Abe}{ca}
\icmlauthor{Alexandre Proutière}{kth}
\end{icmlauthorlist}

\icmlaffiliation{kth}{EECS and Digital Futures, KTH Royal Institute of Technology, Stockholm, Sweden}
\icmlaffiliation{ca}{Cyberagent, Inc., Tokyo, Japan}

\icmlcorrespondingauthor{Kaito Ariu}{ariu@kth.se}

\icmlkeywords{Multi-armed Bandits, Regret Minimization, Linear Bandits, Lasso}

\vskip 0.3in
]

\printAffiliationsAndNotice{}  %

\begin{abstract}
 In this paper, we revisit the regret minimization problem in sparse stochastic contextual linear bandits, where feature vectors may be of large dimension $d$, but where the reward function depends on a few, say $s_0\ll d$, of these features only. We present Thresholded Lasso bandit, an algorithm that (i) estimates the vector defining the reward function as well as its sparse support, i.e., significant feature elements, using the Lasso framework with thresholding, and (ii) selects an arm greedily according to this estimate projected on its support. The algorithm does not require prior knowledge of the sparsity index $s_0$ and can be parameter-free under some symmetric assumptions. For this simple algorithm, we establish non-asymptotic regret upper bounds scaling as $\mathcal{O}( \log d + \sqrt{T} )$ in general, and as $\mathcal{O}( \log d + \log T)$ under the so-called margin condition (a probabilistic condition on the separation of the arm rewards). The regret of previous algorithms scales as $\mathcal{O}( \log d + \sqrt{T \log (d T)})$ and $\mathcal{O}( \log T \log d)$ in the two settings, respectively. Through numerical experiments, we confirm that our algorithm outperforms existing methods. 
\end{abstract}

\section{Introduction}

The linear contextual bandit \cite{abe1999, li2010contextual} is a sequential decision-making problem that generalizes the classical stochastic Multi-Armed Bandit (MAB) problem \cite{lai1985asymptotically,robbins1952some}, where (i) in each round, the decision-maker is provided with a {\it context} in the form of a feature vector for each arm and where (ii) the expected reward is a linear function of these vectors. More precisely, at the beginning of round $t\ge 1$, the decision-maker receives for each arm $k$, a feature vector $A_{t, k}\in \mathbb{R}^d$. She then selects an arm, say $k$, and observes a sample of a random reward with mean $\langle A_{t,k}, \theta\rangle$. The parameter vector $\theta\in \mathbb{R}^d$ is fixed but initially unknown. Linear contextual bandits have been extensively applied in online services such as online advertisement and recommendation systems \cite{li2010contextual, li2016collaborative, zeng2016online}, and constitute arguably the most relevant structured bandit model in practice.

The major challenge in the design of efficient algorithms for contextual linear bandits stems from the high dimensionality of the feature space. For example, for display ad systems as studied in \citet{chapelle2011empirical,weinberger2009}, the joint information about a user, an ad and its publisher is encoded in a feature vector of dimension $d=2^{24}$. Fortunately, typically only a few features significantly impact the expected reward. This observation has motivated the analysis of problems where the unknown parameter vector $\theta$ is sparse \cite{Bastani2015,kim2019doubly, oh2020sparsity,wang2018minimax}. In this paper, we also investigate  sparse contextual linear bandits, and assume that $\theta$ only has at most $s_0\ll d$ non-zero components. The set of these components and its cardinality $s_0$ are unknown to the decision-maker. Sparse contextual linear bandits have attracted a lot of attention recently. State-of-the-art algorithms developed to exploit the sparse structure achieve regrets scaling as $\mathcal{O}( \log d + \sqrt{T \log (dT)})$ in general, and $ \mathcal{O}( \log d \log T)$ under the co-called margin condition (a setting where arms are well separated); refer to Section \ref{sec:related} for details.              

We develop a novel algorithm, referred to as Thresholded Lasso bandit\footnote{An implementation of our method is available at \url{https://github.com/CyberAgentAILab/thresholded-lasso-bandit}.}, with improved regret guarantees. Our algorithm first uses the Lasso framework with thresholding to maintain and update in each round estimates of the parameter vector $\theta$ and of its support. It then greedily picks an arm based on these estimates (the {\it thresholded} estimates of $\theta$). The regret of the algorithm strongly depends on the accuracy of these estimates. We derive strong guarantees on this accuracy, which in turn leads to regret guarantees. Our contributions are as follows.

\paragraph{(i) Thresholded Lasso Estimation Performance.} The performance of the Lasso-based estimation procedure is now fairly well understood, see e.g., \citet{buhlmann2011statistics,tibshirani1996regression, zhou2010thresholded}. For example, \citet{zhou2010thresholded} provides an analysis of the estimation of the support of $\theta$, and specifically, gives upper bounds on the number of false positives (components that are not in the support, but estimated as part of it) and false negatives (components that are in the support, but not estimated as part of it). These analyses, however, critically rely on the assumption that the observed data is i.i.d.. This assumption does not hold for the bandit problem, as the algorithm adapts its arm selection strategy depending on the past observations. Despite the non i.i.d. nature of the data, we manage to derive performance guarantees of the estimate of $\theta$. In particular, we establish high probability guarantees that are independent of the dimension $d$ (see Lemma~\ref{lm:LS_estim_risk_klarge}).

\paragraph{(ii) Regret Guarantees.} Based on the analysis of the Thresholded Lasso estimation procedure, we provide a finite-time analysis of the regret of our algorithm under certain symmetry assumptions made in \citet{oh2020sparsity}. 
The regret scales at most as ${\mathcal{O}}( \log d + \sqrt{T} )$ in general and ${\mathcal{O}}(\log d + \log T)$ under the margin condition. More precisely, the estimation error of $\theta$ induces a regret scaling as $\mathcal{O}( \sqrt{ T})$ (or $ \mathcal{O}( \log T)$ under the margin condition). The additional term $ \mathcal{O}(\log d )$ in our regret upper bounds comes from the errors made when estimating the support of $\theta$. It is worth noting that when using the plain Lasso estimator (without thresholding), one would obtain weaker performance guarantees for the estimation of $\theta$, typically depending on $d$, see e.g., \citet{Bastani2015}. This dependence causes an additional multiplicative term $\log d$ in the regret.

\paragraph{(iii) Numerical Experiments.} We present extensive numerical experiments to illustrate the performance of the Thresholded Lasso bandit algorithm. We compare the estimation accuracy for $\theta$ and the regret of our algorithm to those of the Lasso bandit, Doubly-Robust Lasso bandit, and Sparsity-Agnostic Lasso bandit algorithms \cite{bastani2020online, kim2019doubly, oh2020sparsity}. These experiments confirm the benefit of the use of the Lasso procedure with thresholding.

\begin{table*}[t]
\caption{Algorithms and their regret guarantees for scaling with respect to $d$ and $T$. $\mathcal{O}$ notations are hiding sub-logarithmic factors in $d$ and logarithmic factors in $s_0$. The 'Compatibility' and 'Margin' conditions refer to Assumptions \ref{asm:comp_cond} and \ref{asm:margin}.}
\label{tab:related_work}
\vskip 0.15in 
\begin{center}
\begin{tabular}{lcccc}
\toprule
Paper & Regret & 
Compatibility & Margin & Other \\
\midrule
{\small \citet{abbasi2012online}} & ${\mathcal{O}}(d s_0 (\log { T})^2)$ & & & \\
{\small \citet{bastani2020online}}  & $\mathcal{O}(s_0^2(\log d + \log T)^2)$ & \boldcheckmark  & \boldcheckmark &   \\
{\small \citet{wang2018minimax}}   & $\mathcal{O}(s_0^2( \log d + s_0) \log T)$ & \boldcheckmark  & \boldcheckmark  & \\
{\small \citet{kim2019doubly}} & $\mathcal{O}(s_0 \sqrt{T} \log (d T))$ & \boldcheckmark   &  & \\
{\small \citet{ren2020dynamic}} & {\small $\mathcal{O}(s_0 \mbox{polylog}(d) + \log(T)\sqrt{s_0T \log d })$} &    &  & {\small Restricted Bounded Density}
\\
{\small  \citet{oh2020sparsity}}   & $\mathcal{O} ( s_0^2 \log d + s_0 \sqrt{T \log (dT)})$& \boldcheckmark  &  & Asm \ref{asm:relax_sym}, \ref{asm:balanced_cov} \\
{\small This work: Theorem~\ref{thm:regret_ub_klarge}}&  $\mathcal{O}(s_0^2\log d  +  s_0\log T)$ & \boldcheckmark & \boldcheckmark & Asm \ref{asm:relax_sym}, \ref{asm:balanced_cov}, \ref{asm:cov_div_klarge}\\
{\small This work: Theorem~\ref{thm:regret_ub_wo_margin_klarge}}& $\mathcal{O}(s_0^2 \log d  +  
\sqrt{s_0 T })$& \boldcheckmark  &  & 
Asm \ref{asm:relax_sym}, \ref{asm:balanced_cov}, \ref{asm:cov_div_klarge} 
\\
\begin{tabular}{c}
{\small \citet{bastani2021mostly}}\\
{\small (non-sparse setting) }
\end{tabular}
& $\mathcal{O}(d \log d \log T)$ & & \boldcheckmark  & 
\begin{tabular}{c}
{\small Covariate diversity condition}\\
{\small (Stronger than Asm~\ref{asm:cov_div_klarge})}
\end{tabular}
\\
\bottomrule
\end{tabular}
\end{center}
\vskip -0.1in
\end{table*}

\section{Related Work}\label{sec:related}
 
Stochastic linear bandit problems have attracted a lot of attention over the last decade. \citet{carpentier2012bandit} addresses sparse linear bandits where  $\|\theta\|_0 \le s_0$ and where the set of arms is restricted to the $\ell_2$ unit ball. For regimes where the time horizon is much smaller than the dimension $d$, i.e., $T \ll d$, the authors propose an algorithm whose regret scales at most as ${\mathcal{O}}(s_0\sqrt{T} \log(d T))$. \citet{botao2020high} studied high-dimensional linear bandit problems where the number of actions is larger than or equal to $d$. Under some signal strength conditions, they propose an algorithm that achieves a regret of $\mathcal{O}(s_0 \log d + \sqrt{s_0 T \log (dT)})$. These studies, however, do not consider problems with contextual information.
    
Recently, high-dimensional contextual linear bandits have been investigated under the sparsity assumption $\|\theta\|_0 \le s_0$. In this line of research, the decision-maker is provided in each round with a set of arms defined by a finite set of feature vectors. This set is uniformly bounded across rounds. In this setting, the authors of \citet{abbasi2012online} devise an algorithm with both a minimax (problem independent) regret upper bound $\tilde{\mathcal{O}}(\sqrt{s_0 dT})$ and problem dependent upper bound ${\mathcal{O}}(d s_0 (\log { T})^2)$ (the notation $\tilde{\mathcal{O}}$ hides the polylogarithmic terms) without any assumption on the distribution (other than the assumptions similar to our Assumption~\ref{asm:sparsity_param_klarge}).

In \citet{bastani2020online} (initially published in 2015 as \citet{Bastani2015}), the authors address a high-dimensional contextual linear bandit problem where the unknown parameter defining the reward function is arm-specific ($\theta$ is different for the various arms). In the proposed algorithm, arms are explored uniformly at random for $\mathcal{O}( s_0^2 \log d \log T)$ prespecified rounds. Under the margin condition, similar to our Assumption~\ref{asm:margin}, the algorithm achieves a regret of $\mathcal{O}(s_0^2(\log d + \log T)^2)$. For the same problem, \citet{wang2018minimax} develops the so-called MCP-Bandit algorithm. The latter also uses the uniform exploration for $\mathcal{O}( s_0^2 \log d \log T)$ prespecified rounds, and has improved regret guarantees: the regret scales as $\mathcal{O}(s_0^2( \log d + s_0) \log T)$. 
  
High-dimensional contextual linear bandits have been also studied without the margin condition, but with a unique parameter $\theta$ defining the reward function. \citet{kim2019doubly} designs an algorithm with uniform exploration phases of $\mathcal{O}(\sqrt{T  \log(d T) \log T})$ rounds, and with regret $\mathcal{O}(s_0 \sqrt{T} \log (d T))$. All the aforementioned algorithms require the knowledge of the sparsity index $s_0$. In \citet{oh2020sparsity}, the authors propose an algorithm, referred to as SA Lasso bandit, that does not require this knowledge, and with regret $\mathcal{O} (s_0^2 \log d +  s_0 \sqrt{T \log (dT)})$. 
In addition, SA Lasso bandit does not include any uniform exploration phase. Its regret guarantees are derived under specific assumptions on the context distribution, the so-called relaxed symmetry assumption and the balanced covariance assumption. 
The authors also establish a $\mathcal{O}(s_0^2 \log d + \sqrt{s_0 T \log (d T)})$ regret upper bound under the so-called {\it restricted eigenvalue} condition. In \citet{ren2020dynamic}, under the restricted eigenvalue condition induced by the restricted bounded density assumption, the authors established a regret bound of $\mathcal{O}(s_0 \mbox{polylog}(d) + \sqrt{s_0T \log(d) \mbox{polylog}(T)})$.

One notable recent development in contextual bandit is the regret analyses of the exploration-free or greedy algorithms. Under some symmetry assumptions on the context distribution, these algorithms exhibit sub-linear regret  \cite{bastani2021mostly, kannan2018smoothed, ren2020dynamic, oh2020sparsity}.  Our analysis is also inspired by the results on exploration-free algorithms.

In this paper, we develop an algorithm with improved regret guarantees with and without the margin condition. The algorithm does not rely on the knowledge of $s_0$ and can be made parameter-free.  Such a parameter-free algorithm is not proposed in other recent papers. We have summarized the relevant studies and our work in Table~\ref{tab:related_work}. We will come back to this table when we discuss the assumptions.

\section{Model and Assumptions}\label{sec:model}

\subsection{Model and Notation}
 We consider a contextual linear stochastic bandit problem in a high-dimensional space. In each round $t \in [T]:=\{1,\ldots,T\}$, the algorithm is given a set of context vectors $ \mathcal{A}_t = \{ A_{t, k} \in \mathbb{R}^{d}: k \in [K]\}$. The successive sets $( \mathcal{A}_t )_{t \ge 1}$ form an i.i.d. sequence with distribution $p_{{A}}$. In round $t$, the algorithm selects an arm $A_t \in \mathcal{A}_t $ based on past observations, and collects a random reward $r_t$. Formally, if $\mathcal{F}_{t}$ is the $\sigma$-algebra generated by random variables $(\mathcal{A}_1,A_1,r_1, \ldots,\mathcal{A}_{t-1},A_{t-1},r_{t-1}, \mathcal{A}_{t})$, $A_t$ is $\mathcal{F}_{t}$-measurable. We assume that $r_t = \langle A_{t}, \theta\rangle + \varepsilon_t$, where $\varepsilon_t$ is a  zero mean sub-Gaussian random variable with variance proxy $\sigma^2$ given $\mathcal{F}_{t}$ and $A_t$.
Our objective is to devise an algorithm with minimal regret, where regret is defined as: 
 \begin{align*}
 	R(T) & \coloneqq \EXP \left[ \sum_{t =1}^T \max_{A \in \mathcal{A}_t} \langle A, \theta\rangle - r_t\right]
 	\\
 	& =  \EXP \left[ \sum_{t =1}^T \max_{A \in \mathcal{A}_t} \langle A - A_t, \theta\rangle\right].
 \end{align*}

\paragraph{Notation.} The $\ell_0$-norm of a vector $\theta\in \mathbb{R}^d$ is $\|\theta\|_0= \sum_{i = 1}^d  \indicator \left\{ \theta_i \neq 0\right\}$. We denote $\hat{\Sigma}_t = \frac{1}{t}\sum_{s=1}^t A_s A_s^\top$ as the empirical Gram matrix generated by the arms selected under a specific algorithm. For any $B \subset [d]$, we define $\theta_{B} \coloneqq (\theta_{1, B}, \ldots, \theta_{d, B})^\top $ where for all $i \in [d]$, $\theta_{i, B} \coloneqq \theta_i \indicator \{ i \in B\}$.
For each $ B \subset [d]$, we define the submatrix $A(B) \in \mathbb{R}^{n \times |B|}$ of $ A \in \mathbb{R}^{n \times d}$ where for $A(B)$, we extract the rows that are in $B$. We denote $\textnormal{supp}(x)$ as the set of the non-zero element indices of $x\in\mathbb{R}^d$. We also define $\theta_{\min}$ as the minimal value of $|\theta_i|$ on the support: $\theta_{\min} \coloneqq \min_{i \in \text{supp}(\theta)} |\theta_i|$. We denote by $S(\theta)= \text{supp}(\theta) = \{i \in [d] : \theta_i \neq 0\}$ the support of $\theta$. Definitions and notations are also summarized in Appendix~\ref{app:table_of_notations}, Table~\ref{tab:notions}.

\subsection{Assumptions} 

We present a set of assumptions used throughout the paper. 
Many assumptions are essentially from \citet{oh2020sparsity}. However, there are some differences, and these will be discussed.
We also discuss and compare these assumptions to those made in the related literature.  

\begin{assumption}[Sparsity and parameter constraints]\label{asm:sparsity_param_klarge}
The parameter $\theta$ defining the reward function is sparse, i.e., $\| \theta\|_0\le s_0$ for some fixed but unknown integer $s_0$ ($s_0$ does not depend on $d$). 
We further assume that $\|\theta\|_1 \le s_1 $ for some unknown constant\footnote{Given this assumption of $s_1$ to be a constant, note that the value of $\theta_{\min}$ would scale as $\Theta\left(1/s_0\right)$, as $s_0 \theta_{\min} \le \|\theta\|_1 \le s_1 $ may hold. Note that in  \citet{oh2020sparsity}, $\|\theta\|_2 \le \Theta(1)$ and $\|A\|_2 \le \Theta(1)$ (for all $A \in \mathcal{A}_t$) are assumed.} $s_1$  and $\theta_{\min} \ge s_2 /s_0 $ with some unknown constant $s_2\; (< s_1)$.
Finally, we assume that the $\ell_\infty$-norm of the context vector is bounded: for all $t$ and for all $A \in \mathcal{A}_t$, $\|A\|_\infty \le s_A$, where $s_A>0$ is a constant. 
\end{assumption}

\begin{assumption}[Compatibility condition]\label{asm:comp_cond}
  For a matrix $M \in \mathbb{R}^{d \times d}$ and a set $S_0 \subset [d]$, we define the compatibility constant $\phi(M, S_0)$ as:
$$
     \phi^2(M, S_0) \!\coloneqq \!\min_{x:  \|x_{S_0}\|_1 \neq 0} \left\{\! \frac{s_0 x^\top M x}{\|x_{S_0}\|_1^2} \!:\! \|x_{S_0^c}\|_1 \!\le\! 3 \|x_{S_0}\|_1 \!\right\}\!.
$$
We assume that for the Gram matrix of the action set $\Sigma \coloneqq \frac{1}{K} \sum_{k=1}^K \EXP_{\mathcal{A}\sim p_A} \left[A_{k} A_{k}^\top\right]$ satisfies $\phi^2(\Sigma, S(\theta)) \ge \phi^2_0$, where $\phi_0>0$ is some positive constant. 
 \end{assumption}
 The compatibility condition was introduced in the high-dimensional statistics literature \cite{buhlmann2011statistics}. It ensures that the Lasso estimate \cite{tibshirani1996regression} of the parameter $\theta$ approaches to its true value as the number of samples grows large. Note that it is easy to check that the compatibility condition is strictly weaker than assuming the positive definiteness of $\Sigma$. It allows us to consider feature vectors with strongly correlated components.  Assumption~\ref{asm:comp_cond} is considered to be essential for the Lasso estimate to be consistent and assumed in many of the relevant studies. See Table~\ref{tab:related_work} for studies using the compatibility condition. $\phi_0$ can be a constant that does not depend on $d$. This is the case, for example, when the context distribution is multivariate Gaussian, uniform distribution. 
 In these examples, the minimum eigenvalue of the Gram matrix is lower bounded by some constant. When the minimum eigenvalue is lower bounded by some constant, the compatibility constant is also lower bounded by some constant \cite{buhlmann2011statistics}.
 
 \begin{assumption}[Relaxed symmetry \cite{oh2020sparsity}]\label{asm:relax_sym}
 For the distribution $p_A$ of $\mathcal{A}$, there exists a constant $\nu\ge1$ such that for all $\vec{A} \in \mathbb{R}^{K \times d}$ such that $p_A(\vec{A})>0$, $\frac{p_A(\vec{A})}{p_A(- \vec{A})} \le \nu$.
 \end{assumption}
 
\begin{assumption}[Balanced covariance \cite{oh2020sparsity}]\label{asm:balanced_cov}
For any permutation $\gamma$ of $[K]$,  for any integer $k \in \{2,...,K-1\}$ and a fixed $\theta$, there exists a constant $C_{\textnormal{b}}> 1$ such that
\small
\begin{align*}
    & C_{\textnormal{b}}  \mathbb{E}_{\mathcal{A} \sim p_A} \left[ ( A_{\gamma(1)} A^\top_{\gamma(1)} + A_{\gamma(K)} A^\top_{\gamma(K)}) \right.
    \\
    & \qquad \qquad \quad \left. \cdot \indicator \{ \langle A_{\gamma(1)},  \theta \rangle < \ldots < \langle A_{\gamma(K)}, \theta \rangle \} \right]
    \\
    &\succeq \mathbb{E}_{\mathcal{A} \sim p_A}\left[ A_{\gamma(k)} A^\top_{\gamma(k)}  \indicator \{ \langle A_{\gamma(1)}, \theta \rangle < \ldots < \langle A_{\gamma(K)}, \theta \rangle \} \right].
\end{align*}
\normalsize
\end{assumption}

\begin{assumption}[Sparse positive definiteness]\label{asm:cov_div_klarge}
Define for each $B \subset [d]$,
$
\Sigma_B \coloneqq \frac{1}{K} \sum_{k=1}^K \EXP_{\mathcal{A}\sim p_A} \left[A_{k}(B) A_{k}(B)^\top\right],
$
where $ A_{k}(B)$ is a $|B|$-dimensional vector, which is extracted from the elements of $A_{k}$ with indices in $B$.  There exists a positive constant $\alpha>0$ such that $\forall B \subset [d]$,
 \begin{align*}
 & |B| \le s_0 + (4 \nu C_{\textnormal{b}}\sqrt{s_0})/\phi_0^2 \; \text{and} \; S(\theta) \subset B 
 \\
 & \qquad \Longrightarrow  \min_{v \in \mathbb{R}^{|B|} : \|v\|_2 = 1} v^\top \Sigma_B v  \ge \alpha.
 \end{align*}
 The parameters $\phi_0, \nu, C_{\textnormal{b}}$ are those of Assumptions \ref{asm:comp_cond}, \ref{asm:relax_sym}, and~\ref{asm:balanced_cov}.
 \end{assumption}
 Assumption~\ref{asm:relax_sym} comes from \citet{oh2020sparsity}. This assumption is satisfied for the wide range of distributions including multivariate Gaussian, uniform, and  Bernoulli distributions. 
 Assumption~\ref{asm:balanced_cov} is also adopted from \citet{oh2020sparsity}. 
 This assumption holds for a wide range of distributions including multivariate Gaussian distribution, uniform distribution on sphere.
It also holds when contexts are independent across arms with any arbitrary distributions \cite{oh2020sparsity}.  Assumption~\ref{asm:cov_div_klarge} implies that the context distribution is diverse enough in the neighborhood of the support of $\theta$. Note that Assumption~\ref{asm:cov_div_klarge} is standard in low dimensional linear bandit literature (e.g., \citet{lattimore2020bandit, degenne2020gamification, jedra2020optimal,hao2020adaptive}). There, if $S(\theta)=[d]$, the only choice for $B$ is $[d]$, and the set of action has to span $\mathbb{R}^d$ (hence Assumption~\ref{asm:cov_div_klarge} is satisfied). 
We will see that after an accurate estimate of the support $S(\theta)$ (Lemma~\ref{lm:support_recovery_klarge}), Assumption~\ref{asm:cov_div_klarge} is used only to analyze the performance of the least square estimator of low-dimensional (order of $\mathcal{O}(s_0)$) vector. Assumption~\ref{asm:cov_div_klarge} is strictly weaker than the covariate diversity condition of \citet{bastani2021mostly}, where the positive definiteness must be guaranteed for the Gram matrix generated by the greedy algorithm.   We also discuss the details of assumptions in Appendix~\ref{app:assumption_K}.

\section{Algorithm} 

 In this section, we present the Thresholded (TH) Lasso bandit algorithm. The algorithm adapts the idea of Lasso with thresholding proposed in \citet{zhou2010thresholded} to estimate $\theta$ and its support. The main challenge in the analysis of the Lasso with thresholding stems from the fact that here, the data is non i.i.d. (the arm selection is adaptive).

\begin{algorithm}[htb]
  \caption{TH Lasso Bandit}
  \label{algo:th_lasso_bandit}
  \begin{algorithmic}[1]
    \STATE \textbf{Input:} $\lambda_0 $
    \FOR {$t=1,\cdots,T$}
      \STATE Receive a context set $\mathcal{A}_t: = \{A_{t,k} : k \in [K]\}$
      \STATE Pull arm $A_t=\argmax_{A \in \mathcal{A}_t} \langle A, \hat{\theta}_t \rangle$ (ties are broken uniformly at random) and observe $r_t$
      \STATE $\lambda_t \gets  \lambda_0 \sqrt{\frac{2 \log t \log d}{t}}$
      \STATE $A \gets (A_1, A_2, \ldots, A_t)^\top, \quad R \gets (r_1, r_2, \ldots, r_t)^\top$
      \STATE  $\hat{\theta}_0^{(t)} \gets \argmin_{\theta} \frac{1}{t}\|R - A\theta \|_2^2 + \lambda_t \|\theta\|_1$
      \STATE $\hat{S}_0^{(t)} \gets \{j\in [d] :\; |(\hat{\theta}^{(t)}_0)_j| > 4 \lambda_t\}$ 
      \STATE $\hat{S}_1^{(t)} \gets \{j\in \hat{S}_0^{(t)} : \; | (\hat{\theta}^{(t)}_0)_j| \geq 4 \lambda_t \sqrt{|\hat{S}_0^{(t)}|}\}$
      \STATE $A_S \gets (A_{1, \hat{S}_1^{(t)}}, A_{2, \hat{S}_1^{(t)}}, \ldots, A_{t, \hat{S}_1^{(t)}})^\top$
      \STATE $\hat{\theta}_{t+1} \gets \argmin_{\theta}\|R - A_S \theta\|_2^2$
      \ENDFOR
    \end{algorithmic}
\end{algorithm}

The pseudo-code of our algorithm is presented in Algorithm~\ref{algo:th_lasso_bandit}. In round $t$, the algorithm pulls the arm in a greedy way using the estimated value $\hat{\theta}_t$ of $\theta$. From the past selected arms and rewards, we get via the Lasso a first estimate $\hat{\theta}_0^{(t)}$ of $\theta$. This estimate is then used to estimate the support of $\theta$ using appropriate thresholding. The regularizer $\lambda_t \coloneqq \lambda_0 \sqrt{(2 \log t \log d)/t}$ is set at a much larger value than that in  the previous work (they typically have the order of $ \sqrt{(\log d + \log t)/t}$), as we are only focusing on the support recovery here. Note that we apply a thresholding procedure twice to $\hat{\theta}_0^{(t)}$ to provide the support estimate $\hat{S}_1^{(t)}$. The final estimate $\hat{\theta}_{t+1}$ is obtained as the least squares estimator of $\theta$, when restricted to $\hat{S}_1^{(t)}$. The initial support estimate done by Lasso contains too many false positives. By including thresholding steps in the algorithm, we remove the unnecessary false positives and improve the support estimate. We quantify this improvement in the next section.

\section{Performance Guarantees}\label{sec:upper_bounds}

We analyze the regret of the Thresholded Lasso bandit algorithm both when the margin condition holds and when it does not. 
We show that better guarantees can be obtained with a single-parameter or parameter-free algorithm.

\subsection{With the Margin Condition}

 \begin{assumption}[Margin condition]\label{asm:margin}
 There exists a constant $C_{\textnormal{m}} >0$ such that for all $\kappa>0$, 
 \begin{align*}
     \forall k\neq k', \quad \Pr_{A \sim p_{A}}(0 < |\langle A_{k} - A_{k'}, \theta\rangle| \le \kappa) \le C_{\textnormal{m}} \kappa.
 \end{align*}
 \end{assumption}
The margin condition controls the probability that under $p_A$ two arms yield very similar rewards (and hence are hard to separate) and is 
widely used in the classification literature (see e.g., \citet{tsybakov2004optimal, audibert2007fast}). 
For the (low-dimensional) linear bandit literature, it was first introduced in \citet{goldenshluger2013linear}. 
The margin condition holds for the most usual context distributions (including the uniform distribution and Multivariate Gaussian distributions) and a much weaker assumption than requiring the strict separation between the arms.

The following theorem provides a non-asymptotic regret upper bound of TH Lasso bandit under the margin condition. To simplify the presentation of our regret upper bound, define  $\tau = \left\lfloor\frac{2 \log (2 d^2)}{C_0^2} (\log s_0) (\log \log d) \right\rfloor $, where $C_0 =\min \left\{\frac{1}{2}, \frac{\phi_0^2}{512 s_0 s_A^2 \nu C_{\textnormal{b}}}\right\}$.
Note that $\tau = \mathcal{O}\left(s_0^2 (\log s_0) (\log d) (\log \log d)\right)$.
\begin{theorem}\label{thm:regret_ub_klarge}
Assume that Assumptions \ref{asm:sparsity_param_klarge} -- \ref{asm:cov_div_klarge}, \ref{asm:margin} hold.\\
(i) (TH Lasso Bandit with parameter-dependent input) There are universal positive constants $c_1, c_2, c_3$ depending on $ \sigma, s_A, s_1, s_2, \phi_0, \nu, C_{\textnormal{b}},   K,  \alpha, C_{\textnormal{m}}$, such that if we set $\lambda_0=c_1$, then for all $d \ge c_2$, for all $T \ge 2$:
\begin{align*}
& R(T) \le c_3\left( \tau + s_0 (\log s_0)^{\frac{3}{2}}\log T + s_0^2\right).
\end{align*}
(ii) (TH Lasso Bandit with parameter-free input) There are universal positive constants $c_4, c_5$ depending on $ \sigma, s_A, s_1, s_2, \phi_0, \nu, C_{\textnormal{b}},   K,  \alpha, C_{\textnormal{m}}$, such that if we set $ \lambda_0 = {1}/{(\log \log d)^{\frac{1}{4}}}$ in TH Lasso Bandit, then for all $d \ge c_4$, for all $T \ge 2$,
\begin{align*}
     R(T)\le c_5\left( \tau + s_0 (\log s_0)^{\frac{3}{2}}\log T + s_0^2\right).
\end{align*}
The precise definitions of $c_1$-$c_5$ are given in Appendix~\ref{subsec:proof_regret_ub_klarge}.
\end{theorem}
We provide the proof of Theorem~\ref{thm:regret_ub_klarge} in Appendix~\ref{subsec:proof_regret_ub_klarge}.

\subsection{Without the Margin Condition}

\begin{theorem}\label{thm:regret_ub_wo_margin_klarge}
Assume that Assumptions \ref{asm:sparsity_param_klarge} -- \ref{asm:cov_div_klarge} hold.\\
(i) (TH Lasso Bandit with parameter-dependent input) There are universal positive constants $c_1, c_2, c_3$ depending on $ \sigma, s_A, s_1, s_2, \phi_0, \nu, C_{\textnormal{b}},   K,  \alpha$ such that if we set
$\lambda_0 = c_1 $, then for all $d \ge c_2$. for all $T \ge 2$:
\begin{align*}
& R(T) \le c_3\left(  \tau +  (\log s_0) \sqrt{s_0 T} + s_0^2\right).
\end{align*}
(ii) (TH Lasso Bandit with parameter-free input) There are universal positive constants $c_4, c_5$ depending on $ \sigma, s_A, s_1, s_2, \phi_0, \nu, C_{\textnormal{b}},   K,  \alpha$ such that if we set $ \lambda_0 = {1}/{(\log \log d)^{\frac{1}{4}}}$ in TH Lasso Bandit, then for all $d \ge c_4$, for all $T \ge 2$,
\begin{align*}
     R(T)\le c_5\left( \tau +  (\log s_0) \sqrt{s_0 T} + s_0^2\right).
\end{align*}
The precise definitions of $c_1$-$c_5$ are given in Appendix~\ref{subsec:regret_ub_wo_margin_klarge}
\end{theorem}
The proof of Theorem~\ref{thm:regret_ub_wo_margin_klarge} is presented in Appendix~\ref{subsec:regret_ub_wo_margin_klarge}.

Theorems \ref{thm:regret_ub_klarge} and \ref{thm:regret_ub_wo_margin_klarge} state that TH Lasso bandit achieves much lower regret than the existing algorithms. Indeed, upper regret bounds for the latter had a term scaling as $\log d \log T$ (resp. $\log d + \sqrt{T\log (dT)}$) with (resp. without) the margin condition. TH Lasso bandit removes the $\log d$ and $\log T$ multiplicative factors. In most applications of the sparse linear contextual bandit, both $T$ and $d$ are typically very large, and the regret improvement obtained by TH Lasso bandit is significant. 
Also note that our regret upper bounds match the minimax lower bound $\Omega(\sqrt{s_0 T}) $ proved in \citet{ren2020dynamic}.

\subsection{Sketch of the Proof of Theorems}

We sketch below the proof of Theorem \ref{thm:regret_ub_klarge} and \ref{thm:regret_ub_wo_margin_klarge}. Complete proofs of Theorems and associated Lemmas are presented in Appendix~\ref{sec:Proof_theorems} and Appendix~\ref{sec:proof_lemmas_klarge}, respectively.

\paragraph{(1) Estimation of the Support of $\theta$.} First, we prove that the estimated support contains the true support $S(\theta)$ with high probability.

\begin{lemma}\label{lm:support_recovery_klarge}
 Let $t \ge  \frac{2 \log (2 d^2)}{C_0^2}$ such that $4 \left( \frac{4 \nu C_{\textnormal{b}} s_0}{\phi_0^2} + \sqrt{\left(1 + \frac{4\nu C_{\textnormal{b}}}{\phi_0^2}\right)s_0}\right)\lambda_t \le \theta_{\min}$. Under Assumptions \ref{asm:sparsity_param_klarge}, \ref{asm:comp_cond}, \ref{asm:relax_sym}, and \ref{asm:balanced_cov}, 
$
    \Pr \left(S(\theta) \subset \hat{S}_1^{(t)} \text{ and } |\hat{S}_1^{(t)} \setminus S(\theta)| \le \frac{4\nu C_{\textnormal{b}} \sqrt{s_0}}{\phi^2_0}\right)
    $
    $
     \ge 1 -  2 \exp\left( -\frac{t \lambda_t^2}{32 \sigma^2 s_A^2} + \log d\right) - \exp\left( - \frac{t C_0^2}{2}\right).
$
\end{lemma}
\noindent Lemma~\ref{lm:support_recovery_klarge} extends the support recovery result of the Thresholded Lasso \cite{zhou2010thresholded} to the case of non-i.i.d data (generated by the bandit algorithm). 
The dependence on $s_0$ is analogous to the offline result (Theorem~3.1 of \citet{zhou2010thresholded}). As it can be seen from the proof, even after the single-step thresholding, for all sufficiently large $t$, we have the guarantee of $S(\theta) \subset \hat{S}_0^{(t)}$ and $ | \hat{S}_0^{(t)} \setminus S(\theta)| = \mathcal{O}(s_0)$ with high probability. However, with the two-step thresholding, we have $ | \hat{S}_1^{(t)} \setminus S(\theta)| = \mathcal{O}(\sqrt{s_0})$ (See Appendix~\ref{app:benefit_thresholding} for the benefit of the two-step thresholding in detail). 

\begin{remark}
In the proof of Lemma~\ref{lm:support_recovery_klarge}, we also obtain an bound on the estimation error of $\hat{\theta}_0^{(t)}$ (by the Lasso).
One may directly use $\hat{\theta}_0^{(t)}$ for the arm selection as is in \citet{oh2020sparsity}, however, this results in a weaker performance guarantee of the order $\mathcal{O} ( \log d + \sqrt{T \log (dT)})$ (without the margin condition).
This is due to the fact that the estimation error of $\hat{\theta}_0^{(t)}$ has a dependence on $d$, which impacts the order of the regret. 
This motivates the use of the thresholding procedure. With this procedure (i.e., using $\hat{\theta}_t$), we remove the dependence in $d$ of the estimation error when $t$ is larger than $\tau$, which, in turn, leads to an instantaneous regret bound independent of $d$. In summary, the thresholding procedure allows us to derive better regret bounds than those in existing work (e.g., \citet{oh2020sparsity}).
\end{remark}

Define
    $\mathcal{E}_t \coloneqq \left\{S(\theta) \subset \hat{S}_1^{(t)} \;\text{and}\; |\hat{S}_{1}^{(t)} \setminus S(\theta)| \le  \frac{4 \nu C_{\textnormal{b}} \sqrt{s_0}}{\phi^2_0} \right\}.$
In the remaining of the proof, in view of Lemma~\ref{lm:support_recovery_klarge}, we can assume that the event $\mathcal{E}_t$ holds.

\paragraph{(2) Minimal Eigenvalue of the Empirical Gram Matrix.}  
We write $\hat{S}_1^{(t)} = \hat{S}$ for the simplicity. Let $\hat{\Sigma}_{\hat{S}} \coloneqq \frac{1}{t} \sum_{s = 1}^t A_{s}(\hat{S})  A_{s}(\hat{S})^\top$ be the empirical Gram matrix on the estimated support.  We prove that the positive definiteness of the empirical Gram matrix on the estimated support is guaranteed. 
\begin{lemma}\label{lm:lowerbound_eig_S_klarge}
 Let $t \in [T]$. Under Assumptions \ref{asm:sparsity_param_klarge} and \ref{asm:cov_div_klarge}, we have:
$
    \Pr\left(\lambda_{\min} (\hat{\Sigma}_{\hat{S}}) \ge   \frac{\alpha}{4\nu  C_{\textnormal{b}}} \ \big| \ \mathcal{E}_t \right)   \ge 1 -  \exp\left( \log\left(s_0 + \frac{4 \nu C_{\textnormal{b}}\sqrt{s_0}}{\phi_0^2} \right)
 - \frac{t \alpha}{20 s_A^2\nu C_{\textnormal{b}}\left(s_0 + (4 \nu C_{\textnormal{b}} \sqrt{s_0})/\phi_0^2\right) } \right). 
$
\end{lemma}

\paragraph{(3) Estimation of $\theta$ after Thresholding.} Next, we study the accuracy of $\hat{\theta}_{t}$.

\begin{lemma}\label{lm:LS_estim_risk_klarge}
 Let $t \in [T]$ and $s' = s_0 + 4 \nu C_{\textnormal{b}} \sqrt{s_0}/\phi_0^2$. Under Assumption~\ref{asm:sparsity_param_klarge}, we have, for all $x, \lambda >0$:
$
     \Pr \left( \|\hat{\theta}_{t+1} - \theta \|_2 \ge x \; \text{and} \;\lambda_{\min}(\hat{\Sigma}_{\hat{S}}) \ge \lambda \ \big| \ \mathcal{E}_t \right)  
$
$
\le2 s' \exp \left( - \frac{\lambda^2 t x^2}{2 \sigma^2 s_A^2 s'}\right).$
\end{lemma}

From the above lemma, we conclude that $\theta$ is well estimated with high probability. Note that in the above estimation error, the dependency in $s_0$ can be also improved from linear to square root compared with the analysis of Lemma~1 (Oracle inequality) in \citet{oh2020sparsity} (SA Lasso bandit). This stems from the fact that using the compatibility condition, one can only control the $\ell_1$ norm of the estimation error of $\theta$, while using the OLS leads to an $\ell_2$ guarantee. 

\paragraph{(4) Instantaneous Regret Upper Bound with the Margin Condition.}
For the previous lemmas, we can derive an upper bound on the instantaneous regret with the margin condition. Define $h_0 = \lfloor \sqrt{\log(4 (s_0 + \frac{4 \nu C_{\textnormal{b}}\sqrt{s_0}}{\phi_0^2}))}  + 1 \rfloor$.
 
\begin{lemma}[With the margin condition]\label{lm:inst_regret_bound_klarge}
Define $\mathcal{G}_{t}^{\frac{\alpha}{4\nu C_{\textnormal{b}}}} \coloneqq \left\{\lambda_{\min } ( \hat{\Sigma}_{\hat{S}} ) \ge \frac{\alpha}{4\nu C_{\textnormal{b}}} \right\}$. Let $t \ge 2$. Under Assumptions \ref{asm:sparsity_param_klarge}, \ref{asm:comp_cond}, \ref{asm:relax_sym}, \ref{asm:balanced_cov}, \ref{asm:cov_div_klarge}, and \ref{asm:margin}, the expected instantaneous regret $\EXP[\max_{A \in \mathcal{A}_t} \langle A - A_t, \theta \rangle]$ is upper bounded by:
\begin{align*}
& \frac{1408 \sigma^2 s_A^4 C_{\textnormal{m}} (K-1) h_0^3 \nu^2 C_{\textnormal{b}}^2 \left(s_0 + \frac{4 \nu C_{\textnormal{b}}\sqrt{s_0}}{\phi_0^2}\right)}{\alpha^2} \frac{1}{t - 1} 
\\
& + 2 (K-1)  s_A s_1  \left(\Pr(\mathcal{E}_t^c) + \Pr\left(\left(\mathcal{G}_{t}^{\frac{\alpha}{4 \nu C_{\textnormal{b}}}}\right)^c \middle| \mathcal{E}_t\right) \right).
\end{align*}
\end{lemma}
Notice that the first term of this instantaneous regret bound does not depend on $d$. This leads to the better regret order.

On the other hand, without the margin condition, we present the key lemma, which is proven by a novel application of the discretization technique:
\begin{lemma}[Without the margin condition]\label{lm:inst_regret_bound_wo_margin_klarge}
Under Assumptions \ref{asm:sparsity_param_klarge},  \ref{asm:comp_cond},  \ref{asm:relax_sym}, \ref{asm:balanced_cov}, and \ref{asm:cov_div_klarge}, for any $t \in [T]$, $\EXP[\max_{A \in \mathcal{A}_t} \langle A - A_t, \theta\rangle]$ is upper bounded by
\begin{align*}
& \frac{36 \sigma s_A  (K-1) h_0^2 \nu C_{\textnormal{b}}}{\alpha}\sqrt{\frac{2 \left(s_0 + \frac{4 \nu C_{\textnormal{b}}\sqrt{s_0} }{\phi_0^2}\right)}{t-1}}
\\
& + 2  (K-1) s_A s_1  \left(\Pr(\mathcal{E}_t^c) + \Pr\left(\left(\mathcal{G}_{t}^{\frac{\alpha}{4\nu C_{\textnormal{b}}}}\right)^c \middle| \mathcal{E}_t\right) \right).
\end{align*}
\end{lemma}
Again, the first term in the above bound is independent of $d$. Note also that compared to existing work, we improve the dependence in $t$: we get $1/\sqrt{t}$ instantaneous upper bound while in the other recent work (e.g., \citet{kim2019doubly, oh2020sparsity}) they get $\sqrt{\log t /t}$. As a consequence, we obtain better regret guarantees without the margin condition. More precisely, we manage to remove the unnecessary $\log T$ factors in the regret that was present in all previous studies.

By summing up these instantaneous regret bounds, we get Theorem~\ref{thm:regret_ub_klarge} and \ref{thm:regret_ub_wo_margin_klarge}. 
In Appendix~\ref{sec:additional_theorems}, we also provide regret guarantees without Assumption~\ref{asm:balanced_cov} but when $K=2$ (Theorem~\ref{thm:regret_ub_k2}) and without the margin condition (Theorem~\ref{thm:regret_ub_wo_margin_k2}). 
These theorems are established using the relaxed symmetry assumption only; see also the proof of Lemma~2 in \citet{oh2020sparsity}.

\begin{figure*}[t!]
    \centering
    \begin{minipage}[t]{0.32\textwidth}
        \centering
        \includegraphics[width=1.1\textwidth]{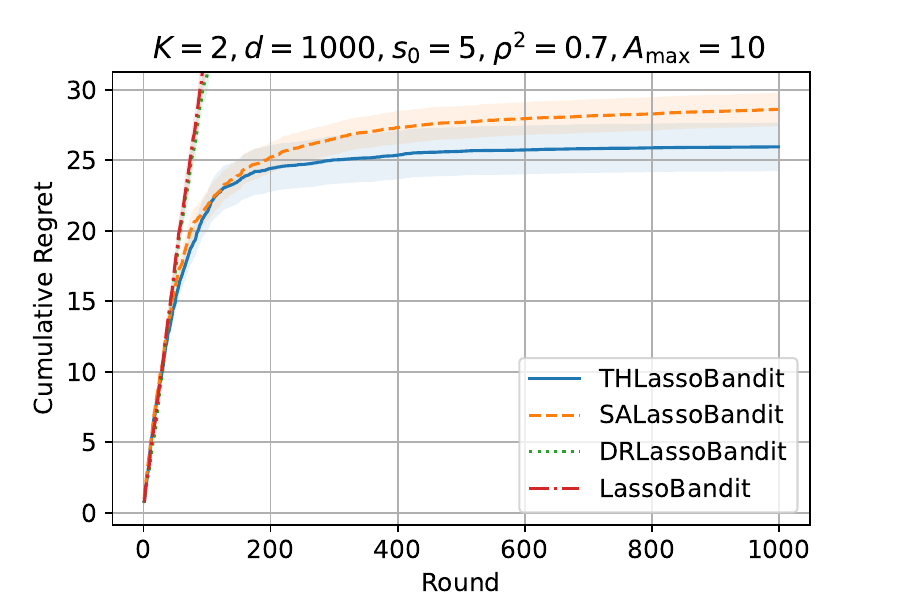}
    \end{minipage}
    \begin{minipage}[t]{0.32\textwidth}
        \centering
        \includegraphics[width=1.1\textwidth]{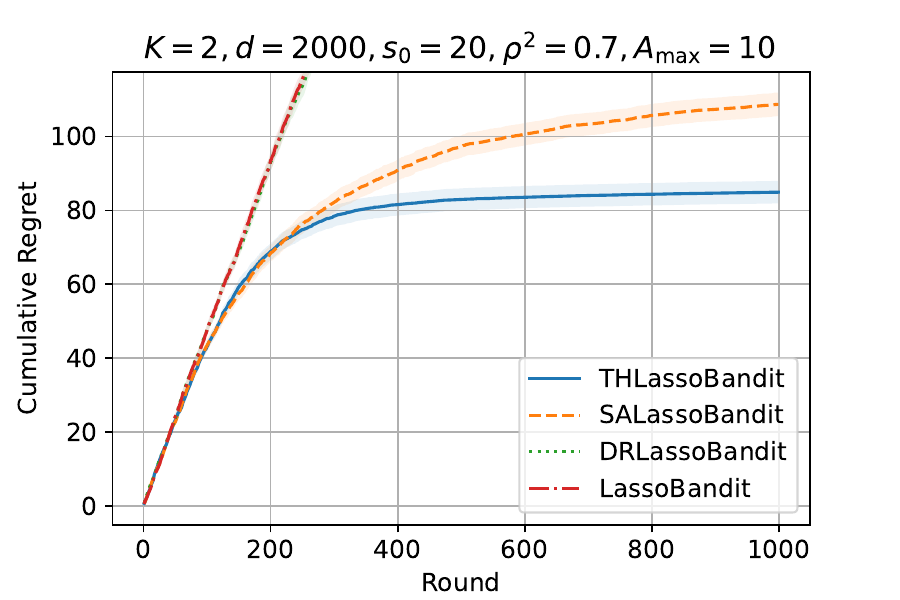}
    \end{minipage}
    \begin{minipage}[t]{0.32\textwidth}
        \centering
        \includegraphics[width=1.1\textwidth]{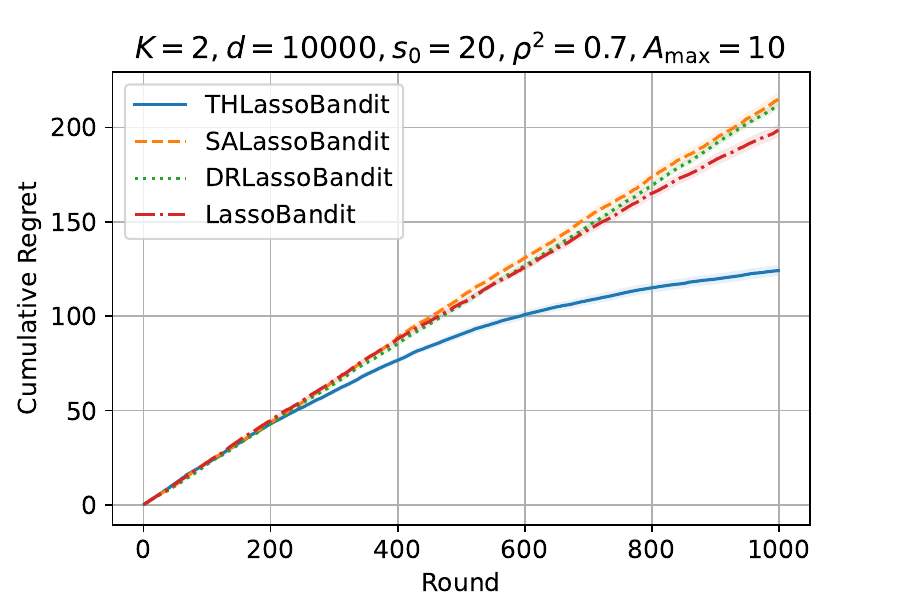}
    \end{minipage} \\
    \begin{minipage}[t]{0.32\textwidth}
        \centering
        \includegraphics[width=1.1\textwidth]{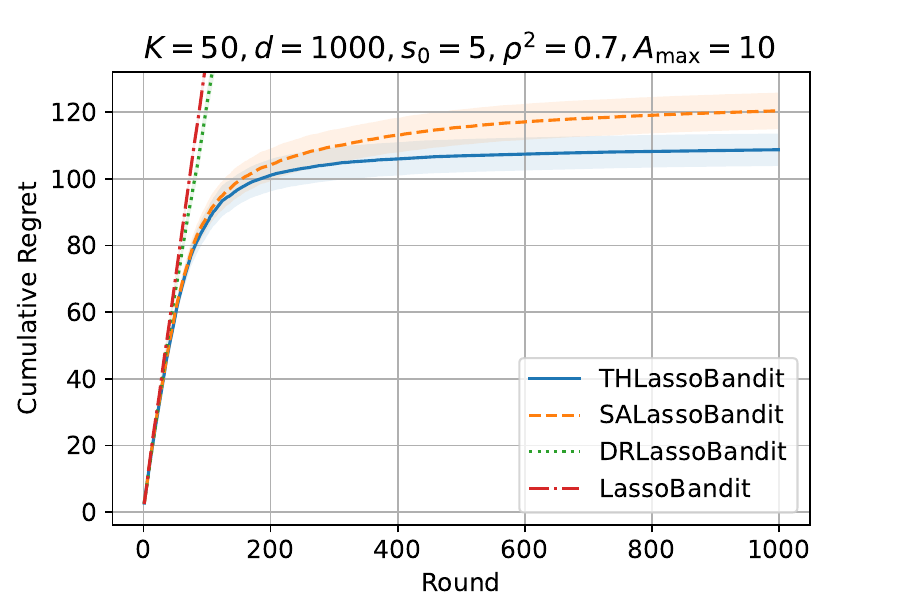}
    \end{minipage}
    \begin{minipage}[t]{0.32\textwidth}
        \centering
        \includegraphics[width=1.1\textwidth]{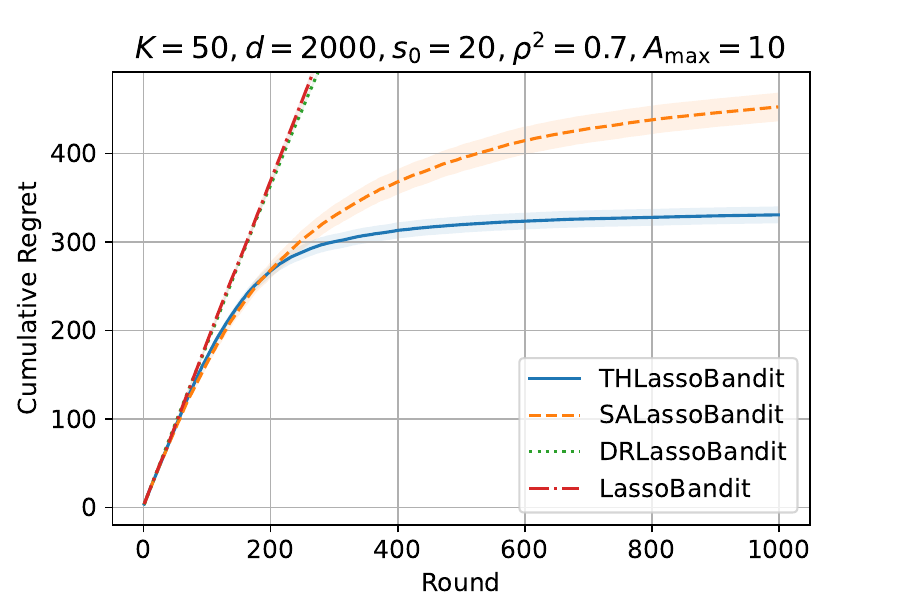}
    \end{minipage}
    \begin{minipage}[t]{0.32\textwidth}
        \centering
        \includegraphics[width=1.1\textwidth]{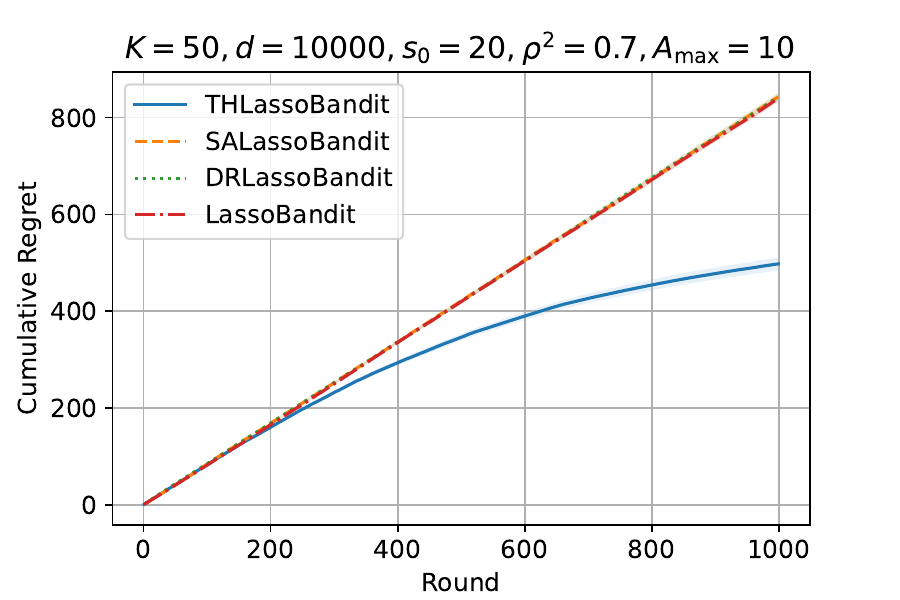}
    \end{minipage}
    \caption{Cumulative regret of the three algorithms with $\rho^2=0.7$, $A_{\max}=10$ in six scenarios selected using $K\in \{2, 50\}$, $d\in \{1000, 2000, 10000\}$, and $s_0\in \{5, 20\}$.
    The shaded area represents the standard errors.}
    \label{fig:experiment1_regrets}
\end{figure*}

\begin{figure*}[t!]
    \centering
    \begin{minipage}[t]{0.32\textwidth}
        \centering
        \includegraphics[width=1.1\textwidth]{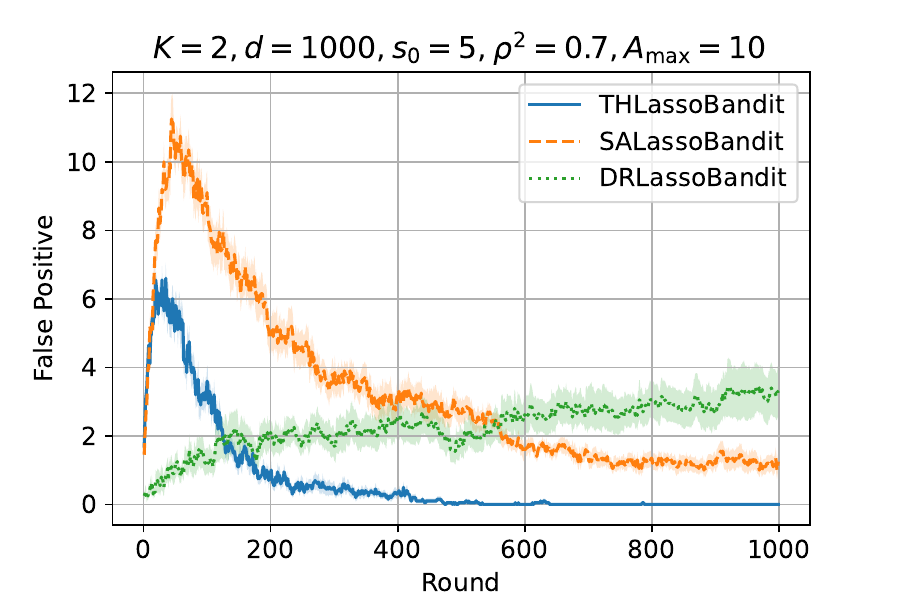}
    \end{minipage}
    \begin{minipage}[t]{0.32\textwidth}
        \centering
        \includegraphics[width=1.1\textwidth]{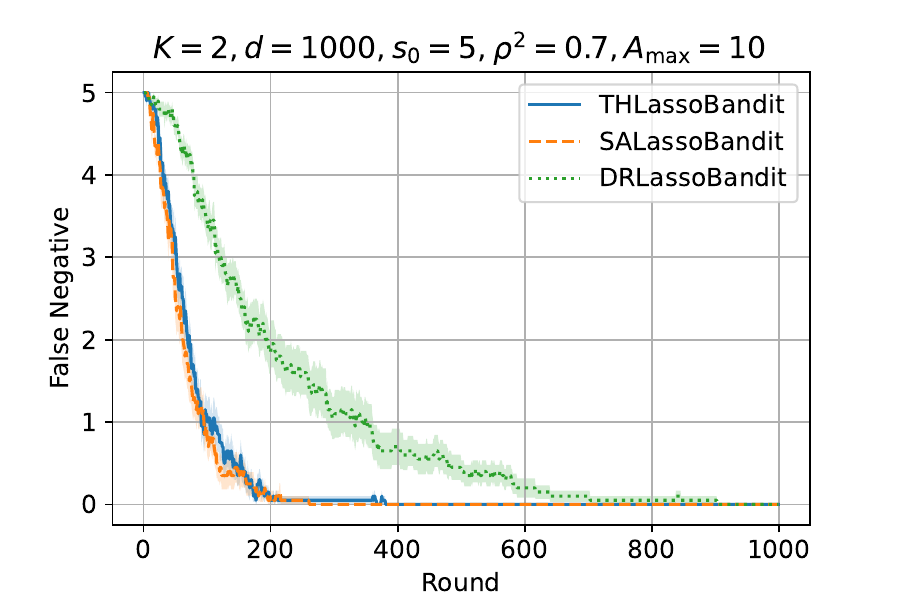}
    \end{minipage}
    \begin{minipage}[t]{0.32\textwidth}
        \centering
        \includegraphics[width=1.1\textwidth]{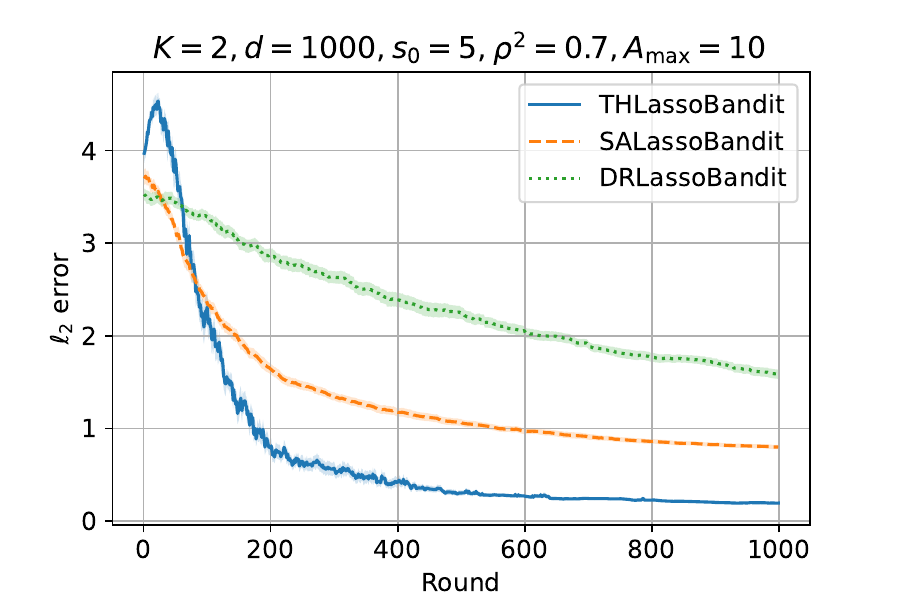}
    \end{minipage}
    \caption{(Left) Number of false positives $|\hat{S}_1^{(t)} \setminus S(\theta)|$, (center) false negatives $|S(\theta) \setminus \hat{S}_1^{(t)}|$, (right) $\ell_2$-norm error $\|\hat{\theta}_t - \theta\|_2$ of the three algorithms with $\rho^2=0.7$, $A_{\max}=10$, $K=2$, $s_0=5$, and $d=1000$. The shaded area represents the standard errors.}
    \label{fig:experiment1_error}
\end{figure*}

\begin{figure*}[t!]
\begin{minipage}{0.48\textwidth}
\centering
\includegraphics[width=0.96\textwidth]{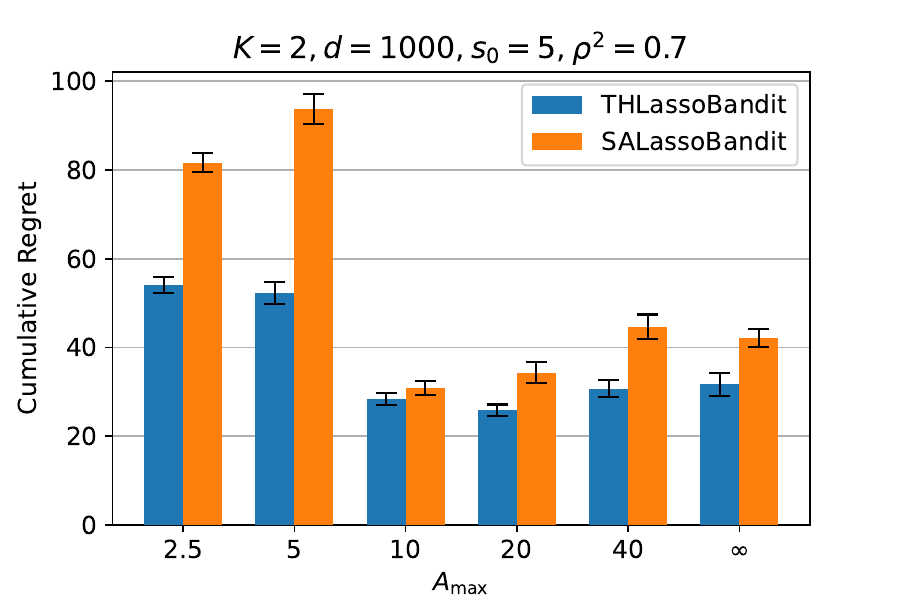}
\caption{Cumulative regret at round $t=1000$ of TH Lasso bandit and SA Lasso bandit with $\rho^2=0.7$, $K=2$, $d=1000$, $s_0=5$, and varying $A_{\max}\in \{2.5, 5, 10, 20, 40, \infty\}$.
The error bars represent the standard errors.}
\label{fig:experiment2_regrets}
\end{minipage}
\hfill
\begin{minipage}{0.48\textwidth}
\centering
\includegraphics[width=0.96\textwidth]{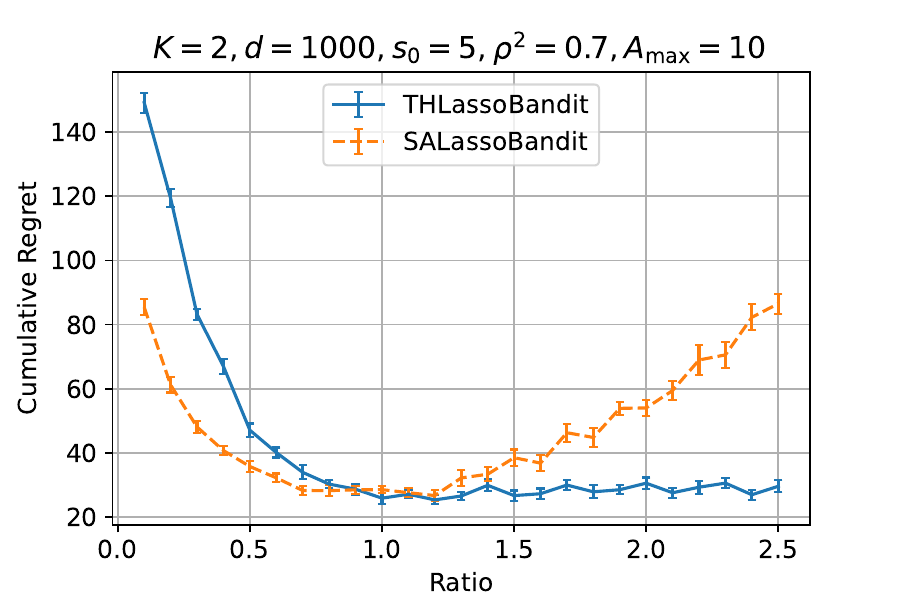}
\caption{Cumulative regret at round $t=1000$ of TH Lasso bandit and SA Lasso bandit with $\rho^2=0.7$, $A_{\max}=10$, $K=2$, $d=1000$, $s_0=5$, and varying $\lambda_0/\lambda^{\ast}\in [0.1, 2.5]$.
The error bars represent the standard errors.}
\label{fig:discussion_regrets}
\end{minipage}
\end{figure*}

\section{Experiments}

In this section, we empirically evaluate the TH Lasso bandit algorithm. 
We compare its performance to those of the Lasso Bandit \cite{bastani2020online}, Doubly-Robust (DR) Lasso bandit \cite{kim2019doubly}, and SA Lasso bandit \cite{oh2020sparsity} algorithms. Note that the Lasso bandit algorithm \cite{bastani2020online} deals with a slightly different problem setting ($\theta$ varies across arms in their setting). We follow the comparison ideas in \citet{kim2019doubly} (considered the $Kd$-dimensional context vectors and the $Kd$-dimensional regression parameters for each arm. For details, see \citet{kim2019doubly}).

\paragraph{Reward Parameter and Contexts.} We consider problems where $\theta\in \mathbb{R}^d$ is sparse, i.e., $\|\theta\|_0=s_0$. We generate each non-zero components of $\theta$ in an i.i.d. manner using the uniform distribution on $[1, 2]$.
In each round $t$, for each component $i\in [d]$, we sample $((A_{t,1})_{i}, \cdots, (A_{t,K})_{i})^{\top} \in \mathbb{R}^K$ from a Gaussian distribution $\mathcal{N}(\mathbf{0}_{K}, V)$ where $V_{i,i}=1$ for all $i\in [K]$ and $V_{i,k}=\rho^2=0.7$ for all $i\neq k\in [K]$. We then normalize each $A_{t,k} = ((A_{t,k})_1, \ldots, (A_{t,k})_d)^{\top} \in \mathbb{R}^d$ so that its $\ell_2$-norm is at most $A_{\max}$ for all $k \in [K]$.
Note that the components of the feature vectors are correlated over $[d]$ and over $[K]$. The noise process is Gaussian, i.i.d. over rounds: $\varepsilon_t\sim\mathcal{N}(0, 1)$. We test the algorithms for different values of $K, d, s_0$, and $A_{\max}$. For each experimental setting, we averaged the results for $20$ instances. We also provide additional experimental results with non-Gaussian distributions in Appendix~\ref{sec:exp_non_gaussian}.

\begin{remark}
In most of our experiments, the context is drawn from a multivariate Gaussian, or uniform distribution on $[-1,1]^d$.
In this case, the minimum eigenvalue of the gram matrix $\Sigma$ is lower bounded by some constant. 
Hence, Assumptions~\ref{asm:comp_cond} and \ref{asm:cov_div_klarge} are satisfied. 
Clearly, Assumption~\ref{asm:relax_sym} is satisfied by the symmetry of the distribution. 
When the distribution is independent over the arms, from Proposition~1 in \citet{oh2020sparsity}, Assumption~\ref{asm:balanced_cov} is satisfied. 
Since each element of the context distribution has a bounded density everywhere, Assumption~\ref{asm:margin} is also satisfied. 
Furthermore, in Appendix~\ref{subsec:numerical_hard_instances}, we empirically tested our algorithm for some hard problems where the covariate diversity condition \cite{bastani2021mostly} does not hold. 
\end{remark}

\paragraph{Algorithms.} For DR Lasso bandit and Lasso bandit, we use the tuned hyperparameter at\\ \url{https://github.com/gisoo1989/Doubly-Robust-Lasso-Bandit}.\\
For the SA Lasso bandit and TH Lasso bandit algorithms, we tune the hyperparameter $\lambda_0$ in $[0.01, 0.5]$ to roughly optimize the algorithm performance when $K=2, d=1000, A_{\max}=10$, and $s_0=5$. As a result, we set $\lambda_0=0.16$ for SA Lasso bandit, and set $\lambda_0=0.02$ for TH Lasso bandit.

\paragraph{Results.} We first compare the regret of each algorithm with $A_{\max}=10$, $K\in\{2, 50\}$, $d=\{1000, 2000, 10000\}$, and $s_0\in \{5, 20\}$. We experimented with larger values of $d$, in addition to the one in existing studies.
Figure~\ref{fig:experiment1_regrets} shows the average cumulative regret for each algorithm.
We find that TH Lasso bandit outperforms the other algorithms in all scenarios. We provide additional experimental results, including experiments with different correlation levels $\rho^2(\in\{0, 0.3\})$ and dimension $d$, in Appendix~\ref{sec:exp_non_gaussian}. 

Next, we compare the estimation accuracy for $\theta$ under  three algorithms (DR, SA, and TH Lasso bandit) in the scenario: $K=2, d=1000, A_{\max}=10, s_0=5$.
Figure~\ref{fig:experiment1_error} shows the number of false positives $|\hat{S}_1^{(t)} \setminus S(\theta)|$, the number of false negatives $|S(\theta) \setminus \hat{S}_1^{(t)}|$, and $\ell_2$-norm error $\|\hat{\theta}_t - \theta\|_2$.
Note that, for DR Lasso bandit and SA Lasso bandit, we define the estimated support as $\hat{S}_1^{(t)}=\{i\in [d]: \hat{\theta}_{t,i}\neq 0\}$. We can observe that the number of false positives of our algorithm converge to zero faster than those of DR Lasso bandit and SA Lasso bandit.
Furthermore, our algorithm yields a smaller estimation error $\|\hat{\theta}_t - \theta\|_2$ than the two other algorithms, as is shown in right column of Figure~\ref{fig:experiment1_error}.

We also conduct experiments varying $A_{\max}\in \{2.5, 5, 10, 20, 40, \infty\}$. As in the previous experiments, for each $A_{\max}$, we normalize each feature vector $A_{t,k}$ so that its $\ell_2$-norm is at most $A_{\max}$ for all $k \in [K]$.
We set $K=2, d=1000$, and $s_0=5$. Figure~\ref{fig:experiment2_regrets} shows the average cumulative regret at $t=1000$ of TH Lasso bandit and SA Lasso bandit for each $A_{\max}$. This experiment confirms that TH Lasso exhibits lower regret than SA Lasso bandit. Additional results when $K=50$ and $s_0=20$ are also included in Appendix~\ref{subsec:additional_Amax}.

Finally, we examine the robustness of TH Lasso bandit and SA Lasso bandit with respect to the hyperparameter $\lambda_0$. We vary $\lambda_0 \in [0.1\lambda^{\ast}, 2.5\lambda^{\ast}]$ where $\lambda^{\ast}= 0.02$ for TH Lasso bandit and $\lambda^{\ast}= 0.16$ for SA Lasso bandit. We set $K=2, d=1000, s_0=5$, and $A_{\max}=10$. Figure~\ref{fig:discussion_regrets} shows the average cumulative regret at $t=1000$ for TH Lasso bandit and SA Lasso bandit for different ratios $\lambda_0/\lambda^{\ast}$.
Observe that the regret of TH Lasso bandit is more stable than that of SA Lasso bandit as the  ratio grows. Indeed, the performance of TH Lasso bandit is not very sensitive to the choice of $\lambda_0$: it is robust. This contrasts with the SA Lasso bandit algorithm, for which a careful tuning of $\lambda_0$ is needed to get good performance.

\section{Conclusion}
In this paper, we studied the high-dimensional contextual linear bandit problem with sparsity. We devised TH Lasso bandit, a simple algorithm that applies a Lasso procedure with thresholding  to estimate the support of the unknown parameter. We established finite-time regret upper bounds under various assumptions, and in particular with and without the margin condition. These bounds exhibit a better regret scaling than those derived for previous algorithms. We also numerically compared TH Lasso bandit to previous algorithms in a variety of settings, and showed that it outperformed other algorithms in these settings.

In future work, it would be interesting to consider scenarios where the assumptions made in this paper may not hold. In particular, it is worth investigating the case where the relaxed symmetry condition (Assumption~\ref{asm:relax_sym}) is not satisfied. In this case, being greedy in the successive arm selections may not work. It is intriguing to know whether devising an algorithm without forced uniform exploration and with reasonable regret guarantees is possible.

\section*{Acknowledgements}

We would like to thank Komei Fujita, Yusuke Kaneko, Hiroaki Shiino, and Shota Yasui for the fruitful discussions. 
We also thank anonymous reviewers for helpful comments on the previous version of our manuscript.
K. Ariu was partially supported by the Nakajima Foundation Scholarship. A. Proutiere's research is partially supported by the Wallenberg AI, Autonomous Systems and Software Program (WASP) funded by the Knut and Alice Wallenberg Foundation, and by Digital Futures.

\bibliography{references}
\bibliographystyle{icml2022}

\newpage
\appendix
\onecolumn

\newpage
\noindent {\Large \bf Appendix}

\section{Table of Notations}\label{app:table_of_notations}

Table~\ref{tab:notions} summarizes the notations used in the paper.
\begin{table}[htbp]	
	\caption{Table of notations}
	\label{tab:notions}
	\begin{center}%
		\footnotesize
		\begin{tabular}{c c p{10cm} }
			\toprule
			\multicolumn{3}{l}{\bf Problem-specific notations}\\
			\hline
			$A_{t,k} $ &   & Feature vector associated with the arm $k$
			\\ 
			$\theta$ &   & Parameter vector
			\\ 
			$d$ & & Dimension of feature vectors 
			\\
			$ s_0$ & & Sparsity index
			\\
			 $T$ & & Total number of rounds 
			 \\
			$\mathcal{A}_t$ & & Set of context vectors at round $t$ 
			\\
			$p_{A}$ & & Distribution for $\mathcal{A}_t$
			\\
			$r_t$ & & Reward at round $t$
			\\
			$\mathcal{F}_t$ & & $\sigma$-algebra generated by random variables~$(\mathcal{A}_1,A_1,r_1, \ldots,\mathcal{A}_{t-1},A_{t-1},r_{t-1}, \mathcal{A}_{t})$
			\\
			$\varepsilon_t$ & & Zero mean sub-Gaussian noise 
			\\
			$\sigma^2$ & & Variance proxy of $\varepsilon_t$
			\\
			$R(T)$ & & Regret 
			\\
			$\hat{\Sigma}_t$  & & Empirical Gram matrix generated by the arms selected under a specific algorithm, i.e., $\frac{1}{t}\sum_{s=1}^t A_s A_s^\top$
			\\
			$S(\theta)$ & & Support of $\theta$: $\{i \in [d] : \theta_i \neq 0\}$
			\\
            $\theta_{\min}$ & & $\min_{i \in S}|\theta_i|$
			\\
			$s_A$ & & $\ell_\infty$ norm upper bound on $A \in \mathcal{A}_t$ (see Assumption~\ref{asm:sparsity_param_klarge})
			\\
			$s_1$ & & $\ell_1$ norm upper bound on $\theta$ (see Assumption~\ref{asm:sparsity_param_klarge})
			\\
			$s_2$ & & Used for the lower bound on $\theta_{\min}$ (see Assumption~\ref{asm:sparsity_param_klarge})
			\\
			$\phi^2(M, S_0)$ & & Compatibility constant (see Assumption~\ref{asm:comp_cond})
			\\
			$\Sigma$& & Expected Gram matrix $ \frac{1}{K} \sum_{k=1}^K \EXP_{\mathcal{A}\sim p_A} \left[A_{k} A_{k}^\top\right]$
			\\
			$\phi^2_0$& &  Lower bound on $\phi^2(\Sigma, S(\theta))$
			\\
			$\nu$ & & Constant for Relaxed symmetry (see Assumption~\ref{asm:relax_sym})
			\\
			$C_{\textnormal{b}}$ & & Constant for Balanced covariance (see Assumption~\ref{asm:balanced_cov})
			\\
			$\alpha$ & & Constant for Sparse positive definiteness (see Assumption~\ref{asm:cov_div_klarge} and \ref{asm:cov_div})
			\\
			$\lambda_t$ & & Regularizer at round $t$
			\\
			$\lambda_0$ & & Coefficient of the regularizer 
			\\
			$ \hat{S}_0^{(t)}, \hat{S}_1^{(t)}$ & & Estimate of the support after the first and the second thresholding, respectively.
			\\
			$\hat{\theta}_t$& &  Estimated vector of $\theta$
			\\
			$ C_{\textnormal{m}}$ & & Constant for the margin condition (see Assumption~\ref{asm:margin})\\
			$h_0$ & & Term whose order is $\mathcal{O}((\log s_0)^{\frac{1}{2}})$ (see definitions before the Lemmas)
			\\
			$C_0$& &  Term whose order is $\mathcal{O}(1/s_0)$ (see definitions before the Theorems)
			\\
			$\tau$ & &  $ \left\lfloor\frac{2 \log (2 d^2)}{C_0^2} (\log s_0) (\log \log d) \right\rfloor $
			\\
			$\mathcal{E}_t$ & & Event $\left\{S \subset \hat{S}_1^{(t)}  \text{ and } |\hat{S}_{1}^{(t)} \setminus S| \le  \frac{4 \nu C_{\textnormal{b}} \sqrt{s_0}}{\phi^2_0} \right\}$ or 
			\\
			 & & $\qquad \left\{S \subset \hat{S}_1^{(t)}  \text{ and } |\hat{S}_{1}^{(t)} \setminus S| \le  \frac{2 \nu  \sqrt{s_0}}{\phi^2_0} \right\}$ 
			\\
			$\hat{S}$ & & Estimate of the support after the second thresholding (Equivalent to $\hat{S}_1^{(t)}$)
			\\
			$\hat{\Sigma}_{\hat{S}}$ & & $\frac{1}{t} \sum_{s = 1}^t A_{s}(\hat{S})  A_{s}(\hat{S})^\top$
			\\
			$\mathcal{G}_{t}^{\lambda}$  & & Event $\left\{\lambda_{\min } ( \hat{\Sigma}_{\hat{S}} ) \ge \lambda \right\}$
			\\
			$A_{\max}$ & & $ \ell_2$ norm bound on $A_{t, k}$ (used in the experiments)
			\\ \multicolumn{3}{c}{}\\
			\hline
			 \multicolumn{3}{l}{\bf Generic notations}\\
			 \hline
			 $\|x\|_0$& & $\ell_0$ norm of $x$, i.e., $\|x\|_0 = \sum_{i=1}^d\indicator\{\theta_i \neq 0\}$
			 \\
			 $[x]$ & & Set of positive integers upto $x$, i.e., $[x] = \{1, \ldots, x\}$
			 \\
			 $\langle x, y \rangle$ & & Inner product of $x$ and $y$
			 \\
			 $\Pr(A)$& & Probability that event $A$ occurs
			 \\
			 $\EXP[a]$ & & Expected value of $a$
			 \\
			 $\theta_{i, B}$ & &  $\theta_i \indicator \{ i \in B\}$
			 \\
			 $\theta_B$ & &  $(\theta_{1, B}, \ldots, \theta_{d, B})^\top$
			 \\
			 $A(B)$ & & $ n \times |B|$ submatrix of $A\in \mathbb{R}^{n\times d}$ where $B\subset [d]$
			 \\
			 $\text{supp}(x)$ & & Set of the non-zero element indices of $x$
			\\ \bottomrule
		\end{tabular}
		\normalsize
	\end{center}
\end{table}

\section{Discussion on the Assumptions and Regret Dependence on $K$}
\label{app:assumption_K}

 Our assumptions are in principle following the literature \citet{oh2020sparsity}.
In the contextual linear bandit setting, Assumptions \ref{asm:relax_sym} and \ref{asm:balanced_cov} or the covariate diversity condition are standard (at least in the experimental settings). They hold for many context distributions including multivariate Gaussian distribution, uniform distribution on the sphere, and arbitrary independent distribution
for each arm \cite{oh2020sparsity}. For example, the covariate diversity condition holds in the experimental settings of \citet{bastani2020online} and \citet{wang2018minimax}. 

Regarding the regret dependence of $K$, we have at least linear scaling with $K$.  The constant $C_{\textnormal{b}}$ does not scale with $K$ when the context distribution is a multivariate Gaussian distribution or a uniform distribution on a unit sphere (see Proposition~1 of \citet{oh2020sparsity}). However, for general distribution, $C_{\textnormal{b}}$ can scale exponentially with $K$. We conjecture that we can improve this dependency: numerical results show that the dependence on $K$ is mild (See Appendix~\ref{sec:appendix_exp}).

\section{On the Benefit of the Two Step Thresholding Procedure}
\label{app:benefit_thresholding}

In our choice of the thresholding parameter ($ 4\lambda_t$ in the first step and $ 4\lambda_t \sqrt{|\hat{S}_0^{(t)}|}$ in the second step), we aim at a partial recovery of the support so that the trade-off between the duration of the phase with linearly growing regret and the support recovery accuracy is optimized in the design. 
Using two-step thresholding, we achieve better regret guarantees than single-step thresholding. This improvement is due to the fact that with two-step thresholding, the estimated support of $\theta$ is improved (with two-step thresholding, we have $\mathcal{O}(\sqrt{s_0})$ false positives on the estimated support, whereas with single thresholding, there are $ \mathcal{O}(s_0)$). While this difference in results does not contribute to changing the order of the regret, it does contribute to improving the coefficients on $\log T$ and $\sqrt{T}$ terms in regret.

\section{Additional Theorems}
\label{sec:additional_theorems}

Before presenting the additional theorems, we introduce the following assumption (which is a slightly modified version of Assumption~\ref{asm:cov_div_klarge}).

\begin{assumption}[Sparse positive definiteness, $K=2$]
\label{asm:cov_div}
Let $K=2$. Define $\Sigma_B \coloneqq \frac{1}{2} \sum_{k=1}^2 \EXP_{\mathcal{A}\sim p_A} \left[A_{k}(B) A_{k}(B)^\top\right]$, for any $B \subset [d]$, where $ A_{k}(B)$ is a $|B|$-dimensional vector, which is extracted from the elements of $A_{k}$ with indices in $B$. 
There exists a positive constant $\alpha>0$ such that $\forall B \subset [d]$,
 \begin{align*}
 \bigg( |B| \le s_0 + (2 \nu \sqrt{s_0})/\phi_0^2 \; \text{and} \; S(\theta) \subset B\bigg) \Longrightarrow \bigg( \min_{v \in \mathbb{R}^{|B|} : \|v\|_2 = 1} v^\top \Sigma_B v  \ge \alpha \bigg).
 \end{align*}
 The parameters $\phi_0, \nu$ are those of Assumptions \ref{asm:comp_cond}, \ref{asm:relax_sym}.
 \end{assumption}
We redefine the parameters $\tau = \left\lfloor\frac{2 \log (2 d^2)}{C_0^2} (\log s_0) (\log \log d) \right\rfloor$, where $C_0 =\min \left\{\frac{1}{2}, \frac{\phi_0^2}{256 s_0 s_A^2 \nu }\right\}$. 

The following theorem provides the regret guarantees when $K=2$, without Balanced covariance (Assumption~\ref{asm:balanced_cov}), and with the margin condition.
\begin{theorem}[with margin, without balanced covariance]\label{thm:regret_ub_k2}
Assume that Assumptions \ref{asm:sparsity_param_klarge}--\ref{asm:relax_sym}, \ref{asm:cov_div}, and \ref{asm:margin} hold and $K=2$.\\
(i) (TH Lasso Bandit with parameter-dependent input) There are universal positive constants $c_1, c_2, c_3$ depending on $ \sigma, s_A, s_1, s_2, \phi_0, \nu,  \alpha, C_{\textnormal{m}}$, such that if we set $\lambda_0=c_1$, then for all $d \ge c_2$, for all $T \ge 2$:
\begin{align*}
& R(T) \le c_3\left( \tau + s_0 (\log s_0)^{\frac{3}{2}}\log T + s_0^2\right).
\end{align*}
(ii) (TH Lasso Bandit with parameter-free input) There are universal positive constants $c_4, c_5$ depending on $ \sigma, s_A, s_1, s_2, \phi_0, \nu,  \alpha, C_{\textnormal{m}}$, such that if we set $ \lambda_0 = {1}/{(\log \log d)^{\frac{1}{4}}}$ in TH Lasso Bandit, then for all $d \ge c_4$, for all $T \ge 2$,
\begin{align*}
     R(T)\le c_5\left( \tau + s_0 (\log s_0)^{\frac{3}{2}}\log T + s_0^2\right).
\end{align*}
The precise definitions of $c_1$-$c_5$ are given in Appendix~\ref{subsec:proof_regret_ub_k2}.
\end{theorem}
We provide the proof of Theorem~\ref{thm:regret_ub_k2} in Appendix~\ref{subsec:proof_regret_ub_k2}.

Next, the following theorem provides the regret guarantees when $K=2$, without Balanced covariance (Assumption~\ref{asm:balanced_cov}), and without the margin condition.
\begin{theorem}[without margin, without balanced covariance]\label{thm:regret_ub_wo_margin_k2}
Assume that Assumptions \ref{asm:sparsity_param_klarge}-- \ref{asm:relax_sym}, and \ref{asm:cov_div} hold and $K=2$.\\
(i) (TH Lasso Bandit with parameter-dependent input) There are universal positive constants $c_1, c_2, c_3$ depending on $ \sigma, s_A, s_1, s_2, \phi_0, \nu,  \alpha$ such that if we set
$\lambda_0 = c_1 $, then for all $d \ge c_2$. for all $T \ge 2$:
\begin{align*}
& R(T) \le c_3\left(  \tau +  (\log s_0) \sqrt{s_0 T} + s_0^2\right).
\end{align*}
(ii) (TH Lasso Bandit with parameter-free input) There are universal positive constants $c_4, c_5$ depending on $ \sigma, s_A, s_1, s_2, \phi_0, \nu,  \alpha$ such that if we set $ \lambda_0 = {1}/{(\log \log d)^{\frac{1}{4}}}$ in TH Lasso Bandit, then for all $d \ge c_4$, for all $T \ge 2$,
\begin{align*}
     R(T)\le c_5\left( \tau +  (\log s_0) \sqrt{s_0 T} + s_0^2\right).
\end{align*}
The precise definitions of $c_1$-$c_5$ are given in Appendix~\ref{subsec:proof_ub_wo_margin_wo_balance}.
\end{theorem}
We present the proof of Theorem~\ref{thm:regret_ub_wo_margin_k2} in Appendix~\ref{subsec:proof_ub_wo_margin_wo_balance}.
Furthermore, Lemmas associated with Theorems~\ref{thm:regret_ub_k2} and \ref{thm:regret_ub_wo_margin_k2} and their proofs are presented in Appendix~\ref{sec:proof_lemmas_k2}.

\section{Proof of Theorems}
\label{sec:Proof_theorems}

\subsection{Proof of Theorem~\ref{thm:regret_ub_klarge} (with margin)}
\label{subsec:proof_regret_ub_klarge}

First, we determine the constants $c_1$, $c_2$ as follows.
Set $\lambda_0 = 4 \sigma s_A \sqrt{c} $ with constant $c>0$ (independent of $d$, $T$, and $s_0$) such that $4 \left( \frac{4 \nu C_{\textnormal{b}}s_0}{\phi_0^2} + \sqrt{\left(1 + \frac{4\nu C_{\textnormal{b}}}{\phi_0^2}\right)s_0}\right) \underbrace{\left(4 \sigma s_A \sqrt{c} \sqrt{\frac{2 \log \tau \log d}{\tau}} \right)}_{\lambda_\tau} \le \theta_{\min}$. Note that such a constant $c$ exists as 
\begin{align*}
    \lambda_\tau & = 4 \sigma s_A \sqrt{c} \sqrt{\frac{2 \log \tau \log d}{\tau}} 
 \\
 & = 4 \sigma s_A \sqrt{c} \sqrt{\frac{2 \log (\Theta(s_0^2 (\log s_0) (\log \log d) (\log d)))\log d}{\Theta(s_0^2 (\log s_0) (\log \log d) (\log d))}} 
 \\
 & =  4 \sigma s_A \sqrt{c} \sqrt{\frac{2 \log (\Theta(s_0^2 (\log s_0) (\log \log d) (\log d)))}{\Theta(s_0^2 (\log s_0) (\log \log d) )}} 
 \\
 & \Longrightarrow   \lambda_\tau = \mathcal{O}\left(\frac{1}{s_0}\sqrt{\frac{1}{\log \log d} + \frac{1}{\log s_0}} \right) \qquad \text{and}
  \\
 \theta_{\min} & \ge s_2/s_0 \qquad(\textnormal{from Assumption~\ref{asm:sparsity_param_klarge}}).
\end{align*}
We can take $c_1 = 4 \sigma s_A \sqrt{c}$. Assume that $\tau$ (increasing function of $d$) satisfies $ \tau \ge \exp(4/c).$ This facilitates a constant lower bound on $d$, hence $ c_2$ is determined. 

We upper bound the instantaneous regret in round $t \ge 1$. We have:
\begin{align}
    \EXP[\max_{A \in \mathcal{A}_t} \langle A - A_t, \theta\rangle] & = \EXP[\max_{A \in \mathcal{A}_t} \langle A - A_t, \theta\rangle] \nonumber
    \\
    & \le \EXP[| \max_{A \in \mathcal{A}_t} \langle A, \theta\rangle| ]  +  \EXP[| \langle A_t, \theta \rangle| ] \nonumber
    \\
    & \le s_A s_1  + s_A s_1  \nonumber
    \\
    & = 2 s_A s_1, \label{eq:CS_inst_reg_klarge}
\end{align}
where the second inequality stems from H\"{o}lder's inequality. We deduce the following upper bound on the expected regret up to round $T$:

\begin{align*}
    R(T) & =  \EXP \left[ \sum_{t =1}^T \max_{A \in \mathcal{A}_t} \langle A - A_t, \theta\rangle\right]
    \\
    & \stackrel{(a)}{\le} 2s_A s_1  \tau + \sum_{t = \tau + 1}^T \EXP \left[  \max_{A \in \mathcal{A}_t} \langle A - A_t, \theta\rangle\right]
    \\
    & \stackrel{(b)}{\le}  2s_A s_1 \tau + \sum_{t = \tau + 1}^T\bigg(\frac{1408 \sigma^2 s_A^4 C_{\text{m}} (K-1) h_0^3 \nu^2 C_{\textnormal{b}}^2 \left(s_0 + \frac{4 \nu C_{\textnormal{b}}\sqrt{s_0}}{\phi_0^2}\right)}{\alpha^2} \frac{1}{t - 1} 
    \\
    &\qquad\qquad +  2 (K-1)  s_A s_1  \left(\Pr(\mathcal{E}_t^c) + \Pr\left(\left(\mathcal{G}_{t}^{\frac{\alpha}{4 \nu C_{\textnormal{b}}}}\right)^c \middle| \mathcal{E}_t\right) \right) \bigg)
    \\
    & \stackrel{(c)}{\le}  2s_A s_1 \tau + \sum_{t = \tau + 1}^T\bigg(\frac{1408 \sigma^2 s_A^4 C_{\text{m}} (K-1) h_0^3 \nu^2 C_{\textnormal{b}}^2 \left(s_0 + \frac{4 \nu  C_{\textnormal{b}}\sqrt{s_0}}{\phi_0^2}\right)}{\alpha^2} \frac{1}{t - 1}
    \\
    & \qquad\qquad + 2 (K-1)  s_A s_1  \bigg(2 \exp\left( -\frac{t \lambda_t^2}{32 \sigma^2 s_A^2} + \log d\right) + \exp\left( - \frac{t C_0^2}{2}\right) 
    \\
    &\qquad\qquad\qquad + \exp\left(\log\left(s_0 + \frac{4 \nu C_{\textnormal{b}}\sqrt{s_0}}{\phi_0^2}\right) - \frac{t \alpha}{20 s_A^2  \nu C_{\textnormal{b}}\left(s_0 + (4 \nu C_{\textnormal{b}} \sqrt{s_0})/\phi_0^2\right)}\right) \bigg) \bigg),
\end{align*}
where for $ (a)$, we used equation \eqref{eq:CS_inst_reg_klarge} for $ 1 \le t \le \tau$; for $(b)$, we used Lemma~\ref{lm:inst_regret_bound_klarge}; for $(c)$, we used Lemma~\ref{lm:support_recovery_klarge} (for $\mathcal{E}_t$) and  Lemma~\ref{lm:lowerbound_eig_S_klarge} (for $\mathcal{G}_t^{\frac{\alpha}{4 \nu C_{\textnormal{b}}}}$). 
Now we have:
\begin{align*}
    \sum_{t = \tau + 1}^T  & \frac{1408 \sigma^2 s_A^4 C_{\text{m}} (K-1) h_0^3 \nu^2 C_{\textnormal{b}}^2 \left(s_0 + \frac{4 \nu C_{\textnormal{b}}\sqrt{s_0}}{\phi_0^2}\right)}{\alpha^2} \frac{1}{t - 1} \\
    &\qquad\qquad\qquad = \sum_{t = \tau }^{T - 1}\frac{1408 \sigma^2 s_A^4 C_{\text{m}} (K-1) h_0^3 \nu^2 C_{\textnormal{b}}^2 \left(s_0 + \frac{4 \nu C_{\textnormal{b}}\sqrt{s_0}}{\phi_0^2}\right)}{\alpha^2} \frac{1}{t}
    \\
    &\qquad\qquad\qquad  \le \frac{1408 \sigma^2 s_A^4 C_{\text{m}} (K-1) h_0^3 \nu^2 C_{\textnormal{b}}^2 \left(s_0 + \frac{4 \nu C_{\textnormal{b}}\sqrt{s_0}}{\phi_0^2}\right)}{\alpha^2}(1 + \int_{1}^T \frac{1}{t} dt )
    \\
    &\qquad\qquad\qquad  = \frac{1408 \sigma^2 s_A^4 C_{\text{m}} (K-1) h_0^3 \nu^2 C_{\textnormal{b}}^2 \left(s_0 + \frac{4 \nu C_{\textnormal{b}}\sqrt{s_0}}{\phi_0^2}\right)}{\alpha^2}(1 + \log T ),
\end{align*}
and
\begin{align*}
    \sum_{t = \tau + 1}^T \exp\left( -\frac{t \lambda_t^2}{32 \sigma^2 s_A^2} + \log d\right) & = \sum_{t = \tau + 1}^T \exp\left( - c\log t \log d + \log d\right)
    \\
    & \stackrel{(a)}{\le} \sum_{t = \tau + 1}^T \exp \left( - \frac{c \log d \log t}{2}\right)
    \\
    & \stackrel{(b)}{\le} \sum_{t = \tau + 1}^T \exp \left( - 2\log t\right)
    \\
    & = \sum_{t = \tau + 1}^T  \frac{1}{t^2}
    \\
    & \le \sum_{t = 1}^\infty  \frac{1}{t^2}
    \\
    & = \frac{\pi^2}{6},
\end{align*}
where for $(a)$ and $(b)$, we used the assumption $ \tau \ge \exp(4/c)$.
 In addition, 
\begin{align*}
    & \sum_{t = \tau + 1}^T \exp\left( - \frac{t C_0^2}{2}\right)  + \exp\left(\log\left(s_0 + \frac{4 \nu C_{\textnormal{b}}\sqrt{s_0}}{\phi_0^2}\right) - \frac{t \alpha}{20 s_A^2 \nu C_{\textnormal{b}} \left(s_0 + (4 \nu C_{\textnormal{b}} \sqrt{s_0})/\phi_0^2\right)}\right) 
    \\
    & \le \int_{0}^\infty \left( \exp\left( - \frac{t C_0^2}{2}\right) + \left(s_0 + \frac{4 \nu C_{\textnormal{b}} \sqrt{s_0}}{\phi_0^2}\right) \exp\left(- \frac{t \alpha}{20 s_A^2 \nu C_{\textnormal{b}} \left(s_0 + (4 \nu C_{\textnormal{b}} \sqrt{s_0})/\phi_0^2\right)}\right) \right) dt 
    \\
    & = \frac{2}{C_0^2} + \left(s_0 + \frac{4 \nu C_{\textnormal{b}} \sqrt{s_0}}{\phi_0^2}\right)^2 \frac{20 s_A^2 \nu C_{\textnormal{b}}}{\alpha}.
\end{align*}
In summary, we obtain:
\begin{align*}
    R(T) &  \le 2s_A s_1  \tau + \frac{1408 \sigma^2 s_A^4 C_{\text{m}} (K-1) h_0^3 \nu^2 C_{\textnormal{b}}^2 \left(s_0 + \frac{4 \nu C_{\textnormal{b}}\sqrt{s_0}}{\phi_0^2}\right)}{\alpha^2}(1 + \log T )
    \\
    & \ \ \ + 2 (K-1)  s_A s_1 \left( \frac{\pi^2}{3} + \frac{2}{C_0^2} + \left(s_0 + \frac{4 \nu C_{\textnormal{b}} \sqrt{s_0}}{\phi_0^2}\right)^2 \frac{20 s_A^2 \nu C_{\textnormal{b}}}{\alpha} \right).
\end{align*}
Looking at the scaling with respect to $d$, $T$, and $s_0$, one can determine $c_3$ (note that $1/C_0^2 = \mathcal{O}(s_0^2)$ and $ h_0^3 = \mathcal{O}((\log(s_0))^{\frac{3}{2}})$). This concludes the proof of the first part of the theorem.

Regarding the second part of the theorem, we impose the following condition on $d$:
\begin{itemize}
    \item[(i)] $ \log \log d \ge 4096 \sigma^4 s_A^4 $
    \item[(ii)] $
4 \left( \frac{4 \nu C_{\textnormal{b}}s_0}{\phi_0^2} + \sqrt{ \left(1 + \frac{4 \nu C_{\textnormal{b}}}{\phi_0^2} \right) s_0}\right) \lambda_\tau \le \theta_{\min},
$
\item[(iii)] $ d \ge 100$.
\end{itemize}
When $ d \ge c_4$ with some constant $c_4$, it should be noted that the condition (ii) can hold, as
\begin{align}
    \sqrt{\frac{2 \log \tau \log d}{\tau}}  = \mathcal{O}\left(\frac{1}{s_0}\sqrt{\frac{1}{\log \log d} + \frac{1}{\log s_0}} \right)
\end{align}
and 
\begin{align}
    \lambda_0 & = 1/(\log \log d)^{\frac{1}{4}}
    \\
    & = o(1) \quad (\text{as } d \to \infty).
\end{align}

Then, we have the following computations:
\begin{align*}
    \sum_{t = \tau + 1}^T \exp\left( -\frac{t \lambda_t^2}{32 \sigma^2 s_A^2} + \log d\right)
    & = \sum_{t = \tau + 1}^T \exp\left( -\left( \frac{ \log t}{16 \sigma^2 s_A^2 (\log \log d)^{\frac{1}{2}}} - 1\right)\log d\right)
    \\
    & \stackrel{(a)}{\le}  \sum_{t = \tau + 1}^T \exp\left( - \frac{ \log t \log d}{32 \sigma^2 s_A^2 (\log \log d)^{\frac{1}{2}} } \right)
    \\
    & \stackrel{(b)}{\le} \sum_{t = \tau + 1}^T \exp\left( - \frac{ \log t \log \tau}{32 \sigma^2 s_A^2 (\log \log d)^{\frac{1}{2}} } \right)
    \\
    & \stackrel{(c)}{\le}  \sum_{t = \tau + 1}^T \exp\left( - 2 \log t  \right)
    \\
    & = \sum_{t = \tau + 1}^T \frac{1}{t^2}
    \\
    & \le \sum_{t = 1}^\infty \frac{1}{t^2}
    \\
    & = \frac{\pi^2}{6},
\end{align*}
where for $(a)$, we used the fact that $ \log \tau \ge 64 \sigma^2 s_A^2  (\log \log d)^{\frac{1}{2}}$ from (i); for $(b)$, we used the fact that $\log \tau \le \log d$ from $d \ge 100$; for $(c)$, we used again $ \log \tau \ge 64 \sigma^2 s_A^2  (\log \log d)^{\frac{1}{2}}$. 
Therefore, similarly, we get the regret bound
\begin{align*}
    R(T) &  \le 2s_A s_1  \tau + \frac{1408 \sigma^2 s_A^4 C_{\text{m}} (K-1) h_0^3 \nu^2 C_{\textnormal{b}}^2 \left(s_0 + \frac{4 \nu C_{\textnormal{b}}\sqrt{s_0}}{\phi_0^2}\right)}{\alpha^2}(1 + \log T )
    \\
    & \ \ \ + 2 (K-1)  s_A s_1  \left( \frac{\pi^2}{3} + \frac{2}{C_0^2} + \left(s_0 + \frac{4 \nu C_{\textnormal{b}} \sqrt{s_0}}{\phi_0^2}\right)^2 \frac{20 s_A^2 \nu C_{\textnormal{b}}}{\alpha} \right),
\end{align*}
and we can find the constant $c_5$. This concludes the proof.

\ep 

\subsection{Proof of Theorem~\ref{thm:regret_ub_wo_margin_klarge} (without margin)}
\label{subsec:regret_ub_wo_margin_klarge}
First, we determine the constants $c_1$, $c_2$ similarly to those of Theorem~\ref{thm:regret_ub_klarge}.

Using Lemma~\ref{lm:inst_regret_bound_wo_margin_klarge}, we proceed as in the proof of Theorem~\ref{thm:regret_ub_klarge}, and deduce that:
\begin{align*}
    & R(T) =  \EXP \left[ \sum_{t =1}^T \max_{A \in \mathcal{A}_t} \langle A - A_t, \theta\rangle\right]
    \\
    & \le 2s_A s_1 \tau + \sum_{t = \tau + 1}^T \EXP \left[  \max_{A \in \mathcal{A}_t} \langle A - A_t, \theta\rangle\right]
    \\
    & \le  2s_A s_1  \tau\\
    & + \sum_{t = \tau + 1}^T\left(\frac{36 \sigma s_A  (K-1) h_0^2 \nu C_{\textnormal{b}}}{\alpha}\sqrt{\frac{2 \left(s_0 + \frac{4 \nu C_{\textnormal{b}}\sqrt{s_0} }{\phi_0^2}\right)}{t-1}} + 2  (K-1) s_A s_1  \left(\Pr(\mathcal{E}_t^c) + \Pr\left(\left(\mathcal{G}_{t}^{\frac{\alpha}{4\nu C_{\textnormal{b}}}}\right)^c \middle| \mathcal{E}_t\right) \right)\right)
\end{align*}
We also show that:
\begin{align*}
    \sum_{t=\tau+1}^T  2 (K - 1)  s_A s_1  & \left(\Pr(\mathcal{E}_t^c) + \Pr\left(\left(\mathcal{G}_{t}^{\frac{\alpha}{4\nu C_{\textnormal{b}}}}\right)^c \middle| \mathcal{E}_t\right) \right) \\
    &\le 2 (K-1)  s_A s_1 \left( \frac{\pi^2}{3} + \frac{2}{C_0^2} + \left(s_0 + \frac{4\nu C_{\textnormal{b}}\sqrt{s_0}}{\phi_0^2}\right)^2 \frac{20 s_A^2 \nu C_{\textnormal{b}}}{\alpha} \right).
\end{align*}
Now we have:
\begin{align*}
    \sum_{t = \tau + 1}^T & \frac{36 \sigma s_A  (K-1) h_0^2 \nu C_{\textnormal{b}}}{\alpha}\sqrt{\frac{2 \left(s_0 + \frac{4 \nu C_{\textnormal{b}}\sqrt{s_0} }{\phi_0^2}\right)}{t-1}}  = \sum_{t = \tau }^{T - 1}\frac{36 \sigma s_A  (K-1) h_0^2 \nu C_{\textnormal{b}}}{\alpha}\sqrt{\frac{2 \left(s_0 + \frac{4 \nu C_{\textnormal{b}}\sqrt{s_0} }{\phi_0^2}\right)}{t}}
    \\
    & \le \frac{36 \sigma s_A  (K-1) h_0^2 \nu C_{\textnormal{b}}\sqrt{2 \left(s_0 + \frac{4 \nu C_{\textnormal{b}}\sqrt{s_0} }{\phi_0^2}\right)}}{\alpha}(1 + \int_{1}^T \sqrt{\frac{1}{t}} dt )
    \\
    & \le \frac{36 \sigma s_A  (K-1) h_0^2 \nu C_{\textnormal{b}}\sqrt{8 \left(s_0 + \frac{4 \nu C_{\textnormal{b}}\sqrt{s_0} }{\phi_0^2}\right)}}{\alpha}\sqrt{T},
\end{align*}
In summary, we get:
\begin{align*}
    R(T) & \le 2 s_A s_1 \tau + \frac{36 \sigma s_A  (K-1) h_0^2 \nu C_{\textnormal{b}}\sqrt{8 \left(s_0 + \frac{4 \nu C_{\textnormal{b}}\sqrt{s_0} }{\phi_0^2}\right)}}{\alpha}\sqrt{T} 
    \\
    & \ \ \ + 2 (K-1)  s_A s_1 \left( \frac{\pi^2}{3} + \frac{2}{C_0^2} + \left(s_0 + \frac{4 \nu C_{\textnormal{b}} \sqrt{s_0}}{\phi_0^2}\right)^2 \frac{20 s_A^2 \nu C_{\textnormal{b}}}{\alpha} \right).
\end{align*}
Looking at the scaling with respect to $d$, $T$, and $s_0$, one can determine $c_3$ (note that $1/C_0^2 = \mathcal{O}(s_0^2)$ and $ h_0^2 = \mathcal{O}(\log(s_0))$).  This concludes the proof of the first part of the theorem.

Regarding the second part of the theorem, we can determine the constant $c_4$ similarly as in Theorem~\ref{thm:regret_ub_klarge}. 
Then, we proceed as in the proof of Theorem~\ref{thm:regret_ub_klarge} and get the regret bound:
\begin{align*}
    R(T) & \le 2 s_A s_1 \tau + \frac{36 \sigma s_A  (K-1) h_0^2 \nu C_{\textnormal{b}}\sqrt{8 \left(s_0 + \frac{4 \nu C_{\textnormal{b}}\sqrt{s_0} }{\phi_0^2}\right)}}{\alpha}\sqrt{T} 
    \\
    & \ \ \ + 2 (K-1)  s_A s_1 \left( \frac{\pi^2}{3} + \frac{2}{C_0^2} + \left(s_0 + \frac{4 \nu C_{\textnormal{b}} \sqrt{s_0}}{\phi_0^2}\right)^2 \frac{20 s_A^2 \nu C_{\textnormal{b}}}{\alpha} \right).
\end{align*}
We can find the constant $c_5$ from this upper bound.
This concludes the proof. \ep

\subsection{Proof of Theorem~\ref{thm:regret_ub_k2} (with margin, without balanced covariance)}
\label{subsec:proof_regret_ub_k2}
This proof follows that of Theorem~\ref{thm:regret_ub_klarge} to some extent.
First, we determine the constants $c_1$, $c_2$ as follows.
Set $\lambda_0 = 4 \sigma s_A \sqrt{c} $ with constant $c>0$ (independent of $d$, $T$, and $s_0$)  such that $ 4 \left( \frac{2 \nu s_0}{\phi_0^2} + \sqrt{\left(1 + \frac{2\nu }{\phi_0^2}\right)s_0}\right) \underbrace{\left( 4 \sigma s_A \sqrt{c} \sqrt{\frac{2 \log \tau \log d}{\tau}} \right)}_{\lambda_\tau}\le \theta_{\min}$. Note that such a constant $c$ exists as 
\begin{align*}
    \lambda_\tau & = 4 \sigma s_A \sqrt{c} \sqrt{\frac{2 \log \tau \log d}{\tau}} 
 \\
 & = 4 \sigma s_A \sqrt{c} \sqrt{\frac{2 \log (\Theta(s_0^2 (\log s_0) (\log \log d) (\log d)))\log d}{\Theta(s_0^2 (\log s_0) (\log \log d) (\log d))}} 
 \\
 & =  4 \sigma s_A \sqrt{c} \sqrt{\frac{2 \log (\Theta(s_0^2 (\log s_0) (\log \log d) (\log d)))}{\Theta(s_0^2 (\log s_0) (\log \log d) )}} 
 \\
 & \Longrightarrow   \lambda_\tau = \mathcal{O}\left(\frac{1}{s_0}\sqrt{\frac{1}{\log \log d} + \frac{1}{\log s_0}} \right) \qquad \text{and}
  \\
 \theta_{\min} & \ge s_2/s_0 \qquad(\textnormal{from Assumption~\ref{asm:sparsity_param_klarge}}).
\end{align*}
We can take $c_1 = 4 \sigma s_A \sqrt{c}$. Assume that $\tau$ (increasing function of $d$) satisfies $ \tau \ge \exp(4/c).$ This facilitates a constant lower bound on $d$, hence $ c_2$ is determined. 

We deduce the following upper bound on the expected regret up to round $T$:
\begin{align*}
    R(T) & =  \EXP \left[ \sum_{t =1}^T \max_{A \in \mathcal{A}_t} \langle A - A_t, \theta\rangle\right]
    \\
    & \stackrel{(a)}{\le} 2s_A s_1  \tau + \sum_{t = \tau + 1}^T \EXP \left[  \max_{A \in \mathcal{A}_t} \langle A - A_t, \theta\rangle\right]
    \\
    & \stackrel{(b)}{\le}  2s_A s_1  \tau + \sum_{t = \tau + 1}^T\left(\frac{352 \sigma^2 s_A^4 C_{\text{m}}  h_0^3\nu^2\left(s_0 + \frac{2 \nu \sqrt{s_0}}{\phi_0^2}\right)}{\alpha^2} \frac{1}{t-1} + 2   s_A s_1 \left(\Pr(\mathcal{E}_t^c) + \Pr\left(\left(\mathcal{G}_{t}^{\frac{\alpha}{2\nu}}\right)^c \middle| \mathcal{E}_t\right) \right) \right)
    \\
    & \stackrel{(c)}{\le}  2s_A s_1 \tau + \sum_{t = \tau + 1}^T\Bigg(\frac{352 \sigma^2 s_A^4 C_{\text{m}}  h_0^3 \nu^2 \left(s_0 + \frac{2 \nu \sqrt{s_0}}{\phi_0^2}\right)}{\alpha^2} \frac{1}{t-1}
    \\
    & \ \ \ + 2  s_A s_1  \Bigg(2 \exp\left( -\frac{t \lambda_t^2}{32 \sigma^2 s_A^2} + \log d\right) + \exp\left( - \frac{t C_0^2}{2}\right) 
    \\
    & \ \ \ + \exp\left(\log\left(s_0 + \frac{2 \nu \sqrt{s_0}}{\phi_0^2}\right) - \frac{t \alpha}{10 s_A^2 \nu  \left(s_0 + (2 \nu \sqrt{s_0})/\phi_0^2\right)}\right) \Bigg) \Bigg),
\end{align*}
where for $ (a)$, we used equation \eqref{eq:CS_inst_reg_klarge} for $ 1 \le t \le \tau$; for $(b)$, we used Lemma~\ref{lm:inst_regret_bound_k2}; for $(c)$, we used Lemma~\ref{lm:support_recovery_k2} (for $\mathcal{E}_t$) and  Lemma~\ref{lm:lowerbound_eig_S_k2} (for $\mathcal{G}_t^{\frac{\alpha}{2\nu}}$). 
Now we have:
\begin{align*}
    \sum_{t = \tau + 1}^T\frac{352 \sigma^2 s_A^4 C_{\text{m}} h_0^3 \nu^2\left(s_0 + \frac{2 \nu \sqrt{s_0}}{\phi_0^2}\right)}{\alpha^2} \frac{1}{t-1} & = \sum_{t = \tau }^{T - 1}\frac{352 \sigma^2 s_A^4 C_{\text{m}}  h_0^3\left(s_0 + \frac{2 \nu \sqrt{s_0}}{\phi_0^2}\right)}{\alpha^2} \frac{1}{t}
    \\
    & \le \frac{352 \sigma^2 s_A^4 C_{\text{m}}  h_0^3 \nu^2\left(s_0 + \frac{2 \nu \sqrt{s_0}}{\phi_0^2}\right)}{\alpha^2}(1 + \int_{1}^T \frac{1}{t} dt )
    \\
    & = \frac{352 \sigma^2 s_A^4 C_{\text{m}}  h_0^3 \nu^2\left(s_0 + \frac{2 \nu \sqrt{s_0}}{\phi_0^2}\right)}{\alpha^2}(1 + \log T ),
\end{align*}
and
\begin{align*}
    \sum_{t = \tau + 1}^T \exp\left( -\frac{t \lambda_t^2}{32 \sigma^2 s_A^2} + \log d\right) & = \sum_{t = \tau + 1}^T \exp\left( -c\log t \log d + \log d\right)
    \\
    & \stackrel{(a)}{\le} \sum_{t = \tau + 1}^T \exp \left( - \frac{c \log d \log t}{2}\right)
    \\
    & \stackrel{(b)}{\le} \sum_{t = \tau + 1}^T \exp \left( - 2\log t\right)
    \\
    & = \sum_{t = \tau + 1}^T  \frac{1}{t^2}
    \\
    & \le \sum_{t = 1}^\infty  \frac{1}{t^2}
    \\
    & = \frac{\pi^2}{6},
\end{align*}
where for $(a)$ and $(b)$, we used the assumption $ \tau \ge \exp(4/c)$.
In addition, 
\begin{align*}
    & \sum_{t = \tau + 1}^T\left( \exp\left( - \frac{t C_0^2}{2}\right) + \exp\left(\log\left(s_0 + \frac{2 \nu \sqrt{s_0}}{\phi_0^2}\right) - \frac{t \alpha}{10 s_A^2 \nu  \left(s_0 + (2 \nu \sqrt{s_0})/\phi_0^2\right)} \right) \right)
    \\
    & \le \int_{0}^\infty \left( \exp\left( - \frac{t C_0^2}{2}\right) + \left(s_0 + \frac{2 \nu \sqrt{s_0}}{\phi_0^2}\right) \exp\left(- \frac{t \alpha}{10 s_A^2 \nu  \left(s_0 + (2 \nu \sqrt{s_0})/\phi_0^2\right)}\right) \right) dt 
    \\
    & = \frac{2}{C_0^2} + \left(s_0 + \frac{2 \nu \sqrt{s_0}}{\phi_0^2}\right)^2 \frac{10 s_A^2 \nu }{\alpha}.
\end{align*}
In summary, we obtain that:
\begin{align*}
    R(T) &  \le 2s_A s_1  \tau + \frac{352 \sigma^2 s_A^4 C_{\text{m}}  h_0^3\nu^2\left(s_0 + \frac{2 \nu \sqrt{s_0}}{\phi_0^2}\right)}{\alpha^2} (\log T + 1) 
    \\
    & \ \ \ + 2  s_A s_1 \left( \frac{\pi^2}{3} + \frac{2}{C_0^2} + \left(s_0 + \frac{2 \nu \sqrt{s_0}}{\phi_0^2}\right)^2 \frac{10 s_A^2 \nu }{\alpha} \right).
\end{align*}
Looking at the scaling with respect to $d$, $T$, and $s_0$, one can determine $c_3$. This concludes the first part of the theorem.

Regarding the second part of the theorem, we impose the following condition on $d$:
\begin{itemize}
    \item[(i)] $ \log \log d \ge 4096 \sigma^4 s_A^4 $
    \item[(ii)] $
4 \left( \frac{2 \nu s_0}{\phi_0^2} + \sqrt{ \left(1 + \frac{2 \nu }{\phi_0^2} \right) s_0}\right) \lambda_\tau \le \theta_{\min},
$
\item[(iii)] $ d \ge 100$.
\end{itemize}
With some constant $c_4$, when $ d \ge c_4$, it should be noted that the condition (ii) holds, as
\begin{align*}
     \sqrt{\frac{2 \log \tau \log d}{\tau}}=  \mathcal{O}\left(\frac{1}{s_0}\sqrt{\frac{1}{\log \log d} + \frac{1}{\log s_0}} \right)
\end{align*}
and 
\begin{align*}
    \lambda_0 & = 1/(\log \log d)^{\frac{1}{4}}
    \\
    & = o(1) \quad (\text{as } d \to \infty).
\end{align*}

We have,
\begin{align*}
    \sum_{t = \tau + 1}^T \exp\left( -\frac{t \lambda_t^2}{32 \sigma^2 s_A^2} + \log d\right)
    & = \sum_{t = \tau + 1}^T \exp\left( -\left( \frac{ \log t}{16 \sigma^2 s_A^2 (\log \log d)^{\frac{1}{2}}} - 1\right)\log d\right)
    \\
    & \stackrel{(a)}{\le}  \sum_{t = \tau + 1}^T \exp\left( - \frac{ \log t \log d}{32 \sigma^2 s_A^2 (\log \log d)^{\frac{1}{2}} } \right)
    \\
    & \stackrel{(b)}{\le} \sum_{t = \tau + 1}^T \exp\left( - \frac{ \log t \log \tau}{32 \sigma^2 s_A^2 (\log \log d)^{\frac{1}{2}} } \right)
    \\
    & \stackrel{(c)}{\le}  \sum_{t = \tau + 1}^T \exp\left( - 2 \log t  \right)
    \\
    & = \sum_{t = \tau + 1}^T \frac{1}{t^2}
    \\
    & \le \sum_{t = 1}^\infty \frac{1}{t^2}
    \\
    & = \frac{\pi^2}{6},
\end{align*}
where for $(a)$, we used the fact that $ \log \tau \ge 64 \sigma^2 s_A^2  (\log \log d)^{\frac{1}{2}}$; for $(b)$, we used the fact that $\log \tau \le \log d$ from $d \ge 100$; for $(c)$, we used again $ \log \tau \ge 64 \sigma^2 s_A^2  (\log \log d)^{\frac{1}{2}}$.  Therefore, a similar regret upper bound can be obtained in this case:
\begin{align*}
    R(T) &  \le 2s_A s_1 \tau + \frac{352 \sigma^2 s_A^4 C_{\text{m}}  h_0^3\nu^2\left(s_0 + \frac{2 \nu \sqrt{s_0}}{\phi_0^2}\right)}{\alpha^2} (\log T + 1) 
    \\
    & \ \ \ + 2  s_A s_1 \left( \frac{\pi^2}{3} + \frac{2}{C_0^2} + \left(s_0 + \frac{2 \nu \sqrt{s_0}}{\phi_0^2}\right)^2 \frac{10 s_A^2 \nu }{\alpha} \right)
\end{align*}
and we can find the constant $c_5$. This concludes the proof.

\subsection{Proof of Theorem~\ref{thm:regret_ub_wo_margin_k2}
 (without margin, without balanced covariance)}
 \label{subsec:proof_ub_wo_margin_wo_balance}
 
 This proof follows the proof of Theorem~\ref{thm:regret_ub_wo_margin_klarge} mostly.
 First, we determine the constants $c_1$, $c_2$ similarly to those of Theorem~\ref{thm:regret_ub_k2}.

Using Lemma~\ref{lm:inst_regret_bound_wo_margin_k2}, we proceed as in the proof of Theorem~\ref{thm:regret_ub_wo_margin_klarge}, and deduce that:
\begin{align*}
    R(T)&=\mathbb{E}[\sum_{t=1}^T\max_{A\in \mathcal{A}_t}\langle A-A_t,\theta\rangle] 
    \\
&\leq 2s_As_1 \tau + \sum_{t=\tau+1}^T\mathbb{E}[\max_{A\in \mathcal{A}_t}\langle A-A_t,\theta\rangle] 
\\
&\leq 2s_As_1 \tau + \sum_{t=\tau+1}^T \left( \frac{18 \sigma s_A  h_0^2 \nu }{\alpha}\sqrt{\frac{2 \left(s_0 + \frac{2 \nu \sqrt{s_0}  }{\phi_0^2}\right)}{t-1}} + 2s_As_1\left(\Pr(\mathcal{E}_t^c)+\Pr\left(\left(\mathcal{G}_t^{\frac{\alpha}{2 \nu}}\right)^c | \mathcal{E}_t\right)\right)\right).
\end{align*}
We have:
\begin{align*}
    & \sum_{t=\tau+1}^T  2 s_A s_1 \left(\Pr(\mathcal{E}_t^c) + \Pr\left(\left(\mathcal{G}_{t}^{\frac{\alpha}{2 \nu}}\right)^c \middle| \mathcal{E}_t\right) \right) 
    \\
    & \le \sum_{t=\tau+1}^T2s_As_1 \Bigg(  2 \exp\left( -\frac{t \lambda_t^2}{32 \sigma^2 s_A^2} + \log d\right)
    \\
    & \ \ \ \ +  \exp\left( - \frac{t C_0^2}{2}\right) + \exp\left( \log\left(s_0 + \frac{2 \nu \sqrt{s_0}}{\phi_0^2} \right) - \frac{t \alpha}{10 s_A^2 \nu  \left(s_0 + (2 \nu \sqrt{s_0})/\phi_0^2\right)}\right)\Bigg),
\end{align*}
Regarding the series involving $\sqrt{1 /t}$, we have:
\begin{align*}
    \sum_{t=\tau+1}^T\sqrt{\frac{1}{t-1}}& = \sum_{t=\tau}^{T-1}\sqrt{\frac{1}{t}}
    \\
    & \le 1 + \int_{1}^T\sqrt{\frac{1}{t}}dt  
    \\
    & \le 2\sqrt{T}.
\end{align*}
The bounds for 
\begin{align*}
 & \sum_{t=\tau+1}^T2s_As_1 \Bigg(  2 \exp\left( -\frac{t \lambda_t^2}{32 \sigma^2 s_A^2} + \log d\right)
    \\
    & \ \ \ \ +  \exp\left( - \frac{t C_0^2}{2}\right) + \exp\left( \log\left(s_0 + \frac{2 \nu \sqrt{s_0}}{\phi_0^2} \right) - \frac{t \alpha}{10 s_A^2 \nu  \left(s_0 + (2 \nu \sqrt{s_0})/\phi_0^2\right)}\right)\Bigg)
\end{align*}
    hold similarly as is in Theorem~\ref{thm:regret_ub_k2}. 
In summary, we get:
\begin{align*}
    R(T) & \le 2s_As_1 \tau \\
    & \ \ \ + \frac{36\sigma s_A h_0^2 \nu\sqrt{2 \left(s_0+\frac{2\nu\sqrt{s_0}}{\phi_0^2} \right)}}{\alpha}\sqrt{T} + 2s_A s_1 \left(\frac{\pi^2}{3}+\frac{2}{C_0^2}+\left(s_0+\frac{2\nu\sqrt{s_0}}{\phi_0^2}\right)^2\frac{10s_A^2 \nu}{\alpha}\right).
\end{align*}
Looking at the scaling with respect to $d$, $T$, and $s_0$, one can determine $c_3$. This concludes the proof of the first part of the theorem.

Regarding the second part of the theorem, we impose the following condition on $d$:
\begin{itemize}
    \item[(i)] $ \log \log d \ge 4096 \sigma^4 s_A^4 $
    \item[(ii)] $
4 \left( \frac{2 \nu s_0}{\phi_0^2} + \sqrt{ \left(1 + \frac{2 \nu }{\phi_0^2} \right) s_0}\right) \lambda_\tau \le \theta_{\min},
$
\item[(iii)] $ d \ge 100$.
\end{itemize}
With some constant $c_4$, when $ d \ge c_4$, it should be noted that the condition (ii) holds, as
\begin{align*}
    \sqrt{\frac{2 \log \tau \log d}{\tau}} =  \mathcal{O}\left(\frac{1}{s_0}\sqrt{\frac{1}{\log \log d} + \frac{1}{\log s_0}} \right)
\end{align*}
and 
\begin{align*}
    \lambda_0 & = 1/(\log \log d)^{\frac{1}{4}}
    \\
    & = o(1) \quad (\text{as } d \to \infty).
\end{align*}

Then, similarly to the proof in Appendix~\ref{subsec:proof_regret_ub_k2}, we obtain 
\begin{align*}
    R(T) & \le 2s_As_1 \tau \\
    & \ \ \ + \frac{36\sigma s_A h_0^2 \nu\sqrt{2 \left(s_0+\frac{2\nu\sqrt{s_0}}{\phi_0^2} \right)}}{\alpha}\sqrt{T} + 2s_A s_1 \left(\frac{\pi^2}{3}+\frac{2}{C_0^2}+\left(s_0+\frac{2\nu\sqrt{s_0}}{\phi_0^2}\right)^2\frac{10s_A^2 \nu}{\alpha}\right).
\end{align*}
Investigating the scaling with respect to $d$, $T$, and $s_0$, one can determine $c_5$.

\ep

\section{Proof of Lemmas}
\label{sec:proof_lemmas_klarge}

\subsection{Proof of Lemma~\ref{lm:support_recovery_klarge}}

We define $v \coloneqq \hat{\theta}_{0}^{(t)} - \theta$. We first analyze the performance of the initial Lasso estimate.

\begin{lemma}\label{lm:adaptive_lasso_klarge}
     Let $\hat{\Sigma}_t \coloneqq \frac{\sum_{s=1}^t A_s A_s^\top}{t}$ be the empirical covariance matrix of the selected context vectors. Suppose $\hat{\Sigma}_t$ satisfies the compatibility condition with the support $S(\theta)$ with the compatibility constant $\phi_t$. Then, under Assumption~\ref{asm:sparsity_param_klarge}, we have:
     \begin{align*}
        \Pr\left(\| v\|_1 \le \frac{4 s_0 \lambda_t}{\phi_t^2}\right) \ge 1 - 2 \exp\left( - \frac{t \lambda_t^2}{32 \sigma^2 s_A^2} + \log d\right).
    \end{align*} 
\end{lemma}
The next lemma then states that the compatibility constant of $\hat{\Sigma}_t$ does not deviate much  from  the compatibility constant of $\Sigma$.
\begin{lemma}\label{lm:compatibility_holds_klarge}
    Let $C_0 \coloneqq \min\left\{ \frac{1}{2}, \frac{\phi_0^2}{512 s_0 s_A^2 \nu C_{\textnormal{b}}}\right\}$. For all $t \ge \frac{2 \log (2 d^2)}{C_0^2}$, we have:
    \begin{align*}
        \Pr\left( \phi^2(\hat{\Sigma}_t, S(\theta)) \ge \frac{\phi^2_0}{4\nu C_{\textnormal{b}}}\right) \ge 1 - \exp\left( - \frac{t C_0^2}{2}\right).
    \end{align*}
\end{lemma}
Then, we follow the steps of the proof given by \citet{zhou2010thresholded}. Let us define the event $\mathcal{G}_t$ as:
\begin{align*}
    \mathcal{G}_t \coloneqq \left\{\| v\|_1 \le \frac{4 s_0 \lambda_t}{\phi_t^2} \right\}.
\end{align*}
For the rest of this section, we assume that the event $\mathcal{G}_t$ holds. 
Note that:
\begin{align*}
    \|v\|_1 & \ge \|v_{S(\theta)^c}\|_1
    = \sum_{j \in S(\theta)^c} |(\hat{\theta}_0^{(t)})_j| 
    \\
    & \ge \sum_{j \in S(\theta)^c \cap \hat{S}_0^{(t)}}|(\hat{\theta}_0^{(t)})_j|
    \\
    & = \sum_{j \in  \hat{S}_0^{(t)}  \setminus S(\theta)}|(\hat{\theta}_0^{(t)})_j|
    \\
    & \stackrel{(a)}{\ge} | \hat{S}_0^{(t)}  \setminus S(\theta)|  4 \lambda_t,
\end{align*}
where for $(a)$, we used the construction of $ \hat{S}_0^{(t)}$ in the algorithm. We get:
\begin{align*}
    | \hat{S}_0^{(t)}  \setminus S(\theta)| & \le \frac{\|v\|_1}{4 \lambda_t}
    \\
    & \stackrel{(a)}{\le} \frac{s_0}{\phi_t^2},
\end{align*}
where for $(a)$, we used the definition of $\mathcal{G}_t$. 
We have: $\forall j \in S(\theta)$, 
\begin{align*}
    |(\hat{\theta}_{0}^{(t)})_j| & \ge \theta_{\min}  - \|v_{S(\theta)}\|_\infty
    \\
    & \ge \theta_{\min}  - \|v_{S(\theta)}\|_1
    \\
    & \ge \theta_{\min}  - \frac{4 s_0 \lambda_t}{\phi_t^2}.
\end{align*}
Therefore, when $t$ is large enough so that $ 4 \lambda_t \le \theta_{\min}  - \frac{4 s_0 \lambda_t}{\phi_t^2}$, we have: $S(\theta) \subset \hat{S}_0^{(t)}$. 
Using a similar argument, when $t $ is large enough so that  $ 4 \lambda_t \sqrt{\left( 1 + \frac{1}{\phi_t^2}\right)s_0} \le \theta_{\min} - \frac{4 s_0 \lambda_t}{\phi_t^2}$, it holds that $S(\theta) \subset \hat{S}_1^{(t)}$.
From the construction of $\hat{S}_1^{(t)}$ in the algorithm, it also holds that: $\hat{S}_1^{(t)} \subset \hat{S}_0^{(t)}$. Therefore,
\begin{align*}
    \|v\|_1 & \ge \sum_{i \in \hat{S}_0^{(t)} \setminus S(\theta)} | (\hat{\theta}_0^{(t)})_i|
    \\
    & \ge \sum_{i \in \hat{S}_1^{(t)} \setminus S(\theta)} | (\hat{\theta}_0^{(t)})_i|
    \\
    & \ge |\hat{S}_1^{(t)} \setminus S(\theta)| 4 \lambda_t \sqrt{|\hat{S}_0^{(t)}|},
\end{align*}
and 
\begin{align*}
    |\hat{S}_1^{(t)} \setminus S(\theta)| & \le \frac{\|v\|_1}{4 \lambda_t \sqrt{|\hat{S}_0^{(t)}|}}
    \\
    & \le \frac{1}{4 \lambda_t \sqrt{|\hat{S}_0^{(t)}|}} \cdot \frac{4 s_0 \lambda_t}{\phi_t^2}
    \\
    & \le \frac{\sqrt{s_0}}{\phi_t^2}.
\end{align*}
Note that the condition $ 4 \lambda_t \sqrt{\left( 1 + \frac{1}{\phi_t^2}\right)s_0} \le \theta_{\min} - \frac{4 s_0 \lambda_t}{\phi_t^2}$ is equivalent to $ 4 \lambda_t \left(  \sqrt{\left( 1 + \frac{1}{\phi_t^2}\right)s_0} + \frac{s_0 }{\phi_t^2}\right) \le \theta_{\min}$.
This concludes the proof of Lemma~\ref{lm:support_recovery_klarge} by substituting $\phi_t^2 = \phi_0^2/(4\nu C_{\textnormal{b}})$.\ep

\subsection{Proof of Lemmas used in the proof of Lemma~\ref{lm:support_recovery_klarge}}

\subsubsection{Proof of Lemma~\ref{lm:adaptive_lasso_klarge}}

The proof is similar to that given by \citet{oh2020sparsity}. For the sake of brevity, let $S= S(\theta)$. Let us define the loss function:
\begin{align*}
    \ell_t(\theta) \coloneqq \frac{1}{t} \sum_{s=1}^t (r_t - \langle \theta, A_t\rangle)^2.
\end{align*}
The initial Lasso estimate is given by:
\begin{align*}
    \hat{\theta}_t \coloneqq \arg\min_{\theta'}\left\{\ell_t(\theta') + \lambda_t \|\theta'\|_1\right\}.
\end{align*}
From this definition, we get:
\begin{align*}
    \ell_t(\hat{\theta}_t) + \lambda_t \|\hat \theta_t\|_1 \le \ell_t(\theta) + \lambda_t \|\theta\|_1.
\end{align*}
Let us denote $\EXP[\cdot]$ as the expectation over $r_t$ in this section. Note that in view of the previous inequality, we have:
\begin{align*}
    \ell_t (\hat{\theta}_t) - \EXP[\ell_t (\hat{\theta}_t)] +\EXP[\ell_t (\hat{\theta}_t)] -  \EXP[\ell_t(\theta)]  + \lambda_t \|\hat \theta_t\|_1 \le \ell_t(\theta) - \EXP[\ell_t(\theta)] + \lambda_t \|\theta\|_1.
\end{align*}
Denoting $ v_t(\theta) \coloneqq \ell_t (\theta) - \EXP[\ell_t (\theta)]$ and $\mathcal{E}(\theta') \coloneqq \EXP[\ell_t (\theta')] - \EXP[\ell_t (\theta)]$, 
\begin{align*}
    v_t(\hat{\theta}_t) + \mathcal{E}(\hat{\theta}_t) + \lambda_t \|\hat \theta_t\|_1  \le  v_t({\theta}) + \lambda_t \|\theta\|_1.
\end{align*}
Let us define the event $\mathcal{T}_t$: 
\begin{align*}
    \mathcal{T}_t \coloneqq \{|v_t(\hat{\theta}_t) - v_t({\theta})| \le \frac{1}{2}\lambda_t \| \hat{\theta}_t - {\theta}\|_1 \}.
\end{align*}
We can condition on this event in the rest of the proof:
\begin{lemma}\label{lm:empirical_process_bound}
     We have:
    \begin{align*}
        \Pr\left(|v_t(\hat{\theta}_t) - v_t({\theta})| \le \frac{1}{2} \lambda_t \| \hat{\theta}_t - {\theta}\|_1\right) \ge 1 - 2 \exp\left( - \frac{t \lambda_t^2}{32 \sigma^2 s_A^2} + \log d\right).
    \end{align*} 
\end{lemma}
Given the event $\mathcal{T}_t$, we have:
\begin{align*}
    2 \mathcal{E}(\hat{\theta}_t) \le 2 \lambda_t (\|\theta\|_1 - \|\hat{\theta}_t\|_1) + \lambda_t \|\hat{\theta}_t - \theta\|_1.
\end{align*}
By the triangle inequality, 
\begin{align*}
    \|\hat{\theta}_t\|_1 &  =  \|\hat{\theta}_{t, S}\|_1  +  \|\hat{\theta}_{t, S^c}\|_1 
    \\
    & \ge \|{\theta}_S\|_1  - \|\hat{\theta}_{t,S} -  {\theta}_S\| +  \|\hat{\theta}_{t, S^c}\|_1.
\end{align*}
We also have:
\begin{align*}
    \|\hat{\theta}_{t} - \theta \|_1 & = \|(\hat{\theta}_{t} - \theta)_S \|_1 + \|(\hat{\theta}_{t} - \theta)_{S^c} \|_1
    \\
    & = \|\hat{\theta}_{t, S} - \theta_S \|_1 + \|\hat{\theta}_{t, S^c} \|_1.
\end{align*}
Therefore, we get:
\begin{align}
    2 \mathcal{E}(\hat{\theta}_t) & \le 2 \lambda_t \|\theta\|_1 - 2 \lambda_t (\|\theta_S\|_1 - \|\hat{\theta}_{t, S} - \theta_S \|_1 + \|\hat{\theta}_{t, S^c}\|_1) + \lambda_t(\|\hat{\theta}_{t, S} - \theta_S\|_1 + \|\hat{\theta}_{t, S^c}\|_1) \nonumber
    \\
    & = 3 \lambda_t \|\hat{\theta}_{t, S} - \theta_S\|_1 - \lambda_t \|\hat{\theta}_{t, S^c}\|_1. \label{eq:bound_setE}
\end{align}
From the compatibility condition, we get:
\begin{align}
    \|\hat{\theta}_{t, S} - \theta_S\|_1^2 \le \frac{s_0 (\hat{\theta}_{t} - \theta)^\top \hat{\Sigma}_t (\hat{\theta}_{t} - \theta)}{\phi_t^2} \label{eq:bound_}
\end{align}
Using inequality \eqref{eq:bound_setE}, we get:
\begin{align*}
    2 \mathcal{E}(\hat{\theta}_t) + \lambda_t \|\hat{\theta}_t - {\theta}\|_1 & = 2 \mathcal{E}(\hat{\theta}_t) +  \lambda_t\|\hat{\theta}_{t, S^c}\|_1 + \lambda_t \|\hat{\theta}_{t, S} - {\theta}_S\|_1
    \\
    & \le  3 \lambda_t \|\hat{\theta}_{t, S} - \theta_S\|_1  + \lambda_t \|\hat{\theta}_{t, S} - {\theta}_S\|_1
    \\
    & = 4 \lambda_t \|\hat{\theta}_{t, S} - {\theta}_S\|_1
    \\
    & \le  \frac{4 \lambda_t}{{\phi_t}}\sqrt{{s_0 (\hat{\theta}_{t} - \theta)^\top \hat{\Sigma}_t (\hat{\theta}_{t} - \theta)}}
    \\
    & \le { (\hat{\theta}_{t} - \theta)^\top \hat{\Sigma}_t (\hat{\theta}_{t} - \theta)} + \frac{4 \lambda_t^2 s_0}{\phi_t^2}
    \\
    & \le 2 \mathcal{E}(\hat{\theta}_t) + \frac{4 \lambda_t^2 s_0}{\phi_t^2},
\end{align*}
where for the third inequality, we used $4uv \le u^2 + 4v^2$ with $ u =\sqrt{{ (\hat{\theta}_{t} - \theta)^\top \hat{\Sigma}_t (\hat{\theta}_{t} - \theta)}} $ and $v = \frac{\lambda_t \sqrt{s_0}}{\phi_t}$. The last inequality is due to Lemma~\ref{lm:risk_lower_bound}:
\begin{lemma}\label{lm:risk_lower_bound}
    We have:
    \begin{align*}
        \mathcal{E}(\hat{\theta}_t) \ge \frac{1}{2}  { (\hat{\theta}_{t} - \theta)^\top \hat{\Sigma}_t (\hat{\theta}_{t} - \theta)}\;.
    \end{align*}
\end{lemma}
Thus, we get:
\begin{align*}
    \|\hat{\theta}_t - {\theta}\|_1 \le \frac{4 \lambda_t s_0}{\phi_t^2}.
\end{align*}
 This concludes the proof. \ep

 \subsubsection{Proof of Lemma~\ref{lm:compatibility_holds_klarge}}
 
 For the sake of brevity, let $S= S(\theta)$. First, we define the adapted Gram matrix $\Sigma_t \coloneqq \frac{1}{t}\sum_{s=1}^t\EXP[A_sA_s^\top | \mathcal{F}_{s-1}]$. 
From  the construction of the algorithm, $ \EXP[A_s A_s^\top | \mathcal{F}_{s-1}] = \EXP[\sum_{k=1}^K A_{s, k} A_{s, k}^\top \indicator\{k = \argmax_{k'} \langle A_{s,k}, \hat{\theta}_s\rangle\} | \; \hat{\theta}_s]$. 
The following lemma characterizes the expected Gram matrix generated by the algorithm.

\begin{lemma}[Lemma~10 of \citet{oh2020sparsity}]
\label{lem:greedy_compat_klarge}
    Under Assumptions~\ref{asm:relax_sym} and ~\ref{asm:balanced_cov}, for each fixed vector $\theta' \in \mathbb{R}^d$, we have:
    \begin{align*}
        \EXP_{\mathcal{A} \sim p_A}\left[\sum_{k=1,2} A_{k} A_{k}^\top \indicator\{k = \argmax_{k'} \langle A_{k}, \theta'\rangle\}\right]  \succeq \frac{1}{2\nu C_{\textnormal{b}}} \Sigma,
    \end{align*}
    where $A\succeq B$ means that $A - B$  is positive semidefinite.
\end{lemma}
Using Lemma~\ref{lem:greedy_compat_klarge}, we have
\begin{align}
    \Sigma_t \succeq \frac{1}{2\nu C_{\textnormal{b}}} \Sigma. 
\end{align}
By Lemma~6.18 of \citet{buhlmann2011statistics},  Assumption~\ref{asm:comp_cond}, and the definition of the compatibility constant, we get:
\begin{align}
     \phi^2(\Sigma_t, S) \ge  \phi^2(\frac{1}{2\nu C_{\textnormal{b}}}\Sigma, S) \ge \frac{\phi_0^2}{2\nu C_{\textnormal{b}}}.
\end{align}
Furthermore, we have a following adaptive matrix concentration results for $\hat{\Sigma}_t$:
\begin{lemma}\label{lm:matrix_concentration_klarge} Let $C_0 \coloneqq \min\left\{\frac{1}{2}, \frac{\phi_0^2}{512 s_0 s_A^2 \nu C_{\textnormal{b}}}\right\} $. 
We have, for all $t \ge \frac{2 \log(2 d^2)}{C_0^2}$,
\begin{align*}
    \Pr\left(\frac{1}{2s_A^2}\|\hat{\Sigma}_t - \Sigma_t\|_\infty \ge \frac{\phi^2(\Sigma_t, S)}{64 s_0 s_A^2 }\right) \le \exp\left( - \frac{t C_0^2 }{2}\right).
\end{align*}
 \end{lemma}
 We use a following result from \citet{buhlmann2011statistics}:
 \begin{lemma}[Corollary 6.8 in \citet{buhlmann2011statistics}]\label{buhlmann2011_lemma}
Suppose $\Sigma_0$ satisfies the compatibility condition for the set $S$ with $|S| = s_0$, with the compatibility constant $\phi^2(\Sigma_0, S)>0$, and that $ \| \Sigma_0 - \Sigma_1\|_\infty \le \lambda$, where $ \frac{32 \lambda s_0}{\phi^2(\Sigma_0, S)} \le 1$. Then, the compatibility condition also holds for $\Sigma_1$ with the compatibility constant $\frac{\phi^2(\Sigma_0, S)}{2}$, i.e., 
$
     \phi^2(\Sigma_1, S) \ge \frac{\phi^2(\Sigma_0, S)}{2} .
$
  \end{lemma}
Combining the above results, we get, for all $t \ge \frac{2 \log (2 d^2)}{C_0^2}$:
\begin{align*}
    \phi^2(\hat{\Sigma}_t, S) & \ge \frac{\phi^2({\Sigma}_t, S) }{2}
    \\
    & \ge \frac{\phi^2_0}{4 \nu C_{\textnormal{b}}},
\end{align*}
with probability at least $1 - \exp\left( - \frac{t C_0^2}{2}\right)$. This concludes the proof.\ep

 \subsubsection{Proof of Lemma~\ref{lm:empirical_process_bound}}
 Let us denote $\hat{\theta} = \hat{\theta}_t$ for simplicity. We compute $v_t(\hat{\theta})$ as:
 \begin{align*}
     v_t(\hat{\theta}) & = \ell_t(\hat{\theta})  - \EXP[\ell_t(\hat{\theta})]
     \\
     & = \frac{1}{t}\sum_{s =1}^t (r_s - \langle \hat{\theta}, A_s\rangle)^2 - \frac{1}{t}\sum_{s =1}^t\EXP\left[ (r_s - \langle \hat{\theta}, A_s\rangle)^2\right]
     \\
     & = \frac{1}{t}\sum_{s =1}^t ( \langle {\theta}, A_s\rangle + \varepsilon_s - \langle \hat{\theta}, A_s\rangle)^2 - \frac{1}{t}\sum_{s =1}^t\EXP\left[ ( \langle {\theta}, A_s\rangle + \varepsilon_s - \langle \hat{\theta}, A_s\rangle)^2\right]
     \\
     & = \frac{1}{t}\sum_{s =1}^t(2 \varepsilon_s \langle \theta - \hat{\theta}, A_s\rangle  + \varepsilon_s^2 - \EXP[\varepsilon_s^2]).
 \end{align*}
 We also have that:
 \begin{align*}
     v_t({\theta}) & = \frac{1}{t}\sum_{s=1}^t (\varepsilon_s^2 - \EXP[\varepsilon_s^2]).
 \end{align*}
 Therefore, we can compute:
 \begin{align*}
     v_t(\hat{\theta}) - v_t({\theta}) & = \frac{1}{t}\sum_{s=1}^t 2 \varepsilon_s \langle \theta - \hat{\theta}, A_s\rangle 
     \\
     & \le \frac{2}{t} \left\| \sum_{s=1}^t \varepsilon_s A_s \right\|_\infty \| \theta - \hat{\theta}\|_1,
 \end{align*}
 where we used Hölder's inequality in the above inequality.
 We have that:
 \begin{align*}
     \Pr \left(\frac{2}{t}\left\| \sum_{s=1}^t \varepsilon_s A_s \right\|_\infty  \le \lambda \right) \ge 1 - \sum_{i=1}^d \Pr\left(\frac{2}{t} \left| \sum_{s=1}^t \varepsilon_s (A_s)_{i}  \right| > \lambda \right),
 \end{align*}
 where $(A_s)_{i}$ is the $i$-th element of $A_s$. Define $\tilde{\mathcal{F}}_t$ as the $\sigma$-algebra generated by the random variables $(A_1, \mathcal{A}_1, \varepsilon_1, \ldots, A_{t}, \mathcal{A}_{t}, \varepsilon_t, \mathcal{A}_{t+1})$. For each $i \in [d]$, we get $ \EXP[ \varepsilon_s (A_s)_{i} | \tilde{\mathcal{F}}_{s-1}]= (A_s)_{i} \EXP[ \varepsilon_s | \tilde{\mathcal{F}}_{s-1}] = 0$.
 Thus, for each $i \in [d]$, $\{\varepsilon_s (A_s)_{i}\}_{s=1}^t$ is a martingale difference sequence adapted to the filtration $\tilde{\mathcal{F}}_1 \subset \ldots \subset \tilde{\mathcal{F}}_{t-1}$. By Assumption~\ref{asm:sparsity_param_klarge}, we have $|(A_s)_{i}| \le s_A$. We compute, for each $\alpha \in \mathbb{R}$,
 \begin{align*}
     \EXP[\exp(\alpha \varepsilon_s (A_s)_{i}) | \tilde{\mathcal{F}}_{s-1}] & \le \EXP[\exp(\alpha \varepsilon_s s_A) | \tilde{\mathcal{F}}_{s-1}]
     \\
     & \le \exp\left(\frac{\alpha^2 s_A^2 \sigma^2}{2}\right).
 \end{align*}
 Therefore $ \varepsilon_s (A_s)_{i}$ is also a sub-Gaussian random variable with the variance proxy $ (s_A \sigma)^2$. Next, we use the concentration results by \citet{wainwright2019high}, Theorem 2.19:
 \begin{theorem}\label{thm:martingale_bernstein}
    Let $(Z_t, \tilde{\mathcal{F}}_t)_{t=1}^\infty$ be a martingale difference sequence, and assume that for all $\alpha \in \mathbb{R}$, $\EXP[\exp( \alpha Z_s) | \tilde{\mathcal{F}}_{s-1}] \le \exp(\frac{\alpha^2 \sigma^2}{2})$ with probability one. Then, for all $x \ge 0$, we get:
    \begin{align*}
        \Pr\left( \left|\sum_{s=1}^t Z_s \right| \ge  x\right) \le 2 \exp\left( -\frac{x^2}{2 t \sigma^2}\right).
    \end{align*}
 \end{theorem}
 From these results, we get:
 \begin{align*}
     \Pr\left(\left| \sum_{s=1}^t \varepsilon_s (A_s)_{i}  \right| > \frac{t \lambda}{2} \right) & \le 2 \exp\left( - \frac{t \lambda^2}{8 \sigma^2 s_A^2}\right).
 \end{align*}
Taking $\lambda = \frac{1}{2}\lambda_t$,
 \begin{align*}
     \Pr \left(\frac{2}{t}\left\| \sum_{s=1}^t \varepsilon_s A_s \right\|_\infty  \le \frac{\lambda_t}{2} \right) & \ge 1 - 2d \exp\left( - \frac{t \lambda_t^2}{32 \sigma^2 s_A^2}\right)
     \\
     & = 1 - 2 \exp\left( - \frac{t \lambda_t^2}{32 \sigma^2 s_A^2} + \log d\right).
 \end{align*}
 \ep

 \subsubsection{Proof of Lemma~\ref{lm:risk_lower_bound}}
 
 We denote $\hat{\theta}_t = \hat{\theta}$ for brevity. From the definitions of $\mathcal{E}(\theta')$ and $\ell_t(\theta')$,
 \begin{align*}
     \mathcal{E}(\hat{\theta}) & = \EXP[\ell_t (\hat{\theta})] - \EXP[\ell_t (\theta)]
     \\
     & = \frac{1}{t} \EXP\left[\sum_{s=1}^t (\langle \hat{\theta} - \theta, A_s\rangle + \varepsilon_s)^2 - \sum_{s=1}^t \varepsilon_s^2\right]
     \\
     & = \frac{1}{t}\sum_{s=1}^t \langle \hat{\theta} - \theta, A_s\rangle^2
     \\
     & = \frac{1}{t}\sum_{s=1}^t ( \hat{\theta} - \theta)^\top A_s A_s^\top  ( \hat{\theta} - \theta)
     \\
     & =  ( \hat{\theta} - \theta)^\top \hat{\Sigma}_t  ( \hat{\theta} - \theta)
     \\
     & \ge \frac{1}{2} ( \hat{\theta} - \theta)^\top \hat{\Sigma}_t  ( \hat{\theta} - \theta),
 \end{align*}
 where for the inequality, we used the positive semi-definiteness of $\hat{\Sigma}_t$. \ep

 \subsubsection{Proof of Lemma~\ref{lem:greedy_compat_klarge}}
 The proof is almost identical to the proof of Lemma~10 in \citet{oh2020sparsity}. \ep
 
  \subsubsection{Proof of Lemma~\ref{lm:matrix_concentration_klarge}}
 
 Let us define $\gamma^{i j}_t(A_t)$ as:
 \begin{align*}
     \gamma^{i j}_t(A_t) \coloneqq \frac{1}{2 s_A^2} \left( (A_t)_i (A_t)_j - \EXP[ (A_t)_i (A_t)_j \;|\; \mathcal{F}_{t-1}]\right),
 \end{align*}
 where $(A_t)_i$ is the $i$-th element of $A_t$. We have, following a Bernstein-like inequality for the adapted data:
 \begin{lemma}[Bernstein-like inequality for the adapted data \cite{oh2020sparsity}]\label{lm:bernstein-like2}
 Suppose for all $t \ge 1$, for all $ 1 \le i \le j \le d$, $\EXP[\gamma^{i j}_t(A_t) | \mathcal{F}_{t-1}] = 0$ and $\EXP[| \gamma^{i j}_t(A_t)|^m\; | \;\mathcal{F}_{t-1} ] \le m!$ for all integer $m \ge 2$. Then, for all $x >0$, and for all integer $t \ge 1$, we have:
 \begin{align*}
     \Pr\left( \max_{1 \le i \le j \le d} \left|\frac{1}{t} \sum_{s = 1}^t \gamma^{i j}_s(A_s) \right| \ge x + \sqrt{2 x} + \sqrt{\frac{4 \log (2 d^2)}{t}} + \frac{2 \log (2 d^2)}{t}\right) \le \exp\left(- \frac{tx}{2}\right).
 \end{align*}
 \end{lemma}
 Note that $\frac{1}{2 s_A^2} \|\hat{\Sigma}_t - {\Sigma}_t \|_\infty = \max_{1 \le i \le j \le d} \left|\frac{1}{t} \sum_{s = 1}^t \gamma^{i j}_s(A_s) \right| $, $ \EXP[\gamma^{i j}_t(A_t) | \mathcal{F}_{t-1}] = 0$, and $\EXP[| \gamma^{i j}_t(A_t)|^m\; | \;\mathcal{F}_{t-1} ] \le 1$ for all integer $m \ge 2$. Therefore, we can apply Lemma~\ref{lm:bernstein-like2}:
 \begin{align*}
     \Pr \left(\frac{1}{2 s_A^2} \|\hat{\Sigma}_t - {\Sigma}_t \|_\infty \ge x + \sqrt{2 x} + \sqrt{\frac{4 \log (2 d^2)}{t}} + \frac{2 \log (2 d^2)}{t}\right) \le \exp\left(- \frac{tx}{2}\right).
 \end{align*}
 For all $t \ge \frac{2 \log (2 d^2)}{C_0^2}$ with $C_0 \coloneqq \min\left\{\frac{1}{2}, \frac{\phi_0^2}{256 s_0 s_A^2 \nu C_{\textnormal{b}}}\right\} $, taking $x = C_0^2$, 
 \begin{align*}
      x + \sqrt{2 x} + \sqrt{\frac{4 \log (2 d^2)}{t}} + \frac{2 \log (2 d^2)}{t} & \le 2 C_0^2 + 2 \sqrt{2}C_0
      \\
      & \le 4 C_0
      \\
      & \le \frac{\phi_0^2}{128 s_0 s_A^2 \nu C_{\textnormal{b}}}
      \\
      & \le \frac{\phi^2(\Sigma_t, S)}{64 s_0 s_A^2 }\;.
 \end{align*}
 In summary, for all $ t \ge \frac{2 \log (2 d^2)}{C_0^2}$, we get:
 \begin{align*}
      \Pr \left(\frac{1}{2 s_A^2} \|\hat{\Sigma}_t - {\Sigma}_t \|_\infty \ge \frac{\phi^2(\Sigma_t, S)}{64 s_0 s_A^2  }\right) & \le \Pr \left(\frac{1}{2 s_A^2} \|\hat{\Sigma}_t - {\Sigma}_t \|_\infty \ge C_0^2 + \sqrt{2} C_0 + \sqrt{\frac{4 \log (2 d^2)}{t}} + \frac{2 \log (2 d^2)}{t}\right) 
      \\
      & \le  \exp\left( - \frac{t C_0^2}{2}\right).
 \end{align*}
 This concludes the proof. \ep

 \subsection{Proof of Lemma~\ref{lm:lowerbound_eig_S_klarge}}
 
 For a fixed $\hat{S}$, first we define the adapted Gram matrix on the estimated support as 
 $$
 \Sigma_t \coloneqq \frac{1}{t}\sum_{s=1}^t\EXP[A_s(\hat{S})A_s(\hat{S})^\top | \mathcal{F}_{s-1}].
 $$ 
From  the construction of the algorithm, $ \EXP[A_s(\hat{S}) A_s(\hat{S})^\top | \mathcal{F}_{s-1}] = \EXP[\sum_{k=1}^K A_{s, k}(\hat{S}) A_{s, k}(\hat{S})^\top \indicator\{k = \argmax_{k'} \langle A_{s,k}, \hat{\theta}_s\rangle\} | \; \hat{\theta}_s]$. 
Recall that for each $B \subset [d]$, $\Sigma_B \coloneqq \frac{1}{K} \sum_{k=1}^K \EXP_{\mathcal{A}\sim p_A} \left[A_{k}(B) A_{k}(B)^\top\right]$, where $ A_{k}(B)$ is a $|B|$-dimensional vector extracted the elements of $A_{k}$ with indices in $B$. 
The following lemma characterizes the expected Gram matrix generated by the algorithm.

\begin{lemma}\label{lem:greedy_positivedefnite_klarge}
     Fix $\hat{S} $ such that $ S(\theta) \subset \hat{S} $ and $|\hat{S}| \le s_0 + (4 \nu C_{\textnormal{b}} \sqrt{s_0})/\phi_0^2$. Fix $\theta' \in \mathbb{R}^d$. Under Assumption~\ref{asm:comp_cond}, \ref{asm:relax_sym}, and \ref{asm:balanced_cov}, we have:
    \begin{align*}
        \EXP_{\mathcal{A} \sim p_A}\left[\sum_{k \in [K]} A_{k}(\hat{S}) A_{k}(\hat{S})^\top \indicator\{k = \argmax_{k'} \langle A_{k}, \theta'_{\hat{S}}\rangle\}\right]  \succeq \frac{1}{2\nu  C_{\textnormal{b}}} \Sigma_{\hat{S}},
    \end{align*}
    where $A\succeq B$ means that $A - B$  is positive semidefinite.
\end{lemma}
 First, we prove the lower bound on the smallest eigenvalue of the expected covariance matrices. Let $\Sigma_{\hat{S}} \coloneqq \frac{1}{t} \sum_{s =1}^t \EXP[A_{s}( \hat{S}) A_{s}( \hat{S})^\top \mid \mathcal{F}_{t-1}]$. By Assumption~\ref{asm:cov_div_klarge} and the construction of the algorithm, under the event $\mathcal{E}_t$, we get:
 \begin{align*}
     \lambda_{\min} (\Sigma_{\hat{S}}) & =\lambda_{\min}\left(  \frac{1}{t} \sum_{s =1}^t \EXP \left[ \sum_{k =1}^K A_{s, k, \hat{S}} A_{s, k, \hat{S}} \indicator \{ k = \argmax_{k'} \langle A_{k'}, \hat{\theta}_s \rangle\} \mid \hat{\theta}_s \right]  \right)
     \\
     & \ge \sum_{s=1}^t\lambda_{\min} \left(\frac{1}{t} \EXP \left[ \sum_{k =1}^K A_{s, k, \hat{S}} A_{s, k, \hat{S}} \indicator \{ k = \argmax_{k'} \langle A_{k'}, \hat{\theta}_s \rangle\} \mid \hat{\theta}_s \right]\right)
     \\
     & \ge \frac{\alpha}{2\nu  C_{\textnormal{b}}},
 \end{align*}
 where for the first inequality, we used the concavity of $\lambda_{\min} ( \cdot)$ over the positive semi-definite matrices. Next, we prove the upper bound on the largest eigenvalue of $ A_{s}( \hat{S}) A_{s}( \hat{S})^\top$:
 \begin{align*}
     \lambda_{\max} (A_{s}( \hat{S}) A_{s}( \hat{S})^\top)  
     & = \max_{\|v\|_2 = 1} v^\top A_{s}( \hat{S})  A_{s}( \hat{S})^\top v  
      \\
      & \stackrel{(a)}{\le} \max_{\|v\|_2 = 1}\|v\|_1^2 \|A_s ( \hat{S}) \|_\infty^2
      \\
      & \le |\hat{S}| s_A^2
      \\
      & \le \left(s_0 + (4 \nu C_{\textnormal{b}} \sqrt{s_0})/\phi_0^2\right) s_A^2,
 \end{align*}
 where for $(a)$, we used H\"older's inequality. Now recall the matrix Chernoff inequality by \citet{tropp2011user}:
\begin{theorem}[Matrix Chernoff, Theorem 3.1 of \citet{tropp2011user}]\label{lem:tropp2011}
Let $\mathcal{F}_1 \subset \mathcal{F}_2 \subset \ldots \subset \mathcal{F}_t$ be a filtration and a consider a finite sequence $\{X_s\}$ of positive semi-definite matrices with dimension $d$, adapted to the filtration. Suppose $ \lambda_{\max} (X_k) \le R$ almost surely. Define the finite series: $Y \coloneqq \sum_{s=1}^t X_s$ and $W \coloneqq \sum_{s =1}^t \EXP[X_s \mid \mathcal{F}_{s-1}]$. Then, for all $\mu \ge 0$, for all $\delta \in [0,1)$, we have:
\begin{align*}
    \Pr\left( \lambda_{\min} (Y) \le (1 - \delta) \mu \text{ and } \lambda_{\min}( W) \ge \mu  \right) \le d \left( \frac{e^{ - \delta}}{(1 - \delta)^{1 - \delta}}\right)^{\frac{\mu}{R}}.
\end{align*}
\end{theorem} 
Taking $R = \left(s_0 + (4 \nu C_{\textnormal{b}} \sqrt{s_0})/\phi_0^2\right) s_A^2$, $X_s = A_{s, \hat{S}} A_{s, \hat{S}}^\top$, $Y = t \hat{\Sigma}_{\hat{S}}$, $W = t {\Sigma}_{\hat{S}}$, $\delta = 1/2$, $ \mu = t \frac{\alpha}{2\nu  C_{\textnormal{b}}}$: 
\begin{align*}
     & \Pr\left( \lambda_{\min} (t \hat{\Sigma}_{\hat{S}}) \le \frac{1}{2} t \frac{\alpha}{2\nu  C_{\textnormal{b}}} \text{ and } \lambda_{\min}(  t {\Sigma}_{\hat{S}}) \ge t \frac{\alpha}{2\nu  C_{\textnormal{b}}}  \right) 
     \\
     & \le  \left(s_0 + \frac{4 \nu C_{\textnormal{b}}\sqrt{s_0}}{\phi_0^2} \right)\left( \frac{e^{ - 0.5}}{0.5^{0.5}}\right)^{\frac{t }{R} \frac{\alpha}{2\nu  C_{\textnormal{b}}}}
     \\
     & \le \exp\left( \log\left(s_0 + \frac{4 \nu C_{\textnormal{b}}\sqrt{s_0}}{\phi_0^2} \right) - \frac{t \alpha}{20 s_A^2\nu C_{\textnormal{b}}\left(s_0 + (4 \nu C_{\textnormal{b}} \sqrt{s_0})/\phi_0^2\right) } \right),
\end{align*}
where for the last inequality, we used $ -0.5 - 0.5 \log(0.5) < - \frac{1}{10}$.  This concludes the proof.\ep

\subsection{Proof of Lemma~\ref{lm:LS_estim_risk_klarge}}
In this proof, we denote $\hat{S} = \hat{S}_{1}^{(t)}$ and $\varepsilon = (\varepsilon_1, \ldots, \varepsilon_t)^\top$.
Assume $\lambda_{\min} (\hat{\Sigma}_{\hat{S}})\ge \lambda$. We have:
\begin{align*}
    \|\hat{\theta}_{t+1} - \theta\|_2 & = \| (A(\hat{S})^\top A(\hat{S}))^{-1} A(\hat{S})^\top R - \theta\|_2
    \\
    & = \| (A(\hat{S})^\top A(\hat{S}))^{-1} A(\hat{S})^\top (A \theta + \varepsilon) -\theta\|_2
    \\
    & = \| (A(\hat{S})^\top A(\hat{S}))^{-1} A(\hat{S})^\top (A(\hat{S}) \theta({\hat{S}}) + \varepsilon) -\theta\|_2
    \\
    & = \| (A(\hat{S})^\top A(\hat{S}))^{-1} A(\hat{S})^\top \varepsilon \|_2
    \\
    & \le \| (A(\hat{S})^\top A(\hat{S}))^{-1} \|_2 \|A(\hat{S})^\top \varepsilon \|_2
    \\
    & \le \frac{1}{\lambda t} \|A (\hat{S})^\top \varepsilon \|_2.
\end{align*}
We get (note that we are conditioning on a fixed $\hat{S}$ during the proof):
\begin{align*}
    \Pr \left( \|\hat{\theta}_{t+1} - \theta \|_2 \ge x \; \text{and} \; \lambda_{\min}(\hat{\Sigma}_{\hat{S}}) \ge \lambda \right) & =  \Pr \left( \|\hat{\theta}_{t+1} - \theta \|_2 \ge x \; \middle\vert \; \lambda_{\min}( \hat{\Sigma}_{\hat{S}}) \ge \lambda \right)\Pr(\lambda_{\min}( \hat{\Sigma}_{\hat{S}}) \ge \lambda)
    \\
    & \le \Pr \left( \|A (\hat{S})^\top \varepsilon \|_2  \ge \lambda t x \; \middle\vert \; \lambda_{\min}( \hat{\Sigma}_{\hat{S}}) \ge \lambda \right)\Pr(\lambda_{\min}( \hat{\Sigma}_{\hat{S}}) \ge \lambda)
    \\
    & \le \Pr \left( \|A (\hat{S})^\top \varepsilon \|_2  \ge \lambda t x \right)
    \\
    & \le \sum_{i = 1}^d \Pr \left(\left| \sum_{s=1}^t \varepsilon_s (A_{s})_i \indicator \left\{ i \in \hat{S}\right\}\right|\ge \frac{\lambda t x}{\sqrt{s_0 + \frac{4 \nu C_{\textnormal{b}}\sqrt{s_0}}{\phi_0^2}}} \right)
    \\
    & = \sum_{i \in \hat{S}} \Pr \left(\left| \sum_{s=1}^t \varepsilon_s  (A_{s})_i \right|\ge \frac{\lambda t x}{\sqrt{s_0 + \frac{4 \nu C_{\textnormal{b}}\sqrt{s_0}}{\phi_0^2}}} \right)
    \\
    & \stackrel{(a)}{\le} 2 \left( s_0 + \frac{4  \nu C_{\textnormal{b}}\sqrt{s_0}}{\phi_0^2}\right) \exp \left( - \frac{\lambda^2 t x^2}{2 \sigma^2 s_A^2 \left( s_0 + \frac{4  \nu C_{\textnormal{b}} \sqrt{s_0}}{\phi_0^2}\right)}\right),
\end{align*}
where for $(a)$, we used Theorem~\ref{thm:martingale_bernstein}. This concludes the proof.\ep

 \subsection{Proof of Lemma~\ref{lm:inst_regret_bound_klarge}}
 We follow the proof strategy of Lemma~6 in \citet{bastani2021mostly}. Let $r_t^\pi$ be the instantaneous expected regret of algorithm $\pi$ at round $t$ defined as:
 \begin{align*}
     r_t^\pi \coloneqq \EXP\left[ \max_{A \in \mathcal{A}_t } \langle A - A_t, \theta \rangle\right].
 \end{align*}
 Let us define the events $\mathcal{R}_k \coloneqq \{ \mathcal{A}_t \in \mathbb{R}^{K \times d}:  k \in \argmax_{k'} \langle A_{t, k'}, \theta \rangle\}$ and $ \mathcal{G}_{t}^{\lambda} \coloneqq \left\{\lambda_{\min } ( \hat{\Sigma}_{\hat{S}} ) \ge \lambda \right\}$. We have:
 \begin{align*}
     r_t^\pi & \le \sum_{k=1}^K \EXP\left[  r_t^\pi \mid  \mathcal{A}_t \in \mathcal{R}_k\right]  \Pr\left(\mathcal{A}_t \in \mathcal{R}_k\right).
 \end{align*}
 The term $\EXP\left[  r_t^\pi \mid  \mathcal{A}_t \in \mathcal{R}_k\right] $ can be further computed as:
 \begin{align*}
     \EXP\left[  r_t^\pi \mid  \mathcal{A}_t \in \mathcal{R}_k\right]  & = \EXP\left[  \langle A_{t, k} - A_t, \theta \rangle \mid  \mathcal{A}_t \in \mathcal{R}_k\right]
     \\
     & \le \EXP\left[  \indicator\left\{ \langle  A_t, \hat{\theta}_t \rangle  \ge \langle A_{t, k}, \hat{\theta}_t \rangle\right\} \langle A_{t, k} - A_t, \theta \rangle \mid  \mathcal{A}_t \in \mathcal{R}_k\right]
     \\
     & \le \sum_{\ell \neq k} \EXP\left[  \indicator\left\{\langle A_{t, \ell}, \hat{\theta}_t \rangle \ge \langle  A_{t, k}, \hat{\theta}_t \rangle \right\} \langle A_{t, k} - A_{t, \ell}, \theta \rangle \mid  \mathcal{A}_t \in \mathcal{R}_k\right]
     \\
     & \le \sum_{\ell \neq k} \EXP\left[  \indicator\left\{\langle A_{t, \ell}, \hat{\theta}_t \rangle \ge \langle  A_{t, k}, \hat{\theta}_t \rangle \right\} \langle A_{t, k} - A_{t, \ell}, \theta \rangle \mid  \mathcal{A}_t \in \mathcal{R}_k, \ \mathcal{E}_t, \  \mathcal{G}_{t}^{\frac{\alpha}{4 \nu C_{\textnormal{b}}}}\right] 
     \\
     & \ \ \ + 2 (K - 1)  s_A s_1  \left(\Pr(\mathcal{E}_t^c) + \Pr\left(\left(\mathcal{G}_{t}^{\frac{\alpha}{4 \nu C_{\textnormal{b}}}}\right)^c \middle| \mathcal{E}_t\right) \right).
\end{align*}
 Let us denote the event $I_h \coloneqq \{\mathcal{A}_t \in \mathbb{R}^{K \times d} : \langle A_{t, k} - A_{t, \ell}, \theta \rangle \in (2 \delta s_Ah, 2 \delta s_A(h + 1) ]\}$ where
 \begin{align*}
     \delta = \frac{\sigma s_A  \nu C_{\textnormal{b}}}{\alpha} \sqrt{\frac{32 \left(s_0 + \frac{4 \nu C_{\textnormal{b}} s_0 }{\phi_0^2}\right)}{t -1}}.
 \end{align*}
 By conditioning on $I_h$, we get:
 \begin{align*}
     & \EXP\left[  \indicator\left\{\langle A_{t, \ell}, \hat{\theta}_t \rangle \ge \langle  A_{t, k}, \hat{\theta}_t \rangle  \right\} \langle A_{t, k} - A_{t, \ell}, \theta \rangle \mid  \mathcal{A}_t \in \mathcal{R}_k \cap I_h, \ \mathcal{E}_t, \ \mathcal{G}_{t}^{\frac{\alpha}{4 \nu C_{\textnormal{b}}}}\right]
     \\
     & \le \sum_{h=0}^{\lceil s_1 / \delta \rceil}\EXP\left[  \indicator\left\{\langle A_{t, \ell}, \hat{\theta}_t \rangle \ge \langle  A_{t, k}, \hat{\theta}_t \rangle  \right\} \langle A_{t, k} - A_{t, \ell}, \theta \rangle \mid  \mathcal{A}_t \in \mathcal{R}_k \cap I_h, \ \mathcal{E}_t, \ \mathcal{G}_{t}^{\frac{\alpha}{4 \nu C_{\textnormal{b}}}} \right] \Pr(\mathcal{A}_t \in I_h)
     \\
     & \stackrel{(a)}{\le}  \sum_{h=0}^{\lceil s_1 / \delta \rceil} 2 \delta s_A (h + 1)\EXP\left[  \indicator\left\{\langle A_{t, \ell}, \hat{\theta}_t \rangle \ge \langle  A_{t, k}, \hat{\theta}_t \rangle \right\} \mid  \mathcal{A}_t \in \mathcal{R}_k \cap I_h, \ \mathcal{E}_t, \ \mathcal{G}_{t}^{\frac{\alpha}{4 \nu C_{\textnormal{b}}}}
     \right]
     \\
     & \qquad \qquad \qquad \qquad \qquad \qquad \qquad \qquad \cdot \Pr \left(\langle A_{t, k} - A_{t, \ell}, \theta \rangle  \in (0, 2 \delta s_A(h + 1) ] \right)
     \\
     & \stackrel{(b)}{\le} \sum_{h=0}^{\lceil s_1 / \delta \rceil} 4 \delta^2 s_A^2 (h + 1)^2 C_{\text{m}}\Pr\left(  \langle A_{t, \ell}, \hat{\theta}_t \rangle \ge \langle  A_{t, k}, \hat{\theta}_t \rangle  \mid  \mathcal{A}_t \in \mathcal{R}_k \cap I_h, \ \mathcal{E}_t, \ \mathcal{G}_{t}^{\frac{\alpha}{4 \nu C_{\textnormal{b}}}}
     \right),
 \end{align*}
 where for $(a)$, we used the definition of $I_h$ and for $(b)$, we used Assumption~\ref{asm:margin}. Under the event $ \mathcal{A}_t \in I_h$, the event $\langle A_{t, \ell}, \hat{\theta}_t \rangle \ge \langle  A_{t, k}, \hat{\theta}_t \rangle$ happens only when at least one of the events $ \langle A_{t, k} , \theta - \hat{\theta} \rangle \ge \delta s_A h$ or $ \langle A_{t, \ell} , \hat{\theta} - \theta \rangle \ge \delta s_A h $ holds. Therefore, 
 \begin{align*}
     & \Pr\left(  \langle A_{t, \ell}, \hat{\theta}_t \rangle \ge \langle  A_{t, k}, \hat{\theta}_t \rangle
     \mid  \mathcal{A}_t \in \mathcal{R}_k \cap I_h, \ \mathcal{E}_t, \ \mathcal{G}_{t}^{\frac{\alpha}{4 \nu C_{\textnormal{b}}}}
     \right) 
     \\
     & \le \Pr\left(  \langle A_{t, k} , \theta - \hat{\theta} \rangle \ge \delta s_A h   \mid  \mathcal{A}_t \in \mathcal{R}_k \cap I_h, \ \mathcal{E}_t, \ \mathcal{G}_{t}^{\frac{\alpha}{4 \nu C_{\textnormal{b}}}}
     \right)  + \Pr\left( \langle A_{t, \ell} , \hat{\theta} - \theta \rangle \ge \delta s_A h  \mid \mathcal{A}_t \in \mathcal{R}_k \cap I_h, \ \mathcal{E}_t, \ \mathcal{G}_{t}^{\frac{\alpha}{4 \nu C_{\textnormal{b}}}}
     \right) 
     \\
     & \stackrel{(a)}{\le} \Pr\left(   \| \theta - \hat{\theta}\|_2 \ge \delta  h   \mid  \mathcal{A}_t \in \mathcal{R}_k \cap I_h, \ \mathcal{E}_t, \ \mathcal{G}_{t}^{\frac{\alpha}{4 \nu C_{\textnormal{b}}}}
     \right)  + \Pr\left(  \| \theta - \hat{\theta}\|_2 \ge \delta  h  \mid  \mathcal{A}_t \in \mathcal{R}_k \cap I_h, \ \mathcal{E}_t, \ \mathcal{G}_{t}^{\frac{\alpha}{4 \nu C_{\textnormal{b}}}}
     \right) 
     \\
     & = 2 \Pr\left(   \| \theta - \hat{\theta}\|_2 \ge \delta  h  \mid  \mathcal{A}_t \in \mathcal{R}_k \cap I_h, \ \mathcal{E}_t, \ \mathcal{G}_{t}^{\frac{\alpha}{4 \nu C_{\textnormal{b}}}}
     \right),
 \end{align*}
 where for $(a)$, we used the Cauchy–Schwarz inequality. Let us denote $s' = s_0 + \frac{4 \nu C_{\textnormal{b}} \sqrt{s_0}}{\phi_0^2}$. Then, using Lemma~\ref{lm:LS_estim_risk_klarge}, we get:
 \begin{align*}
     \Pr\left(   \| \theta - \hat{\theta}\|_2 \ge \delta  h   \mid  \mathcal{A}_t \in \mathcal{R}_k \cap I_h, \ \mathcal{E}_t, \ \mathcal{G}_{t}^{\frac{\alpha}{4 \nu C_{\textnormal{b}}}}
     \right) & \le 2 s'\exp \left( -\frac{\alpha^2 t \delta^2 h^2}{32 \sigma^2 s_A^2 \nu^2 C_{\textnormal{b}}^2 s'} \right)
     \\
     & = 2 s'\exp \left( -h^2\right).
 \end{align*}
 We also trivially have that:
 \begin{align*}
     \Pr\left(   \| \theta - \hat{\theta}\|_2 \ge \delta  h  \mid  \mathcal{A}_t \in \mathcal{R}_k \cap I_h, \ \mathcal{E}_t, \ \mathcal{G}_{t}^{ \frac{\alpha}{4 \nu C_{\textnormal{b}}}}
     \right) & \le 1.
 \end{align*}
 Therefore, we can bound the expected instantaneous regret as:
 \begin{align*}
    & r_t^\pi  \le \sum_{k=1}^K \EXP\left[  r_t^\pi \mid  \mathcal{A}_t \in \mathcal{R}_k\right]  \Pr\left(\mathcal{A}_t \in \mathcal{R}_k\right)
     \\
     & \le \sum_{k=1}^K \left(\sum_{\ell \neq k}\left( \sum_{h=0}^{\lceil s_1/\delta \rceil} \left(4 \delta^2 s_A^2 (h + 1)^2 C_{\text{m}} \min \left\{ 1, 4s' \exp\left( -h^2\right) \right\} \right) \right) \right.
     \\
     & \ \ \ \left.+ 2 (K-1)  s_A s_1 \left(\Pr(\mathcal{E}_t^c) + \Pr\left(\left(\mathcal{G}_{t}^{\frac{\alpha}{4 \nu C_{\textnormal{b}}}}\right)^c\right) \right) \right)  \cdot \Pr\left(\mathcal{A}_t \in \mathcal{R}_k \right)
     \\
     & \stackrel{(a)}{\le} G \sum_{h=0}^{\lceil s_1/\delta \rceil} (h + 1)^2\min \left\{ 1, 4 s' \exp\left( -h^2\right) \right\} + 2 (K-1)  s_A s_1  \left(\Pr(\mathcal{E}_t^c) + \Pr\left(\left(\mathcal{G}_{t}^{\frac{\alpha}{4 \nu C_{\textnormal{b}}}}\right)^c \middle| \mathcal{E}_t\right) \right)
     \\
     & \le G\left( \sum_{h=0}^{h_0} (h + 1)^2 + \sum_{h= h_0 + 1}^ {h_{\max}}  4 s' (h+1)^2\exp\left( -h^2\right) \right)  + 2 (K-1)  s_A s_1  \left(\Pr(\mathcal{E}_t^c) + \Pr\left(\left(\mathcal{G}_{t}^{\frac{\alpha}{4 \nu C_{\textnormal{b}}}}\right)^c \middle| \mathcal{E}_t\right) \right)
 \end{align*}
 where for brevity $G=4 \delta^2 s_A^2  C_{\text{m}}(K -1) $, 
 where for $(a)$, we used $\sum_{k=1}^K \Pr\left(\mathcal{A}_t \in \mathcal{R}_k\right)= 1$ from Assumption~\ref{asm:margin} and we set $ h_0 \coloneqq \lfloor \sqrt{\log 4 s'} + 1\rfloor$. We have:
 \begin{align*}
     \sum_{h=h_0 + 1}^{h_{\max}} (h+1)^2 \exp(-h^2) & = \sum_{h=h_0 + 1}^{h_{\max}} h^2\exp(-h^2) + 2\sum_{h=h_0 + 1}^{h_{\max}} h \exp(-h^2) + \sum_{h=h_0 + 1}^{h_{\max}} \exp(-h^2)
     \\
     & \le \int_{h_0}^\infty h^2\exp(-h^2)  dh +  2 \int_{h_0}^\infty h\exp(-h^2)  dh+  \int_{h_0}^\infty \exp(-h^2)  dh.
 \end{align*}
 Using an integration by parts, the inequality $ \int_{h_0}^\infty \exp(-h_0^2)dh \le \exp(-h_0^2)/(h_0 + \sqrt{h_0^2 + 4/\pi}) \le \exp(- h_0^2)$, and $ h_0 \ge 1$ from $s_0 \ge 1$,
 we get:
 \begin{align*}
     \int_{h_0}^\infty h^2 \exp(-h^2) dh & \le  \frac{1}{2} h_0 \exp(-h_0^2) + \frac{1}{2} \exp(-h_0^2)
     \\
     2 \int_{h_0}^\infty  h \exp(-h^2) dh & =  \exp(-h_0^2)
     \\
     \int_{h_0}^\infty  \exp(-h^2) dh & \le \exp(-h_0^2).
 \end{align*}
 Therefore, 
 \begin{align*}
      \sum_{h=h_0 + 1}^{h_{\max}} (h+1)^2 \exp(-h^2) & \le \frac{1}{2}h_0 \exp(-h_0^2) + \frac{5}{2}\exp(-h_0^2)
      \\
      & \le h_0 \exp(-h_0^2) + 5\exp(-h_0^2).
 \end{align*}
 We get:
 \begin{align*}
     \sum_{h=0}^{h_0} (h + 1)^2 + \sum_{h= h_0 + 1}^ {h_{\max}}  4 s' (h+1)^2\exp\left( -h^2\right) & \le \frac{(h_0 + 1)(h_0 + 2)(2 h_0 + 3)}{6} + 4 s'  (h_0 + 5) \exp(-h_0^2) 
     \\
     &\le \frac{2h_0^3 + 9 h_0^2 + 13h_0 + 6}{6} + 4s'(h_0 + 5) \frac{1}{4s'}
     \\
     & \stackrel{(a)}{\le} 11 h_0^3,
 \end{align*}
 where for $(a)$, we used $h_0 \ge 1$. Finally, we get:
 \begin{align*}
     & r_t^\pi  \le 44 \delta^2 s_A^2  C_{\text{m}}(K -1) h_0^3 + 2 (K-1)  s_A s_1  \left(\Pr(\mathcal{E}_t^c) + \Pr\left(\left(\mathcal{G}_{t}^{\frac{\alpha}{4 \nu C_{\textnormal{b}}}}\right)^c \middle| \mathcal{E}_t \right) \right)
     \\
     & \le \frac{1408 \sigma^2 s_A^4 C_{\text{m}} (K-1) h_0^3 \nu^2 C_{\textnormal{b}}^2 \left(s_0 + \frac{4 \nu C_{\textnormal{b}}\sqrt{s_0}}{\phi_0^2}\right)}{\alpha^2} \frac{1}{t - 1} + 2 (K-1)  s_A s_1  \left(\Pr(\mathcal{E}_t^c) + \Pr\left(\left(\mathcal{G}_{t}^{\frac{\alpha}{4 \nu C_{\textnormal{b}}}}\right)^c \middle| \mathcal{E}_t\right) \right).
 \end{align*}
 This concludes the proof.\ep
 
 \subsection{Proof of Lemma~\ref{lm:inst_regret_bound_wo_margin_klarge}}
Let $r_t^\pi$ be the instantaneous expected regret of algorithm $\pi$ in round $t$ defined as:
 \begin{align*}
     r_t^\pi \coloneqq \EXP\left[ \max_{A \in \mathcal{A}_t } \langle A - A_t, \theta \rangle\right].
 \end{align*}
 Let us define the events $\mathcal{R}_k \coloneqq \{ \mathcal{A}_t \in \mathbb{R}^{K \times d}:  k \in \argmax_{k'} \langle A_{t, k'}, \theta \rangle\}$ and $ \mathcal{G}_{t}^{\lambda} \coloneqq \left\{\lambda_{\min } ( \hat{\Sigma}_{\hat{S}} ) \ge \lambda \right\}$. As in the proof of Lemma~\ref{lm:inst_regret_bound_klarge}, we get:
 \begin{align*}
     r_t^\pi & \le \sum_{k=1}^K \EXP\left[  r_t^\pi \mid  \mathcal{A}_t \in \mathcal{R}_k\right]  \Pr\left(\mathcal{A}_t \in \mathcal{R}_k\right),
 \end{align*}
 and
 \begin{align*}
     \EXP\left[  r_t^\pi \mid  \mathcal{A}_t \in \mathcal{R}_k\right]  & = \EXP\left[  \langle A_{t, k} - A_t, \theta \rangle \mid  \mathcal{A}_t \in \mathcal{R}_k\right]
    \\
     & \le \sum_{\ell \neq k} \EXP\left[  \indicator\left\{\langle A_{t, \ell}, \hat{\theta}_t \rangle \ge \langle  A_{t, k}, \hat{\theta}_t \rangle  \right\} \langle A_{t, k} - A_{t, \ell}, \theta \rangle \mid  \mathcal{A}_t \in \mathcal{R}_k, \ \mathcal{E}_t, \ \mathcal{G}_{t}^{\frac{\alpha}{4 \nu C_{\textnormal{b}}}} \right] 
     \\
     & \ \ \ + 2 (K - 1)  s_A s_1  \left(\Pr(\mathcal{E}_t^c) + \Pr\left(\left(\mathcal{G}_{t}^{\frac{\alpha}{4 \nu C_{\textnormal{b}}}}\right)^c \middle| \mathcal{E}_t\right) \right).
\end{align*} 
 Let us denote the event $I_h \coloneqq \{\mathcal{A}_t \in \mathbb{R}^{K \times d} : \langle A_{t, k} - A_{t, \ell}, \theta \rangle \in (2 \delta s_Ah, 2 \delta s_A(h + 1) ]\}$ where
 \begin{align*}
      \delta = \frac{\sigma s_A  \nu C_{\textnormal{b}}}{\alpha} \sqrt{\frac{32 \left(s_0 + \frac{4 \nu C_{\textnormal{b}} s_0 }{\phi_0^2}\right)}{t -1}}.
 \end{align*}
 By conditioning on $I_h$, we get:
 \begin{align*}
     & \EXP\left[  \indicator\left\{\langle A_{t, \ell}, \hat{\theta}_t \rangle \ge \langle  A_{t, k}, \hat{\theta}_t \rangle  \right\} \langle A_{t, k} - A_{t, \ell}, \theta \rangle \mid  \mathcal{A}_t \in \mathcal{R}_k \cap I_h, \ \mathcal{E}_t, \ \mathcal{G}_{t}^{\frac{\alpha}{4 \nu C_{\textnormal{b}}}}\right]
     \\
     & \le \sum_{h=0}^{\lceil s_1 / \delta \rceil}\EXP\left[  \indicator\left\{\langle A_{t, \ell}, \hat{\theta}_t \rangle \ge \langle  A_{t, k}, \hat{\theta}_t \rangle  \right\} \langle A_{t, k} - A_{t, \ell}, \theta \rangle \mid  \mathcal{A}_t \in \mathcal{R}_k \cap I_h, \ \mathcal{E}_t, \ \mathcal{G}_{t}^{\frac{\alpha}{4 \nu C_{\textnormal{b}}}} \right] \Pr(\mathcal{A}_t \in I_h)
     \\
     & \stackrel{(a)}{\le}  \sum_{h=0}^{\lceil s_1 / \delta \rceil} 2 \delta s_A (h + 1)\EXP\left[  \indicator\left\{\langle A_{t, \ell}, \hat{\theta}_t \rangle \ge \langle  A_{t, k}, \hat{\theta}_t \rangle \right\} \mid  \mathcal{A}_t \in \mathcal{R}_k \cap I_h, \ \mathcal{E}_t, \ \mathcal{G}_{t}^{\frac{\alpha}{4 \nu C_{\textnormal{b}}}}
     \right]\\
     &\qquad\qquad\qquad\qquad\cdot \Pr \left(\langle A_{t, k} - A_{t, \ell}, \theta \rangle  \in (0, 2 \delta s_A(h + 1) ] \right)
     \\
     & \le \sum_{h=0}^{\lceil s_1 / \delta \rceil} 2 \delta s_A (h + 1)\Pr\left(  \langle A_{t, \ell}, \hat{\theta}_t \rangle \ge \langle  A_{t, k}, \hat{\theta}_t \rangle  \mid  \mathcal{A}_t \in \mathcal{R}_k \cap I_h, \ \mathcal{E}_t, \ \mathcal{G}_{t}^{\frac{\alpha}{4 \nu C_{\textnormal{b}}}}
     \right),
 \end{align*}
 where for $(a)$, we used the definition of $I_h$. Under the event $ \mathcal{A}_t \in I_h$, the event $\langle A_{t, \ell}, \hat{\theta}_t \rangle \ge \langle  A_{t, k}, \hat{\theta}_t \rangle$ happens only when at least one of the events $ \langle A_{t, k} , \theta - \hat{\theta} \rangle \ge \delta s_A h$ or $ \langle A_{t, \ell} , \hat{\theta} - \theta \rangle \ge \delta s_A h $ holds. Therefore, 
 \begin{align*}
     & \Pr\left(  \langle A_{t, \ell}, \hat{\theta}_t \rangle \ge \langle  A_{t, k}, \hat{\theta}_t \rangle
     \mid  \mathcal{A}_t \in \mathcal{R}_k \cap I_h, \ \mathcal{E}_t, \ \mathcal{G}_{t}^{\frac{\alpha}{4 \nu C_{\textnormal{b}}}}
     \right) 
     \\
     & \le \Pr\left(  \langle A_{t, k} , \theta - \hat{\theta} \rangle \ge \delta s_A h   \mid  \mathcal{A}_t \in \mathcal{R}_k \cap I_h, \ \mathcal{E}_t, \ \mathcal{G}_{t}^{\frac{\alpha}{2\nu}}
     \right)  \\
     &\qquad\qquad\qquad + \Pr\left( \langle A_{t, \ell} , \hat{\theta} - \theta \rangle \ge \delta s_A h  \mid \mathcal{A}_t \in \mathcal{R}_k \cap I_h, \ \mathcal{E}_t, \ \mathcal{G}_{t}^{\frac{\alpha}{4 \nu C_{\textnormal{b}}}}
     \right) 
     \\
     & \stackrel{(a)}{\le} \Pr\left(   \| \theta - \hat{\theta}\|_2 \ge \delta  h   \mid  \mathcal{A}_t \in \mathcal{R}_k \cap I_h, \ \mathcal{E}_t, \ \mathcal{G}_{t}^{\frac{\alpha}{4 \nu C_{\textnormal{b}}}}
     \right)  + \Pr\left(  \| \theta - \hat{\theta}\|_2 \ge \delta  h  \mid  \mathcal{A}_t \in \mathcal{R}_k \cap I_h, \ \mathcal{E}_t, \ \mathcal{G}_{t}^{\frac{\alpha}{4 \nu C_{\textnormal{b}}}}
     \right) 
     \\
     & = 2 \Pr\left(   \| \theta - \hat{\theta}\|_2 \ge \delta  h  \mid  \mathcal{A}_t \in \mathcal{R}_k \cap I_h, \ \mathcal{E}_t, \ \mathcal{G}_{t}^{\frac{\alpha}{4 \nu C_{\textnormal{b}}}}
     \right),
 \end{align*}
 where for $(a)$, we used the Cauchy–Schwarz inequality. Let us denote $s' = s_0 + \frac{4 \nu C_{\textnormal{b}}\sqrt{s_0}}{\phi_0^2}$. Then, using Lemma~\ref{lm:LS_estim_risk_klarge}, we get:
 \begin{align*}
     \Pr\left(   \| \theta - \hat{\theta}\|_2 \ge \delta  h   \mid  \mathcal{A}_t \in \mathcal{R}_k \cap I_h, \ \mathcal{E}_t, \ \mathcal{G}_{t}^{\frac{\alpha}{4\nu C_{\textnormal{b}}}}
     \right) & \le 2 s'\exp \left( -\frac{\alpha^2 t \delta^2 h^2}{32 \sigma^2 s_A^2 \nu^2 C_{\textnormal{b}}^2 s'} \right)
     \\
     & = 2 s'\exp \left( -h^2\right).
 \end{align*}
 We also trivially have that:
 \begin{align*}
     \Pr\left(   \| \theta - \hat{\theta}\|_2 \ge \delta  h  \mid  \mathcal{A}_t \in \mathcal{R}_k \cap I_h, \ \mathcal{E}_t, \ \mathcal{G}_{t}^{\frac{\alpha}{4\nu C_{\textnormal{b}}}}
     \right) & \le 1.
 \end{align*}
 Therefore, we can upper bound the expected instantaneous regret as:
 \begin{align*}
     r_t^\pi & \le \sum_{k=1}^K \EXP\left[  r_t^\pi \mid  \mathcal{A}_t \in \mathcal{R}_k\right]  \Pr\left(\mathcal{A}_t \in \mathcal{R}_k\right)
     \\
     & \le \sum_{k=1}^K \left(\sum_{\ell \neq k}\left( \sum_{h=0}^{\lceil s_1/\delta \rceil} \left(2 \delta s_A (h + 1) \min \left\{ 1, 4s' \exp\left( -h^2\right) \right\} \right) \right) \right.
     \\
     & \ \ \ \left.+ 2 (K-1) s_A s_1  \left(\Pr(\mathcal{E}_t^c) + \Pr\left(\left(\mathcal{G}_{t}^{\frac{\alpha}{4\nu C_{\textnormal{b}}}}\right)^c\right) \right) \cdot \Pr\left(\mathcal{A}_t \in \mathcal{R}_k \right)\right)  
     \\
     & \stackrel{(a)}{\le} 2 \delta s_A (K-1) \sum_{h=0}^{\lceil s_1/\delta \rceil} (h + 1)\min \left\{ 1, 4 s' \exp\left( -h^2\right) \right\} \\
     &\qquad\qquad\qquad\qquad\qquad+ 2 (K-1)  s_A s_1  \left(\Pr(\mathcal{E}_t^c) + \Pr\left(\left(\mathcal{G}_{t}^{\frac{\alpha}{4\nu C_{\textnormal{b}}}}\right)^c \middle| \mathcal{E}_t\right) \right)
     \\
     & \le 2 \delta s_A (K-1) \left( \sum_{h=0}^{h_0} (h + 1) + \sum_{h= h_0 + 1}^ {h_{\max}}  4 s' (h+1)\exp\left( -h^2\right) \right)  \\
     &\qquad\qquad\qquad\qquad\qquad + 2 (K-1)  s_A s_1  \left(\Pr(\mathcal{E}_t^c) + \Pr\left(\left(\mathcal{G}_{t}^{\frac{\alpha}{4\nu C_{\textnormal{b}}}}\right)^c \middle| \mathcal{E}_t\right) \right)
 \end{align*}
 where for $(a)$, we used $\sum_{k=1}^K \Pr\left(\mathcal{A}_t \in \mathcal{R}_k\right)= 1$ and we set $ h_0 \coloneqq \lfloor \sqrt{\log 4 s'} + 1\rfloor$. We have:
 \begin{align*}
     \sum_{h=h_0 + 1}^{h_{\max}} (h+1) \exp(-h^2) & = \sum_{h=h_0 + 1}^{h_{\max}} h \exp(-h^2) + \sum_{h=h_0 + 1}^{h_{\max}} \exp(-h^2)
     \\
     & \le \int_{h_0}^\infty h\exp(-h^2)  dh+  \int_{h_0}^\infty \exp(-h^2)  dh.
 \end{align*}
 Since $ h_0 \ge 1$ from $s_0 \ge 1$, we get:
 \begin{align*}
     \int_{h_0}^\infty  h \exp(-h^2) dh & =  \frac{1}{2}\exp(-h_0^2)
     \\
     \int_{h_0}^\infty  \exp(-h^2) dh & \le \exp(-h_0^2).
 \end{align*}
 Therefore, 
 \begin{align*}
      \sum_{h=h_0 + 1}^{h_{\max}} (h+1) \exp(-h^2) & \le \frac{3}{2}\exp(-h_0^2).
 \end{align*}
 We get:
 \begin{align*}
     \sum_{h=0}^{h_0} (h + 1) + \sum_{h= h_0 + 1}^ {h_{\max}}  4 s' (h+1)\exp\left( -h^2\right) & \le \frac{(h_0 + 1)(h_0 + 2)}{2} + 4 s' \frac{3}{2}\exp(-h_0^2) 
     \\
     &\le \frac{h_0^2 + 3h_0 + 2}{2} + \frac{6s'}{4s'}
     \\
     & \stackrel{(a)}{\le} \frac{9}{2}h_0^2,
 \end{align*}
 where for $(a)$, we used $h_0 \ge 1$. Finally, we get:
 \begin{align*}
     r_t^\pi & \le 9 \delta s_A (K-1) h_0^2 + 2 (K-1)  s_A s_1  \left(\Pr(\mathcal{E}_t^c) + \Pr\left(\left(\mathcal{G}_{t}^{\frac{\alpha}{4\nu C_{\textnormal{b}}}}\right)^c \middle| \mathcal{E}_t \right) \right)
     \\
     & \le \frac{36 \sigma s_A  (K-1) h_0^2 \nu C_{\textnormal{b}}}{\alpha}\sqrt{\frac{2 \left(s_0 + \frac{4 \nu C_{\textnormal{b}}\sqrt{s_0} }{\phi_0^2}\right)}{t-1}} + 2  (K-1) s_A s_1  \left(\Pr(\mathcal{E}_t^c) + \Pr\left(\left(\mathcal{G}_{t}^{\frac{\alpha}{4\nu C_{\textnormal{b}}}}\right)^c \middle| \mathcal{E}_t\right) \right).
 \end{align*}
 This concludes the proof.\ep

\section{Proof of Lemmas (without balanced covariance)}
\label{sec:proof_lemmas_k2}

\begin{lemma}\label{lm:support_recovery_k2}
	Let $t \ge  \frac{2 \log (2 d^2)}{C_0^2}$ such that $4 \left( \frac{2 \nu  s_0}{\phi_0^2} + \sqrt{\left(1 + \frac{2\nu }{\phi_0^2}\right)s_0}\right)\lambda_t \le \theta_{\min}$. Under Assumptions \ref{asm:sparsity_param_klarge}, \ref{asm:comp_cond}, \ref{asm:relax_sym}, and \ref{asm:balanced_cov},\\ 
	$
	\Pr \left(S(\theta) \subset \hat{S}_1^{(t)} \text{ and } |\hat{S}_1^{(t)} \setminus S(\theta)| \le \frac{2\nu  \sqrt{s_0}}{\phi^2_0}\right) \ge 1 -  2 \exp\left( -\frac{t \lambda_t^2}{32 \sigma^2 s_A^2} + \log d\right) - \exp\left( - \frac{t C_0^2}{2}\right).$
\end{lemma}

We redefine the event $\mathcal{E}_t$ as
\begin{align*}
    \mathcal{E}_t = \left\{S \subset \hat{S}_1^{(t)}  \text{ and } |\hat{S}_{1}^{(t)} \setminus S| \le  \frac{2 \nu  \sqrt{s_0}}{\phi^2_0} \right\}.
\end{align*}

\begin{lemma}\label{lm:lowerbound_eig_S_k2}
	Let $t \in [T]$. Under Assumptions \ref{asm:sparsity_param_klarge} and \ref{asm:cov_div}, we have:
	\begin{align*}
		&\Pr\left(\lambda_{\min} (\hat{\Sigma}_{\hat{S}}) \ge   \frac{\alpha}{2\nu } \ \big| \ \mathcal{E}_t \right)  \ge 1 -  \exp\left( \log\left(s_0 + \frac{2 \nu \sqrt{s_0}}{\phi_0^2} \right) - \frac{t \alpha}{10 s_A^2 \nu\left(s_0 + (2 \nu \sqrt{s_0})/\phi_0^2\right) }\right). 
	\end{align*}
\end{lemma}

\begin{lemma}\label{lm:LS_estim_risk_k2}
	Let $t \in [T]$ and $s' = s_0 + 2 \nu  \sqrt{s_0}/\phi_0^2$. Under Assumption~\ref{asm:sparsity_param_klarge}, we have for all $x, \lambda >0$:
	\begin{align*}
		\Pr \left( \|\hat{\theta}_{t+1} - \theta \|_2 \ge x \; \text{and} \;\lambda_{\min}(\hat{\Sigma}_{\hat{S}}) \ge \lambda \ \big| \ \mathcal{E}_t \right)  \
		\le  2 s' \exp \left( - \frac{\lambda^2 t x^2}{2 \sigma^2 s_A^2 s'}\right).
	\end{align*}
\end{lemma}

We redefine the parameter $h_0 = \lfloor \sqrt{\log(4 (s_0 + \frac{2 \nu \sqrt{s_0}}{\phi_0^2}))}  + 1 \rfloor$.
\begin{lemma}\label{lm:inst_regret_bound_k2}
	Define $\mathcal{G}_{t}^{\frac{\alpha}{2\nu }} \coloneqq \left\{\lambda_{\min } ( \hat{\Sigma}_{\hat{S}} ) \ge \frac{\alpha}{2\nu } \right\}$. Let $t \ge 2$. Under Assumptions \ref{asm:sparsity_param_klarge}, \ref{asm:comp_cond}, \ref{asm:relax_sym}, \ref{asm:balanced_cov}, \ref{asm:margin}, and  \ref{asm:cov_div}, the expected instantaneous regret $\EXP[\max_{A \in \mathcal{A}_t} \langle A - A_t, \theta \rangle] $ is upper bounded by:
	$$
  \frac{352 \sigma^2 s_A^4 C_{\textnormal{m}}  h_0^3 \nu^2 \left(s_0 + \frac{2 \nu\sqrt{s_0}}{\phi_0^2}\right)}{\alpha^2} \frac{1}{t - 1} + 2  s_A s_1  \left(\Pr(\mathcal{E}_t^c) + \Pr\left(\left(\mathcal{G}_{t}^{\frac{\alpha}{2\nu}}\right)^c \middle| \mathcal{E}_t\right) \right).
	$$
\end{lemma}

\begin{lemma}\label{lm:inst_regret_bound_wo_margin_k2}
	Under Assumptions \ref{asm:sparsity_param_klarge},  \ref{asm:comp_cond},  \ref{asm:relax_sym}, \ref{asm:balanced_cov}, and \ref{asm:cov_div}, for any $t \in [T]$, $\EXP[\max_{A \in \mathcal{A}_t} \langle A - A_t, \theta\rangle]$ is upper bounded by:
	$$
	\frac{18 \sigma s_A   h_0^2 \nu }{\alpha}\sqrt{\frac{2 \left(s_0 + \frac{2 \nu \sqrt{s_0} }{\phi_0^2}\right)}{t-1}} + 2   s_A s_1  \left(\Pr(\mathcal{E}_t^c) + \Pr\left(\left(\mathcal{G}_{t}^{\frac{\alpha}{2\nu }}\right)^c \middle| \mathcal{E}_t\right) \right).
	$$
\end{lemma}

\subsection{Proof of Lemma~\ref{lm:support_recovery_k2}}
For the sake of brevity, let $S= S(\theta)$.  We define $v \coloneqq \hat{\theta}_{0}^{(t)} - \theta$. We first analyze the performance of the initial Lasso estimate.

\begin{lemma}\label{lm:adaptive_lasso}
Let $\hat{\Sigma}_t \coloneqq \frac{\sum_{s=1}^t A_s A_s^\top}{t}$ be the empirical covariance matrix of the selected context vectors. Suppose $\hat{\Sigma}_t$ satisfies the compatibility condition with the support $S$ with the compatibility constant $\phi_t$. Then, under Assumption~\ref{asm:sparsity_param_klarge}, we have:
     \begin{align*}
        \Pr\left(\| v\|_1 \le \frac{4 s_0 \lambda_t}{\phi_t^2}\right) \ge 1 - 2 \exp\left( - \frac{t \lambda_t^2}{32 \sigma^2 s_A^2} + \log d\right).
    \end{align*} 
\end{lemma}
The next lemma then states that the compatibility constant of $\hat{\Sigma}_t$ does not deviate much  from  the compatibility constant of $\Sigma$.

\begin{lemma}\label{lm:compatibility_holds}
    Assume $K=2$. Let $C_0 \coloneqq \min\left\{ \frac{1}{2}, \frac{\phi_0^2}{256 s_0 s_A^2 \nu }\right\}$. For all $t \ge \frac{2 \log (2 d^2)}{C_0^2}$, we have:
    \begin{align*}
        \Pr\left( \phi^2(\hat{\Sigma}_t, S) \ge \frac{\phi^2_0}{2\nu}\right) \ge 1 - \exp\left( - \frac{t C_0^2}{2}\right).
    \end{align*}
\end{lemma}
Then, we follow the steps of the proof given by \citet{zhou2010thresholded}. Let us define the event $\mathcal{G}_t$ as:
\begin{align*}
    \mathcal{G}_t \coloneqq \left\{\| v\|_1 \le \frac{4 s_0 \lambda_t}{\phi_t^2} \right\}.
\end{align*}
For the rest of this section, we assume that the event $\mathcal{G}_t$ holds. 
Note that:
\begin{align*}
    \|v\|_1 & \ge \|v_{S^c}\|_1
    \\
    & = \sum_{j \in S^c} |(\hat{\theta}_0^{(t)})_j| 
    \\
    & \ge \sum_{j \in S^c \cap \hat{S}_0^{(t)}}|(\hat{\theta}_0^{(t)})_j|
    \\
    & = \sum_{j \in  \hat{S}_0^{(t)}  \setminus S}|(\hat{\theta}_0^{(t)})_j|
    \\
    & \stackrel{(a)}{\ge} | \hat{S}_0^{(t)}  \setminus S|  4 \lambda_t,
\end{align*}
where for $(a)$, we used the construction of $ \hat{S}_0^{(t)}$ in the algorithm. We get:
\begin{align*}
    | \hat{S}_0^{(t)}  \setminus S| & \le \frac{\|v\|_1}{4 \lambda_t}
    \stackrel{(a)}{\le} \frac{s_0}{\phi_t^2},
\end{align*}
where for $(a)$, we used the definition of $\mathcal{G}_t$. 
We have: $\forall j \in S$, 
\begin{align*}
    |(\hat{\theta}_{0}^{(t)})_j| & \ge \theta_{\min}  - \|v_S\|_\infty
    \\
    & \ge \theta_{\min}  - \|v_S\|_1
    \\
    & \ge \theta_{\min}  - \frac{4 s_0 \lambda_t}{\phi_t^2}.
\end{align*}
Therefore, when $t$ is large enough so that $ 4 \lambda_t \le \theta_{\min}  - \frac{4 s_0 \lambda_t}{\phi_t^2}$, we have: $S \subset \hat{S}_0^{(t)}$. 
Using a similar argument, when $t $ is large enough so that  $ 4 \lambda_t \sqrt{\left( 1 + \frac{1}{\phi_t^2}\right)s_0} \le \theta_{\min} - \frac{4 s_0 \lambda_t}{\phi_t^2}$, it holds that $S \subset \hat{S}_1^{(t)}$.
From the construction of $\hat{S}_1^{(t)}$ in the algorithm, it also holds that: $\hat{S}_1^{(t)} \subset \hat{S}_0^{(t)}$. Therefore,
\begin{align*}
    \|v\|_1 & \ge \sum_{i \in \hat{S}_0^{(t)} \setminus S} | (\hat{\theta}_0^{(t)})_i|
    \\
    & \ge \sum_{i \in \hat{S}_1^{(t)} \setminus S} | (\hat{\theta}_0^{(t)})_i|
    \\
    & \ge |\hat{S}_1^{(t)} \setminus S| 4 \lambda_t \sqrt{|\hat{S}_0^{(t)}|},
\end{align*}
and 
\begin{align*}
    |\hat{S}_1^{(t)} \setminus S| & \le \frac{\|v\|_1}{4 \lambda_t \sqrt{|\hat{S}_0^{(t)}|}}
    \\
    & \le \frac{1}{4 \lambda_t \sqrt{|\hat{S}_0^{(t)}|}} \cdot \frac{4 s_0 \lambda_t}{\phi_t^2}
    \\
    & \le \frac{\sqrt{s_0}}{\phi_t^2}.
\end{align*}
Note that the condition $ 4 \lambda_t \sqrt{\left( 1 + \frac{1}{\phi_t^2}\right)s_0} \le \theta_{\min} - \frac{4 s_0 \lambda_t}{\phi_t^2}$ is equivalent to $ 4 \lambda_t \left(  \sqrt{\left( 1 + \frac{1}{\phi_t^2}\right)s_0} + \frac{s_0 }{\phi_t^2}\right) \le \theta_{\min}$.
This concludes the proof of Lemma~\ref{lm:support_recovery_k2} by substituting $\phi_t^2 = \phi_0^2/(2\nu)$.\ep

\subsection{Proof of Lemmas used in the proof of Lemma~\ref{lm:support_recovery_k2}}
\subsubsection{Proof of Lemma~\ref{lm:adaptive_lasso}}
The proof is identical to that of Lemma~\ref{lm:adaptive_lasso_klarge}.

 \subsubsection{Proof of Lemma~\ref{lm:compatibility_holds}}
For the sake of brevity, let $S= S(\theta)$. First, we define the adapted Gram matrix $\Sigma_t \coloneqq \frac{1}{t}\sum_{s=1}^t\EXP[A_sA_s^\top | \mathcal{F}_{s-1}]$. 
From  the construction of the algorithm, $ \EXP[A_s A_s^\top | \mathcal{F}_{s-1}] = \EXP[\sum_{k=1}^K A_{s, k} A_{s, k}^\top \indicator\{k = \argmax_{k'} \langle A_{s,k}, \hat{\theta}_s\rangle\} | \; \hat{\theta}_s]$. 
The following lemma characterizes the expected Gram matrix generated by the algorithm.

\begin{lemma}\label{lem:greedy_compat}
    Assume $K=2$. Under Assumption~\ref{asm:comp_cond} and Assumption~\ref{asm:relax_sym}, for each fixed vector $\theta' \in \mathbb{R}^d$, we have:
    \begin{align*}
        \EXP_{\mathcal{A} \sim p_A}\left[\sum_{k=1,2} A_{k} A_{k}^\top \indicator\{k = \argmax_{k'} \langle A_{k}, \theta'\rangle\}\right]  \succeq \frac{1}{\nu} \Sigma,
    \end{align*}
    where $A\succeq B$ means that $A - B$  is positive semidefinite.
\end{lemma}
Using Lemma~\ref{lem:greedy_compat}, we have
\begin{align}
    \Sigma_t \succeq \frac{1}{\nu} \Sigma. 
\end{align}
By Lemma~6.18 of \citet{buhlmann2011statistics},  Assumption~\ref{asm:comp_cond}, and the definition of the compatibility constant, we get:
\begin{align}
     \phi^2(\Sigma_t, S) \ge  \phi^2(\frac{1}{\nu}\Sigma, S) \ge \frac{\phi_0^2}{\nu}.
\end{align}
Furthermore, we have a following adaptive matrix concentration results for $\hat{\Sigma}_t$:
\begin{lemma}\label{lm:matrix_concentration} Let $C_0 \coloneqq \min\left\{\frac{1}{2}, \frac{\phi_0^2}{256 s_0 s_A^2 \nu }\right\} $. 
We have, for all $t \ge \frac{2 \log(2 d^2)}{C_0^2}$,
\begin{align*}
    \Pr\left(\frac{1}{2s_A^2}\|\hat{\Sigma}_t - \Sigma_t\|_\infty \ge \frac{\phi^2(\Sigma_t, S)}{64 s_0 s_A^2 \nu }\right) \le \exp\left( - \frac{t C_0^2 }{2}\right).
\end{align*}
 \end{lemma}
Combining  Lemmas~\ref{lm:matrix_concentration} and \ref{buhlmann2011_lemma}, we get, for all $t \ge \frac{2 \log (2 d^2)}{C_0^2}$:
\begin{align*}
    \phi^2(\hat{\Sigma}_t, S) & \ge \frac{\phi^2({\Sigma}_t, S) }{2}
    \\
    & \ge \frac{\phi^2_0}{2 \nu },
\end{align*}
with probability at least $1 - \exp\left( - \frac{t C_0^2}{2}\right)$. This concludes the proof.\ep

 \subsubsection{Proof of Lemma~\ref{lem:greedy_compat}}
 The proof is almost identical to the proof of Lemma~2 in \citet{oh2020sparsity}. \ep
 
 \subsubsection{Proof of Lemma~\ref{lm:matrix_concentration}}
 
 Let us define $\gamma^{i j}_t(A_t)$ as:
 \begin{align*}
     \gamma^{i j}_t(A_t) \coloneqq \frac{1}{2 C^2_A} \left( (A_t)_i (A_t)_j - \EXP[ (A_t)_i (A_t)_j \;|\; \mathcal{F}_{t-1}]\right),
 \end{align*}
 where $(A_t)_i$ is the $i$-th element of $A_t$. 
 Note that $\frac{1}{2 s_A^2} \|\hat{\Sigma}_t - {\Sigma}_t \|_\infty = \max_{1 \le i \le j \le d} \left|\frac{1}{t} \sum_{s = 1}^t \gamma^{i j}_s(A_s) \right| $, $ \EXP[\gamma^{i j}_t(A_t) | \mathcal{F}_{t-1}] = 0$, and $\EXP[| \gamma^{i j}_t(A_t)|^m\; | \;\mathcal{F}_{t-1} ] \le 1$ for all integer $m \ge 2$. Therefore, we can apply Lemma~\ref{lm:bernstein-like2}:
 \begin{align*}
     \Pr \left(\frac{1}{2 s_A^2} \|\hat{\Sigma}_t - {\Sigma}_t \|_\infty \ge x + \sqrt{2 x} + \sqrt{\frac{4 \log (2 d^2)}{t}} + \frac{2 \log (2 d^2)}{t}\right) \le \exp\left(- \frac{tx}{2}\right).
 \end{align*}
 For all $t \ge \frac{2 \log (2 d^2)}{C_0^2}$ with $C_0 \coloneqq \min\left\{\frac{1}{2}, \frac{\phi_0^2}{256 s_0 s_A^2 \nu }\right\} $, taking $x = C_0^2$, 
 \begin{align*}
      x + \sqrt{2 x} + \sqrt{\frac{4 \log (2 d^2)}{t}} + \frac{2 \log (2 d^2)}{t} & \le 2 C_0^2 + 2 \sqrt{2}C_0
      \\
      & \le 4 C_0
      \\
      & \le \frac{\phi_0^2}{64 s_0 s_A^2 \nu}
      \\
      & \le \frac{\phi^2(\Sigma_t, S)}{64 s_0 s_A^2 \nu }\;.
 \end{align*}
 In summary, for all $ t \ge \frac{2 \log (2 d^2)}{C_0^2}$, we get:
 \begin{align*}
      \Pr \left(\frac{1}{2 s_A^2} \|\hat{\Sigma}_t - {\Sigma}_t \|_\infty \ge \frac{\phi^2(\Sigma_t, S)}{64 s_0 s_A^2 \nu }\right) & \le \Pr \left(\frac{1}{2 s_A^2} \|\hat{\Sigma}_t - {\Sigma}_t \|_\infty \ge C_0^2 + \sqrt{2} C_0 + \sqrt{\frac{4 \log (2 d^2)}{t}} + \frac{2 \log (2 d^2)}{t}\right) 
      \\
      & \le  \exp\left( - \frac{t C_0^2}{2}\right).
 \end{align*}
 This concludes the proof. \ep 
 
 \subsection{Proof of Lemma~\ref{lm:lowerbound_eig_S_k2}}
 
 For a fixed $\hat{S}$, first we define the adapted Gram matrix on the estimated support as 
 $$
 \Sigma_t \coloneqq \frac{1}{t}\sum_{s=1}^t\EXP[A_s(\hat{S})A_s(\hat{S})^\top | \mathcal{F}_{s-1}].
 $$
From  the construction of the algorithm, $ \EXP[A_s(\hat{S}) A_s(\hat{S})^\top | \mathcal{F}_{s-1}] = \EXP[\sum_{k=1}^K A_{s, k}(\hat{S}) A_{s, k}(\hat{S})^\top \indicator\{k = \argmax_{k'} \langle A_{s,k}, \hat{\theta}_s\rangle\} | \; \hat{\theta}_s]$. 
Recall that for each $B \subset [d]$, $\Sigma_B \coloneqq \frac{1}{K} \sum_{k=1}^K \EXP_{\mathcal{A}\sim p_A} \left[A_{k}(B) A_{k}(B)^\top\right]$, where $ A_{k}(B)$ is a $|B|$-dimensional vector extracted the elements of $A_{k}$ with indices in $B$. 
The following lemma characterizes the expected Gram matrix generated by the algorithm.

\begin{lemma}\label{lem:greedy_positivedefnite}
    Assume $K=2$. Fix $\hat{S} $ such that $ S(\theta) \subset \hat{S} $ and $|\hat{S}| \le s_0 + (2 \nu \sqrt{s_0})/\phi_0^2$. Fix $\theta' \in \mathbb{R}^d$. Under Assumption~\ref{asm:comp_cond} and Assumption~\ref{asm:relax_sym}, we have:
    \begin{align*}
        \EXP_{\mathcal{A} \sim p_A}\left[\sum_{k=1,2} A_{k}(\hat{S}) A_{k}(\hat{S})^\top \indicator\{k = \argmax_{k'} \langle A_{k}, \theta_{\hat{S}}'\rangle\}\right]  \succeq \frac{1}{\nu} \Sigma_{\hat{S}},
    \end{align*}
    where $A\succeq B$ means that $A - B$  is positive semidefinite.
\end{lemma}
 First, we prove the lower bound on the smallest eigenvalue of the expected covariance matrices. Let $\Sigma_{\hat{S}} \coloneqq \frac{1}{t} \sum_{s =1}^t \EXP[A_{s}( \hat{S}) A_{s}( \hat{S})^\top \mid \mathcal{F}_{t-1}]$. By Assumption~\ref{asm:cov_div} and the construction of the algorithm, under the event $\mathcal{E}_t$, we get:
 \begin{align*}
     \lambda_{\min} (\Sigma_{\hat{S}}) & =\lambda_{\min}\left(  \frac{1}{t} \sum_{s =1}^t \EXP \left[ \sum_{k =1}^K A_{s, k, \hat{S}} A_{s, k, \hat{S}} \indicator \{ k = \argmax_{k'} \langle A_{k'}, \hat{\theta}_s \rangle\} \mid \hat{\theta}_s \right]  \right)
     \\
     & \ge \sum_{s=1}^t\lambda_{\min} \left(\frac{1}{t} \EXP \left[ \sum_{k =1}^K A_{s, k, \hat{S}} A_{s, k, \hat{S}} \indicator \{ k = \argmax_{k'} \langle A_{k'}, \hat{\theta}_s \rangle\} \mid \hat{\theta}_s \right]\right)
     \\
     & = \frac{\alpha}{\nu},
 \end{align*}
 where for the inequality, we used the concavity of $\lambda_{\min} ( \cdot)$ over the positive semi-definite matrices. Next, we prove the upper bound on the largest eigenvalue of $ A_{s}( \hat{S}) A_{s}( \hat{S})^\top$:
 \begin{align*}
     \lambda_{\max} (A_{s}( \hat{S}) A_{s}( \hat{S})^\top)  
     & = \max_{\|v\| = 1} v^\top A_{s}( \hat{S}) A_{s}( \hat{S})\top v  
      \\
      & \stackrel{(a)}{\le} \max_{\|v\| = 1}\|v\|_1^2 \|A_{s}( \hat{S})\|_\infty^2
      \\
      & \le |\hat{S}|s_A^2
      \\
      & \le  \left(s_0 + (2 \nu \sqrt{s_0})/\phi_0^2\right)s_A^2
 \end{align*}
 where for $(a)$, we used H\"older's inequality and Assumption~\ref{asm:sparsity_param_klarge}. 
Taking $R = \left(s_0 + (2 \nu \sqrt{s_0})/\phi_0^2\right)s_A^2$, $X_s = A_{s, \hat{S}} A_{s, \hat{S}}^\top$, $Y = t \hat{\Sigma}_{\hat{S}}$, $W = t {\Sigma}_{\hat{S}}$, $\delta = 1/2$, $ \mu = t \frac{\alpha}{\nu}$ in Theorem~\ref{lem:tropp2011}, we have: 
\begin{align*}
     \Pr\left( \lambda_{\min} (t \hat{\Sigma}_{\hat{S}}) \le \frac{1}{2} t \frac{\alpha}{\nu} \text{ and } \lambda_{\min}(  t {\Sigma}_{\hat{S}}) \ge t \frac{\alpha}{\nu}  \right) & \le  \left(s_0 + \frac{2 \nu \sqrt{s_0}}{\phi_0^2} \right)\left( \frac{e^{ - 0.5}}{0.5^{0.5}}\right)^{\frac{t \alpha}{R \nu }}
     \\
     & \le \exp\left( \log\left(s_0 + \frac{2 \nu \sqrt{s_0}}{\phi_0^2} \right) - \frac{t \alpha}{10 s_A^2 \nu  \left(s_0 + (2 \nu \sqrt{s_0})/\phi_0^2\right)}\right),
\end{align*}
where for the last inequality, we used $ -0.5 - 0.5 \log(0.5) < - \frac{1}{10}$.  This concludes the proof.\ep

\subsubsection{Proof of Lemma \ref{lem:greedy_positivedefnite}}
 The proof is almost identical to the proof of Lemma~2 in \citet{oh2020sparsity}. \ep

 \subsection{Proof of Lemma~\ref{lm:LS_estim_risk_k2}}
In this proof, we denote $\hat{S} = \hat{S}_{1}^{(t)}$ and $\varepsilon = (\varepsilon_1, \ldots, \varepsilon_t)^\top$.
Assume $\lambda_{\min} (\hat{\Sigma}_{\hat{S}})\ge \lambda$. We have:
\begin{align*}
    \|\hat{\theta}_{t+1} - \theta\|_2 & = \| (A(\hat{S})^\top A(\hat{S}))^{-1} A(\hat{S})^\top R - \theta\|_2
    \\
    & = \| (A(\hat{S})^\top A(\hat{S}))^{-1} A(\hat{S})^\top (A \theta + \varepsilon) -\theta\|_2
    \\
    & = \| (A(\hat{S})^\top A(\hat{S}))^{-1} A(\hat{S})^\top (A(\hat{S}) \theta({\hat{S}}) + \varepsilon) -\theta\|_2
    \\
    & = \| (A(\hat{S})^\top A(\hat{S}))^{-1} A(\hat{S})^\top \varepsilon \|_2
    \\
    & \le \| (A(\hat{S})^\top A(\hat{S}))^{-1} \|_2 \|A(\hat{S})^\top \varepsilon \|_2
    \\
    & \le \frac{1}{\lambda t} \|A (\hat{S})^\top \varepsilon \|_2.
\end{align*}
We get (note that we are conditioning on a fixed $\hat{S}$ during the proof):
\begin{align*}
    \Pr \left( \|\hat{\theta}_{t+1} - \theta \|_2 \ge x \; \text{and} \; \lambda_{\min}(\hat{\Sigma}_{\hat{S}}) \ge \lambda \right) & =  \Pr \left( \|\hat{\theta}_{t+1} - \theta \|_2 \ge x \; \middle\vert \; \lambda_{\min}( \hat{\Sigma}_{\hat{S}}) \ge \lambda \right)\Pr(\lambda_{\min}( \hat{\Sigma}_{\hat{S}}) \ge \lambda)
    \\
    & \le \Pr \left( \|A (\hat{S})^\top \varepsilon \|_2  \ge \lambda t x \; \middle\vert \; \lambda_{\min}( \hat{\Sigma}_{\hat{S}}) \ge \lambda \right)\Pr(\lambda_{\min}( \hat{\Sigma}_{\hat{S}}) \ge \lambda)
    \\
    & \le \Pr \left( \|A (\hat{S})^\top \varepsilon \|_2  \ge \lambda t x \right)
    \\
    & \le \sum_{i = 1}^d \Pr \left(\left| \sum_{s=1}^t \varepsilon_s (A_{s})_i \indicator \left\{ i \in \hat{S}\right\}\right|\ge \frac{\lambda t x}{\sqrt{s_0 + \frac{2 \nu \sqrt{s_0}}{\phi_0^2}}} \right)
    \\
    & = \sum_{i \in \hat{S}} \Pr \left(\left| \sum_{s=1}^t \varepsilon_s  (A_{s})_i \right|\ge \frac{\lambda t x}{\sqrt{s_0 + \frac{2 \nu \sqrt{s_0}}{\phi_0^2}}} \right)
    \\
    & \stackrel{(a)}{\le} 2 \left( s_0 + \frac{2 \nu \sqrt{s_0}}{\phi_0^2}\right) \exp \left( - \frac{\lambda^2 t x^2}{2 \sigma^2 s_A^2 \left( s_0 + \frac{2 \nu  \sqrt{s_0}}{\phi_0^2}\right)}\right),
\end{align*}
where for $(a)$, we used Theorem~\ref{thm:martingale_bernstein}. This concludes the proof.\ep

 \subsection{Proof of Lemma~\ref{lm:inst_regret_bound_k2}}
 We follow the proof strategy of Lemma~6 in \citet{bastani2021mostly}. Let $r_t^\pi$ be the instantaneous expected regret of algorithm $\pi$ at round $t$ defined as:
 \begin{align*}
     r_t^\pi \coloneqq \EXP\left[ \max_{A \in \mathcal{A}_t } \langle A - A_t, \theta \rangle\right].
 \end{align*}
 Let us define the events $\mathcal{R}_k \coloneqq \{ \mathcal{A}_t \in \mathbb{R}^{K \times d}:  k \in \argmax_{k'} \langle A_{t, k'}, \theta \rangle\}$ and $ \mathcal{G}_{t}^{\lambda} \coloneqq \left\{\lambda_{\min } ( \hat{\Sigma}_{\hat{S}} ) \ge \lambda \right\}$. We have:
 \begin{align*}
     r_t^\pi & \le \sum_{k=1}^2 \EXP\left[  r_t^\pi \mid  \mathcal{A}_t \in \mathcal{R}_k\right]  \Pr\left(\mathcal{A}_t \in \mathcal{R}_k\right).
 \end{align*}
 The term $\EXP\left[  r_t^\pi \mid  \mathcal{A}_t \in \mathcal{R}_k\right] $ can be further computed as:
 \begin{align*}
     \EXP\left[  r_t^\pi \mid  \mathcal{A}_t \in \mathcal{R}_k\right]  & = \EXP\left[  \langle A_{t, k} - A_t, \theta \rangle \mid  \mathcal{A}_t \in \mathcal{R}_k\right]
     \\
     & \le \EXP\left[  \indicator\left\{ \langle  A_t, \hat{\theta}_t \rangle  \ge \langle A_{t, k}, \hat{\theta}_t \rangle\right\} \langle A_{t, k} - A_t, \theta \rangle \mid  \mathcal{A}_t \in \mathcal{R}_k\right]
     \\
     & \le \sum_{\ell \neq k} \EXP\left[  \indicator\left\{\langle A_{t, \ell}, \hat{\theta}_t \rangle \ge \langle  A_{t, k}, \hat{\theta}_t \rangle \right\} \langle A_{t, k} - A_{t, \ell}, \theta \rangle \mid  \mathcal{A}_t \in \mathcal{R}_k\right]
     \\
     & \le \sum_{\ell \neq k} \EXP\left[  \indicator\left\{\langle A_{t, \ell}, \hat{\theta}_t \rangle \ge \langle  A_{t, k}, \hat{\theta}_t \rangle \right\} \langle A_{t, k} - A_{t, \ell}, \theta \rangle \mid  \mathcal{A}_t \in \mathcal{R}_k, \ \mathcal{E}_t, \  \mathcal{G}_{t}^{\frac{\alpha}{2\nu}}\right] 
     \\
     & \ \ \ + 2  s_A s_1  \left(\Pr(\mathcal{E}_t^c) + \Pr\left(\left(\mathcal{G}_{t}^{\frac{\alpha}{2\nu}}\right)^c \middle| \mathcal{E}_t\right) \right).
\end{align*}
 Let us denote the event $I_h \coloneqq \{\mathcal{A}_t \in \mathbb{R}^{K \times d} : \langle A_{t, k} - A_{t, \ell}, \theta \rangle \in (2 \delta s_Ah, 2 \delta s_A(h + 1) ]\}$ where
 \begin{align*}
     \delta = \frac{\sigma s_A \nu }{\alpha } \sqrt{\frac{8 \left(s_0 + \frac{2 \nu \sqrt{s_0} }{\phi_0^2}\right)}{t -1}}.
 \end{align*}
 By conditioning on $I_h$, we get:
 \begin{align*}
     & \EXP\left[  \indicator\left\{\langle A_{t, \ell}, \hat{\theta}_t \rangle \ge \langle  A_{t, k}, \hat{\theta}_t \rangle  \right\} \langle A_{t, k} - A_{t, \ell}, \theta \rangle \mid  \mathcal{A}_t \in \mathcal{R}_k \cap I_h, \ \mathcal{E}_t, \ \mathcal{G}_{t}^{\frac{\alpha}{2\nu}}\right]
     \\
     & \le \sum_{h=0}^{\lceil s_1 / \delta \rceil}\EXP\left[  \indicator\left\{\langle A_{t, \ell}, \hat{\theta}_t \rangle \ge \langle  A_{t, k}, \hat{\theta}_t \rangle  \right\} \langle A_{t, k} - A_{t, \ell}, \theta \rangle \mid  \mathcal{A}_t \in \mathcal{R}_k \cap I_h, \ \mathcal{E}_t, \ \mathcal{G}_{t}^{\frac{\alpha}{2\nu}} \right] \Pr(\mathcal{A}_t \in I_h)
     \\
     & \stackrel{(a)}{\le}  \sum_{h=0}^{\lceil s_1 / \delta \rceil} 2 \delta s_A (h + 1)\EXP\left[  \indicator\left\{\langle A_{t, \ell}, \hat{\theta}_t \rangle \ge \langle  A_{t, k}, \hat{\theta}_t \rangle \right\} \mid  \mathcal{A}_t \in \mathcal{R}_k \cap I_h, \ \mathcal{E}_t, \ \mathcal{G}_{t}^{\frac{\alpha}{2\nu}}
     \right] \\
     &\qquad\qquad\qquad\qquad \times \Pr \left(\langle A_{t, k} - A_{t, \ell}, \theta \rangle  \in (0, 2 \delta s_A(h + 1) ] \right)
     \\
     & \stackrel{(b)}{\le} \sum_{h=0}^{\lceil s_1 / \delta \rceil} 4 \delta^2 s_A^2 (h + 1)^2 C_{\text{m}}\Pr\left(  \langle A_{t, \ell}, \hat{\theta}_t \rangle \ge \langle  A_{t, k}, \hat{\theta}_t \rangle  \mid  \mathcal{A}_t \in \mathcal{R}_k \cap I_h, \ \mathcal{E}_t, \ \mathcal{G}_{t}^{\frac{\alpha}{2\nu}}
     \right),
 \end{align*}
 where for $(a)$, we used the definition of $I_h$ and for $(b)$, we used Assumption~\ref{asm:margin}. Under the event $ \mathcal{A}_t \in I_h$, the event $\langle A_{t, \ell}, \hat{\theta}_t \rangle \ge \langle  A_{t, k}, \hat{\theta}_t \rangle$ happens only when at least one of the events $ \langle A_{t, k} , \theta - \hat{\theta} \rangle \ge \delta s_A h$ or $ \langle A_{t, \ell} , \hat{\theta} - \theta \rangle \ge \delta s_A h $ holds. Therefore, 
 \begin{align*}
     & \Pr\left(  \langle A_{t, \ell}, \hat{\theta}_t \rangle \ge \langle  A_{t, k}, \hat{\theta}_t \rangle
     \mid  \mathcal{A}_t \in \mathcal{R}_k \cap I_h, \ \mathcal{E}_t, \ \mathcal{G}_{t}^{\frac{\alpha}{2\nu}}
     \right) 
     \\
     & \le \Pr\left(  \langle A_{t, k} , \theta - \hat{\theta} \rangle \ge \delta s_A h   \mid  \mathcal{A}_t \in \mathcal{R}_k \cap I_h, \ \mathcal{E}_t, \ \mathcal{G}_{t}^{\frac{\alpha}{2\nu}}
     \right)  \\
     &\qquad\qquad\qquad + \Pr\left( \langle A_{t, \ell} , \hat{\theta} - \theta \rangle \ge \delta s_A h  \mid \mathcal{A}_t \in \mathcal{R}_k \cap I_h, \ \mathcal{E}_t, \ \mathcal{G}_{t}^{\frac{\alpha}{2\nu}}
     \right) 
     \\
     & \stackrel{(a)}{\le} \Pr\left(   \| \theta - \hat{\theta}\|_2 \ge \delta  h   \mid  \mathcal{A}_t \in \mathcal{R}_k \cap I_h, \ \mathcal{E}_t, \ \mathcal{G}_{t}^{\frac{\alpha}{2\nu}}
     \right)  + \Pr\left(  \| \theta - \hat{\theta}\|_2 \ge \delta  h  \mid  \mathcal{A}_t \in \mathcal{R}_k \cap I_h, \ \mathcal{E}_t, \ \mathcal{G}_{t}^{\frac{\alpha}{2\nu}}
     \right) 
     \\
     & = 2 \Pr\left(   \| \theta - \hat{\theta}\|_2 \ge \delta  h  \mid  \mathcal{A}_t \in \mathcal{R}_k \cap I_h, \ \mathcal{E}_t, \ \mathcal{G}_{t}^{\frac{\alpha}{2\nu}}
     \right),
 \end{align*}
 where for $(a)$, we used the Cauchy–Schwarz inequality. Let us denote $s' = s_0 + \frac{2 \nu \sqrt{s_0}}{\phi_0^2}$. Then, using Lemma~\ref{lm:LS_estim_risk_k2}, we get:
 \begin{align*}
     \Pr\left(   \| \theta - \hat{\theta}\|_2 \ge \delta  h   \mid  \mathcal{A}_t \in \mathcal{R}_k \cap I_h, \ \mathcal{E}_t, \ \mathcal{G}_{t}^{\frac{\alpha}{2\nu}}
     \right) & \le 2 s'\exp \left( -\frac{\alpha^2 t \delta^2 h^2}{8 \sigma^2 s_A^2 \nu^2 s'} \right)
     \\
     & = 2 s'\exp \left( -h^2\right).
 \end{align*}
 We also trivially have that:
 \begin{align*}
     \Pr\left(   \| \theta - \hat{\theta}\|_2 \ge \delta  h  \mid  \mathcal{A}_t \in \mathcal{R}_k \cap I_h, \ \mathcal{E}_t, \ \mathcal{G}_{t}^{\frac{\alpha}{2\nu}}
     \right) & \le 1.
 \end{align*}
 Therefore, we can upper bound the expected instantaneous regret as:
 \begin{align*}
     r_t^\pi & \le \sum_{k=1}^2 \EXP\left[  r_t^\pi \mid  \mathcal{A}_t \in \mathcal{R}_k\right]  \Pr\left(\mathcal{A}_t \in \mathcal{R}_k\right)
     \\
     & \le \sum_{k=1}^2 \left(\sum_{\ell \neq k}\left( \sum_{h=0}^{\lceil s_1/\delta \rceil} \left(4 \delta^2 s_A^2 (h + 1)^2 C_{\text{m}} \min \left\{ 1, 4s' \exp\left( -h^2\right) \right\} \right) \right) \right.
     \\
     & \ \ \ \left.+ 2  s_A s_1  \left(\Pr(\mathcal{E}_t^c) + \Pr\left(\left(\mathcal{G}_{t}^{\frac{\alpha}{2 \nu}}\right)^c\right) \right) \cdot \Pr\left(\mathcal{A}_t \in \mathcal{R}_k \right)\right)  
     \\
     & \stackrel{(a)}{\le} 4 \delta^2 s_A^2  C_{\text{m}} \sum_{h=0}^{\lceil s_1/\delta \rceil} (h + 1)^2\min \left\{ 1, 4 s' \exp\left( -h^2\right) \right\} + 2  s_A s_1 \left(\Pr(\mathcal{E}_t^c) + \Pr\left(\left(\mathcal{G}_{t}^{\frac{\alpha}{2\nu}}\right)^c \middle| \mathcal{E}_t\right) \right)
     \\
     & \le 4 \delta^2 s_A^2  C_{\text{m}} \left( \sum_{h=0}^{h_0} (h + 1)^2 + \sum_{h= h_0 + 1}^ {h_{\max}}  4 s' (h+1)^2\exp\left( -h^2\right) \right)  + 2  s_A s_1  \left(\Pr(\mathcal{E}_t^c) + \Pr\left(\left(\mathcal{G}_{t}^{\frac{\alpha}{2\nu}}\right)^c \middle| \mathcal{E}_t\right) \right)
 \end{align*}
 where for $(a)$, we used $\sum_{k=1}^2 \Pr\left(\mathcal{A}_t \in \mathcal{R}_k\right)= 1$ from Assumption~\ref{asm:margin} and we set $ h_0 \coloneqq \lfloor \sqrt{\log 4 s'} + 1\rfloor$. We have:
 \begin{align*}
     \sum_{h=h_0 + 1}^{h_{\max}} (h+1)^2 \exp(-h^2) & = \sum_{h=h_0 + 1}^{h_{\max}} h^2\exp(-h^2) + 2\sum_{h=h_0 + 1}^{h_{\max}} h \exp(-h^2) + \sum_{h=h_0 + 1}^{h_{\max}} \exp(-h^2)
     \\
     & \le \int_{h_0}^\infty h^2\exp(-h^2)  dh +  2 \int_{h_0}^\infty h\exp(-h^2)  dh+  \int_{h_0}^\infty \exp(-h^2)  dh.
 \end{align*}
 Using an integration by parts, the inequality $ \int_{h_0}^\infty \exp(-h_0^2)dh \le \exp(-h_0^2)/(h_0 + \sqrt{h_0^2 + 4/\pi}) \le \exp(- h_0^2)$, and $ h_0 \ge 1$ from $s_0 \ge 1$,
 we get:
 \begin{align*}
     \int_{h_0}^\infty h^2 \exp(-h^2) dh & \le  \frac{1}{2} h_0 \exp(-h_0^2) + \frac{1}{2} \exp(-h_0^2)
     \\
     2 \int_{h_0}^\infty  h \exp(-h^2) dh & =  \exp(-h_0^2)
     \\
     \int_{h_0}^\infty  \exp(-h^2) dh & \le \exp(-h_0^2).
 \end{align*}
 Therefore, 
 \begin{align*}
      \sum_{h=h_0 + 1}^{h_{\max}} (h+1)^2 \exp(-h^2) & \le \frac{1}{2}h_0 \exp(-h_0^2) + \frac{5}{2}\exp(-h_0^2)
      \\
      & \le h_0 \exp(-h_0^2) + 5\exp(-h_0^2).
 \end{align*}
 We get:
 \begin{align*}
     \sum_{h=0}^{h_0} (h + 1)^2 + \sum_{h= h_0 + 1}^ {h_{\max}}  4 s' (h+1)^2\exp\left( -h^2\right) & \le \frac{(h_0 + 1)(h_0 + 2)(2 h_0 + 3)}{6} + 4 s'  (h_0 + 5) \exp(-h_0^2) 
     \\
     &\le \frac{2h_0^3 + 9 h_0^2 + 13h_0 + 6}{6} + 4s'(h_0 + 5) \frac{1}{4s'}
     \\
     & \stackrel{(a)}{\le} 11 h_0^3,
 \end{align*}
 where for $(a)$, we used $h_0 \ge 1$. Finally, we get:
 \begin{align*}
     r_t^\pi & \le 44 \delta^2 s_A^2  C_{\text{m}} h_0^3 + 2  s_A s_1 \left(\Pr(\mathcal{E}_t^c) + \Pr\left(\left(\mathcal{G}_{t}^{\frac{\alpha}{2\nu}}\right)^c \middle| \mathcal{E}_t \right) \right)
     \\
     & \le \frac{352 \sigma^2 s_A^4 C_{\text{m}}  h_0^3 \nu^2 \left(s_0 + \frac{2 \nu\sqrt{s_0}}{\phi_0^2}\right)}{\alpha^2} \frac{1}{t - 1} + 2  s_A s_1  \left(\Pr(\mathcal{E}_t^c) + \Pr\left(\left(\mathcal{G}_{t}^{\frac{\alpha}{2\nu}}\right)^c \middle| \mathcal{E}_t\right) \right).
 \end{align*}
 This concludes the proof.\ep
 
\subsection{Proof of Lemma~\ref{lm:inst_regret_bound_wo_margin_k2}}

Let $r_t^\pi$ be the instantaneous expected regret of algorithm $\pi$ in round $t$ defined as:
 \begin{align*}
     r_t^\pi \coloneqq \EXP\left[ \max_{A \in \mathcal{A}_t } \langle A - A_t, \theta \rangle\right].
 \end{align*}
 Let us define the events $\mathcal{R}_k \coloneqq \{ \mathcal{A}_t \in \mathbb{R}^{K \times d}:  k \in \argmax_{k'} \langle A_{t, k'}, \theta \rangle\}$ and $ \mathcal{G}_{t}^{\lambda} \coloneqq \left\{\lambda_{\min } ( \hat{\Sigma}_{\hat{S}} ) \ge \lambda \right\}$. As in the proof of Lemma~\ref{lm:inst_regret_bound_klarge}, we get:
 \begin{align*}
     r_t^\pi & \le \sum_{k=1}^2 \EXP\left[  r_t^\pi \mid  \mathcal{A}_t \in \mathcal{R}_k\right]  \Pr\left(\mathcal{A}_t \in \mathcal{R}_k\right),
 \end{align*}
 and
 \begin{align*}
     \EXP\left[  r_t^\pi \mid  \mathcal{A}_t \in \mathcal{R}_k\right]  & = \EXP\left[  \langle A_{t, k} - A_t, \theta \rangle \mid  \mathcal{A}_t \in \mathcal{R}_k\right]
    \\
     & \le \sum_{\ell \neq k} \EXP\left[  \indicator\left\{\langle A_{t, \ell}, \hat{\theta}_t \rangle \ge \langle  A_{t, k}, \hat{\theta}_t \rangle  \right\} \langle A_{t, k} - A_{t, \ell}, \theta \rangle \mid  \mathcal{A}_t \in \mathcal{R}_k, \ \mathcal{E}_t, \ \mathcal{G}_{t}^{\frac{\alpha}{2 \nu }} \right] 
     \\
     & \ \ \ + 2  s_A s_1 \left(\Pr(\mathcal{E}_t^c) + \Pr\left(\left(\mathcal{G}_{t}^{\frac{\alpha}{2 \nu }}\right)^c \middle| \mathcal{E}_t\right) \right).
\end{align*} 
 Let us denote the event $I_h \coloneqq \{\mathcal{A}_t \in \mathbb{R}^{K \times d} : \langle A_{t, k} - A_{t, \ell}, \theta \rangle \in (2 \delta s_Ah, 2 \delta s_A(h + 1) ]\}$ where
 \begin{align*}
      \delta = \frac{\sigma s_A  \nu }{\alpha} \sqrt{\frac{8 \left(s_0 + \frac{2 \nu  s_0 }{\phi_0^2}\right)}{t -1}}.
 \end{align*}
 By conditioning on $I_h$, we get:
 \begin{align*}
     & \EXP\left[  \indicator\left\{\langle A_{t, \ell}, \hat{\theta}_t \rangle \ge \langle  A_{t, k}, \hat{\theta}_t \rangle  \right\} \langle A_{t, k} - A_{t, \ell}, \theta \rangle \mid  \mathcal{A}_t \in \mathcal{R}_k \cap I_h, \ \mathcal{E}_t, \ \mathcal{G}_{t}^{\frac{\alpha}{2 \nu }}\right]
     \\
     & \le \sum_{h=0}^{\lceil s_1 / \delta \rceil}\EXP\left[  \indicator\left\{\langle A_{t, \ell}, \hat{\theta}_t \rangle \ge \langle  A_{t, k}, \hat{\theta}_t \rangle  \right\} \langle A_{t, k} - A_{t, \ell}, \theta \rangle \mid  \mathcal{A}_t \in \mathcal{R}_k \cap I_h, \ \mathcal{E}_t, \ \mathcal{G}_{t}^{\frac{\alpha}{2 \nu }} \right] \Pr(\mathcal{A}_t \in I_h)
     \\
     & \stackrel{(a)}{\le}  \sum_{h=0}^{\lceil s_1 / \delta \rceil} 2 \delta s_A (h + 1)\EXP\left[  \indicator\left\{\langle A_{t, \ell}, \hat{\theta}_t \rangle \ge \langle  A_{t, k}, \hat{\theta}_t \rangle \right\} \mid  \mathcal{A}_t \in \mathcal{R}_k \cap I_h, \ \mathcal{E}_t, \ \mathcal{G}_{t}^{\frac{\alpha}{2 \nu }}
     \right]\\
    &\qquad\qquad\qquad\qquad \times \Pr \left(\langle A_{t, k} - A_{t, \ell}, \theta \rangle  \in (0, 2 \delta s_A(h + 1) ] \right)
     \\
     & \le \sum_{h=0}^{\lceil s_1 / \delta \rceil} 2 \delta s_A (h + 1)\Pr\left(  \langle A_{t, \ell}, \hat{\theta}_t \rangle \ge \langle  A_{t, k}, \hat{\theta}_t \rangle  \mid  \mathcal{A}_t \in \mathcal{R}_k \cap I_h, \ \mathcal{E}_t, \ \mathcal{G}_{t}^{\frac{\alpha}{2 \nu }}
     \right),
 \end{align*}
 where for $(a)$, we used the definition of $I_h$. Under the event $ \mathcal{A}_t \in I_h$, the event $\langle A_{t, \ell}, \hat{\theta}_t \rangle \ge \langle  A_{t, k}, \hat{\theta}_t \rangle$ happens only when at least one of the events $ \langle A_{t, k} , \theta - \hat{\theta} \rangle \ge \delta s_A h$ or $ \langle A_{t, \ell} , \hat{\theta} - \theta \rangle \ge \delta s_A h $ holds. Therefore, 
 \begin{align*}
     & \Pr\left(  \langle A_{t, \ell}, \hat{\theta}_t \rangle \ge \langle  A_{t, k}, \hat{\theta}_t \rangle
     \mid  \mathcal{A}_t \in \mathcal{R}_k \cap I_h, \ \mathcal{E}_t, \ \mathcal{G}_{t}^{\frac{\alpha}{2 \nu }}
     \right) 
     \\
     & \le \Pr\left(  \langle A_{t, k} , \theta - \hat{\theta} \rangle \ge \delta s_A h   \mid  \mathcal{A}_t \in \mathcal{R}_k \cap I_h, \ \mathcal{E}_t, \ \mathcal{G}_{t}^{\frac{\alpha}{2\nu}}
     \right)  \\
     &\qquad\qquad\qquad + \Pr\left( \langle A_{t, \ell} , \hat{\theta} - \theta \rangle \ge \delta s_A h  \mid \mathcal{A}_t \in \mathcal{R}_k \cap I_h, \ \mathcal{E}_t, \ \mathcal{G}_{t}^{\frac{\alpha}{2 \nu }}
     \right) 
     \\
     & \stackrel{(a)}{\le} \Pr\left(   \| \theta - \hat{\theta}\|_2 \ge \delta  h   \mid  \mathcal{A}_t \in \mathcal{R}_k \cap I_h, \ \mathcal{E}_t, \ \mathcal{G}_{t}^{\frac{\alpha}{2 \nu }}
     \right)  + \Pr\left(  \| \theta - \hat{\theta}\|_2 \ge \delta  h  \mid  \mathcal{A}_t \in \mathcal{R}_k \cap I_h, \ \mathcal{E}_t, \ \mathcal{G}_{t}^{\frac{\alpha}{2 \nu }}
     \right) 
     \\
     & = 2 \Pr\left(   \| \theta - \hat{\theta}\|_2 \ge \delta  h  \mid  \mathcal{A}_t \in \mathcal{R}_k \cap I_h, \ \mathcal{E}_t, \ \mathcal{G}_{t}^{\frac{\alpha}{2 \nu }}
     \right),
 \end{align*}
 where for $(a)$, we used the Cauchy–Schwarz inequality. Let us denote $s' = s_0 + \frac{2 \nu \sqrt{s_0}}{\phi_0^2}$. Then, using Lemma~\ref{lm:LS_estim_risk_klarge}, we get:
 \begin{align*}
     \Pr\left(   \| \theta - \hat{\theta}\|_2 \ge \delta  h   \mid  \mathcal{A}_t \in \mathcal{R}_k \cap I_h, \ \mathcal{E}_t, \ \mathcal{G}_{t}^{\frac{\alpha}{2\nu }}
     \right) & \le 2 s'\exp \left( -\frac{\alpha^2 t \delta^2 h^2}{8 \sigma^2 s_A^2 \nu^2 C_{\textnormal{b}}^2 s'} \right)
     \\
     & = 2 s'\exp \left( -h^2\right).
 \end{align*}
 We also trivially have that:
 \begin{align*}
     \Pr\left(   \| \theta - \hat{\theta}\|_2 \ge \delta  h  \mid  \mathcal{A}_t \in \mathcal{R}_k \cap I_h, \ \mathcal{E}_t, \ \mathcal{G}_{t}^{\frac{\alpha}{2\nu }}
     \right) & \le 1.
 \end{align*}
 Therefore, we can upper bound the expected instantaneous regret as:
 \begin{align*}
     r_t^\pi & \le \sum_{k=1}^2 \EXP\left[  r_t^\pi \mid  \mathcal{A}_t \in \mathcal{R}_k\right]  \Pr\left(\mathcal{A}_t \in \mathcal{R}_k\right)
     \\
     & \le \sum_{k=1}^2 \left(\sum_{\ell \neq k}\left( \sum_{h=0}^{\lceil s_1/\delta \rceil} \left(2 \delta s_A (h + 1) \min \left\{ 1, 4s' \exp\left( -h^2\right) \right\} \right) \right) \right.
     \\
     & \ \ \ \left.+ 2  s_A s_1  \left(\Pr(\mathcal{E}_t^c) + \Pr\left(\left(\mathcal{G}_{t}^{\frac{\alpha}{2\nu }}\right)^c\right) \right) \cdot \Pr\left(\mathcal{A}_t \in \mathcal{R}_k \right)\right)  
     \\
     & \stackrel{(a)}{\le} 2 \delta s_A  \sum_{h=0}^{\lceil s_1/\delta \rceil} (h + 1)\min \left\{ 1, 4 s' \exp\left( -h^2\right) \right\} + 2   s_A s_1 \left(\Pr(\mathcal{E}_t^c) + \Pr\left(\left(\mathcal{G}_{t}^{\frac{\alpha}{2\nu }}\right)^c \middle| \mathcal{E}_t\right) \right)
     \\
     & \le 2 \delta s_A  \left( \sum_{h=0}^{h_0} (h + 1) + \sum_{h= h_0 + 1}^ {h_{\max}}  4 s' (h+1)\exp\left( -h^2\right) \right)  + 2  s_A s_1  \left(\Pr(\mathcal{E}_t^c) + \Pr\left(\left(\mathcal{G}_{t}^{\frac{\alpha}{2 \nu }}\right)^c \middle| \mathcal{E}_t\right) \right)
 \end{align*}
 where for $(a)$, we used $\sum_{k=1}^K \Pr\left(\mathcal{A}_t \in \mathcal{R}_k\right)= 1$ and we set $ h_0 \coloneqq \lfloor \sqrt{\log 4 s'} + 1\rfloor$. We have:
 \begin{align*}
     \sum_{h=h_0 + 1}^{h_{\max}} (h+1) \exp(-h^2) & = \sum_{h=h_0 + 1}^{h_{\max}} h \exp(-h^2) + \sum_{h=h_0 + 1}^{h_{\max}} \exp(-h^2)
     \\
     & \le \int_{h_0}^\infty h\exp(-h^2)  dh+  \int_{h_0}^\infty \exp(-h^2)  dh.
 \end{align*}
 Since $ h_0 \ge 1$ from $s_0 \ge 1$, we get:
 \begin{align*}
     \int_{h_0}^\infty  h \exp(-h^2) dh & =  \frac{1}{2}\exp(-h_0^2)
     \\
     \int_{h_0}^\infty  \exp(-h^2) dh & \le \exp(-h_0^2).
 \end{align*}
 Therefore, 
 \begin{align*}
      \sum_{h=h_0 + 1}^{h_{\max}} (h+1) \exp(-h^2) & \le \frac{3}{2}\exp(-h_0^2).
 \end{align*}
 We get:
 \begin{align*}
     \sum_{h=0}^{h_0} (h + 1) + \sum_{h= h_0 + 1}^ {h_{\max}}  4 s' (h+1)\exp\left( -h^2\right) & \le \frac{(h_0 + 1)(h_0 + 2)}{2} + 4 s' \frac{3}{2}\exp(-h_0^2) 
     \\
     &\le \frac{h_0^2 + 3h_0 + 2}{2} + \frac{6s'}{4s'}
     \\
     & \stackrel{(a)}{\le} \frac{9}{2}h_0^2,
 \end{align*}
 where for $(a)$, we used $h_0 \ge 1$. Finally, we get:
 \begin{align*}
     r_t^\pi & \le 9 \delta s_A  h_0^2 + 2 s_A s_1 \left(\Pr(\mathcal{E}_t^c) + \Pr\left(\left(\mathcal{G}_{t}^{\frac{\alpha}{ 2 \nu }}\right)^c \middle| \mathcal{E}_t \right) \right)
     \\
     & \le \frac{18 \sigma s_A   h_0^2 \nu }{\alpha}\sqrt{\frac{2 \left(s_0 + \frac{2 \nu \sqrt{s_0} }{\phi_0^2}\right)}{t-1}} + 2   s_A s_1 \left(\Pr(\mathcal{E}_t^c) + \Pr\left(\left(\mathcal{G}_{t}^{\frac{\alpha}{2\nu }}\right)^c \middle| \mathcal{E}_t\right) \right).
 \end{align*}
 This concludes the proof.\ep

\newpage
\section{Additional Experimental Results and Details}
\label{sec:appendix_exp}

\subsection{The implementation of Lasso bandit}
Although the problem formulation for DR Lasso and SA Lasso is the same as ours, the problem formulation for Lasso bandit \citet{bastani2020online} is different from ours. 
In \citet{bastani2020online}, the unknown regression vectors are defined arm-wise and a common context is observed among the arms. 
In the numerical experiments, we followed the comparison idea in \citet{kim2019doubly} and \citet{oh2020sparsity} to apply Lasso bandit of \citet{bastani2020online} to our problem setting. 
The idea is explained as follows: 
from the action set $\mathcal{A}_t$, the $Kd$-dimensional context vector $X_t = (A_{t,1}^\top, A_{t,2}^\top, \ldots,A_{t,K}^\top)^\top \in \mathbb{R}^{Kd}$ and
the $Kd$-dimensional arm-wise unknown regression vector 
$\beta_k = (\indicator \{k=1\} \theta^\top,  \indicator \{k=2\} \theta^\top, \ldots, \indicator \{k=K\} \theta^\top)^\top \in \mathbb{R}^{Kd}$
for each $k \in [K]$ is considered to enable the comparison.  
Under these transformations, we have problem dimension $Kd$ (instead of $d$). 
Thus, the assumptions and the regret guarantees have different scalings mainly in $K$. 

\subsection{Additional Results with Various Correlation Levels}
\label{subsec:various_correlation_d}
\begin{figure}[!ht]
    \centering
    \begin{minipage}[t]{0.24\columnwidth}
        \centering
        \includegraphics[width=1.1\textwidth]{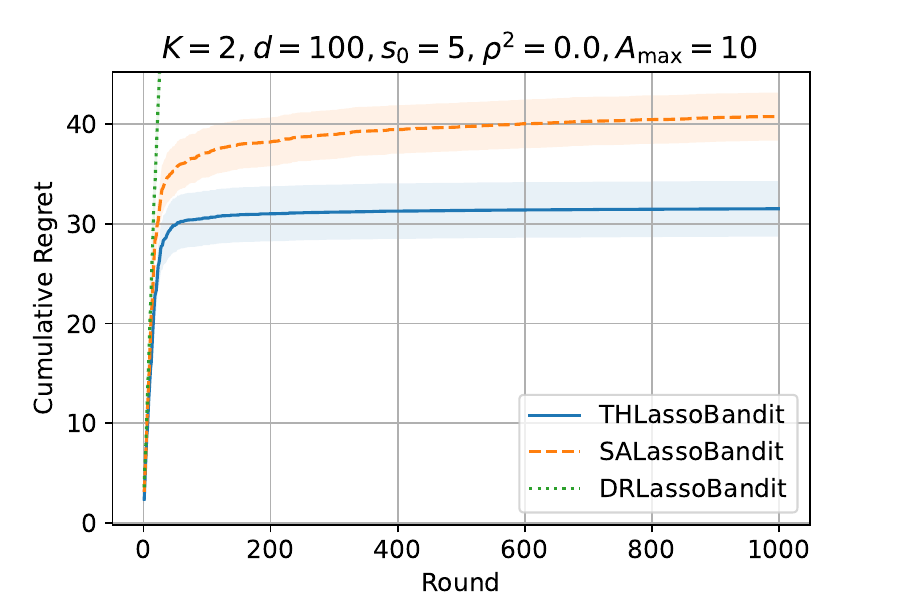}
    \end{minipage}
    \begin{minipage}[t]{0.24\columnwidth}
        \centering
        \includegraphics[width=1.1\textwidth]{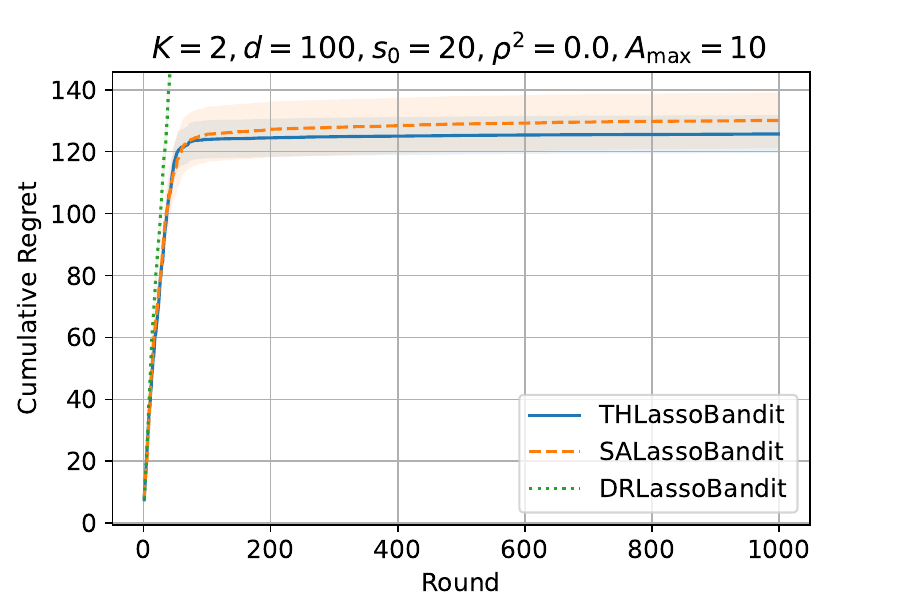}
    \end{minipage}
    \begin{minipage}[t]{0.24\columnwidth}
        \centering
        \includegraphics[width=1.1\textwidth]{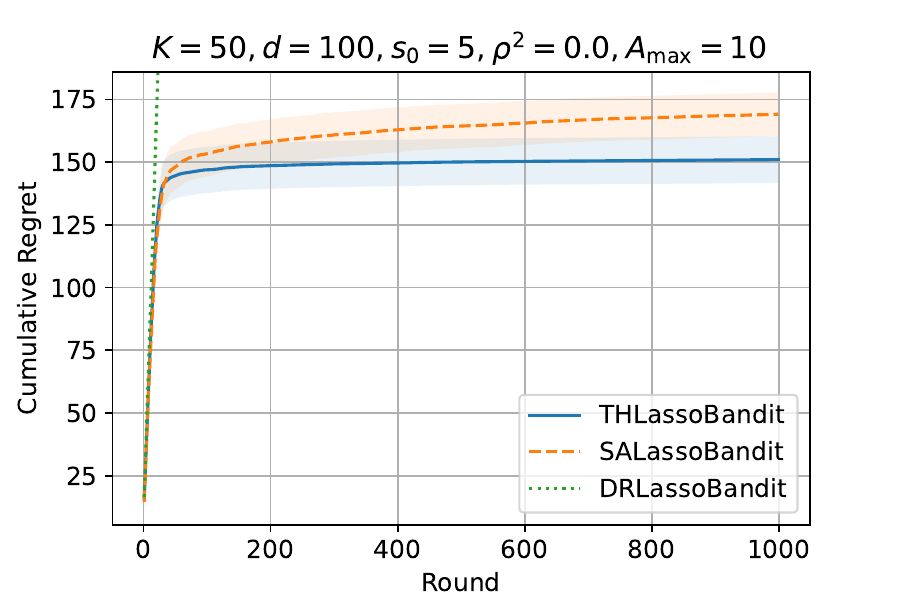}
    \end{minipage}
    \begin{minipage}[t]{0.24\columnwidth}
        \centering
        \includegraphics[width=1.1\textwidth]{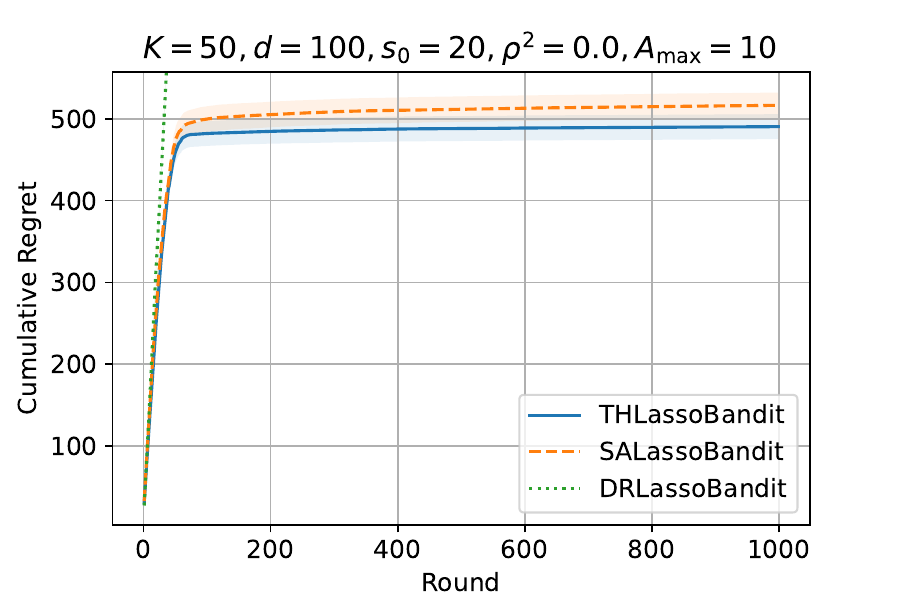}
    \end{minipage} \\
    \begin{minipage}[t]{0.24\columnwidth}
        \centering
        \includegraphics[width=1.1\textwidth]{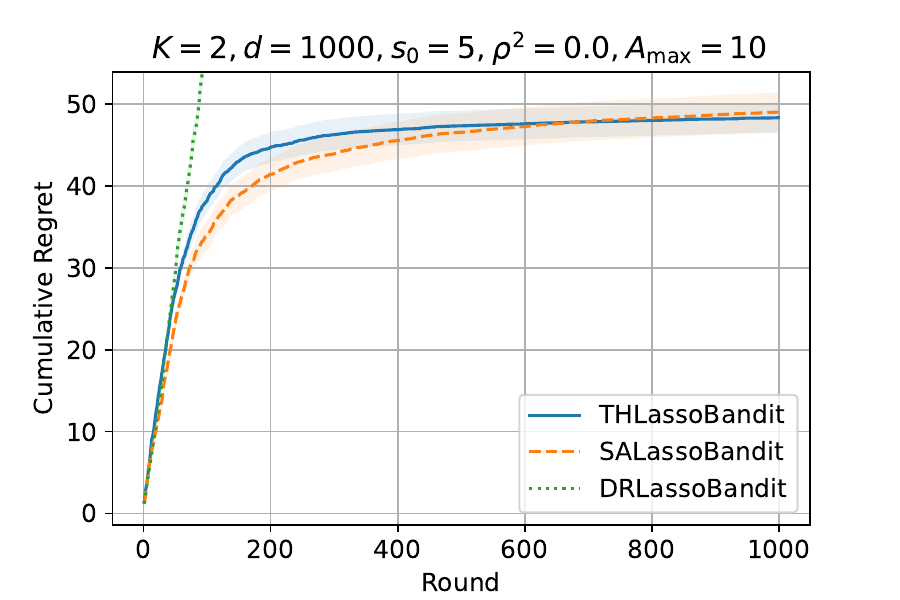}
    \end{minipage}
    \begin{minipage}[t]{0.24\columnwidth}
        \centering
        \includegraphics[width=1.1\textwidth]{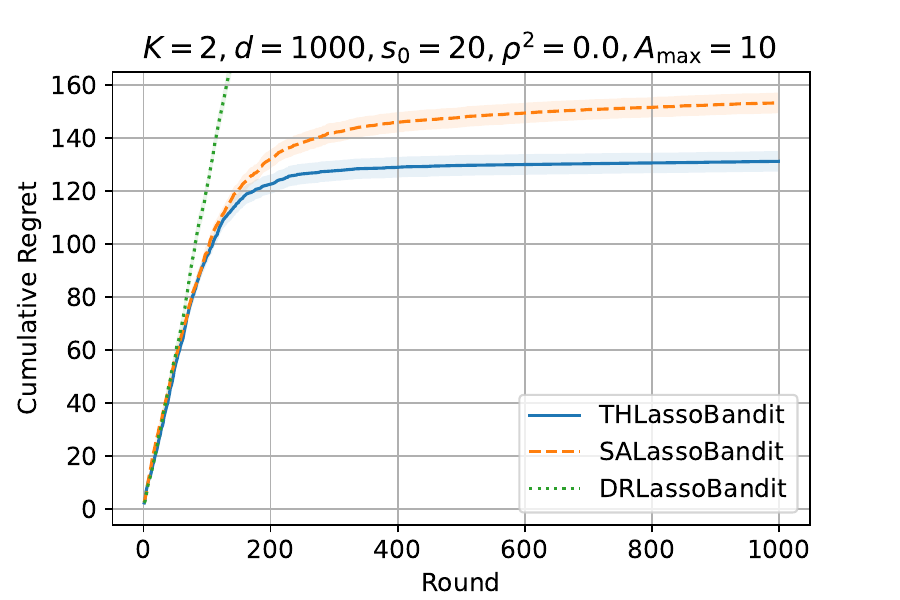}
    \end{minipage}
    \begin{minipage}[t]{0.24\columnwidth}
        \centering
        \includegraphics[width=1.1\textwidth]{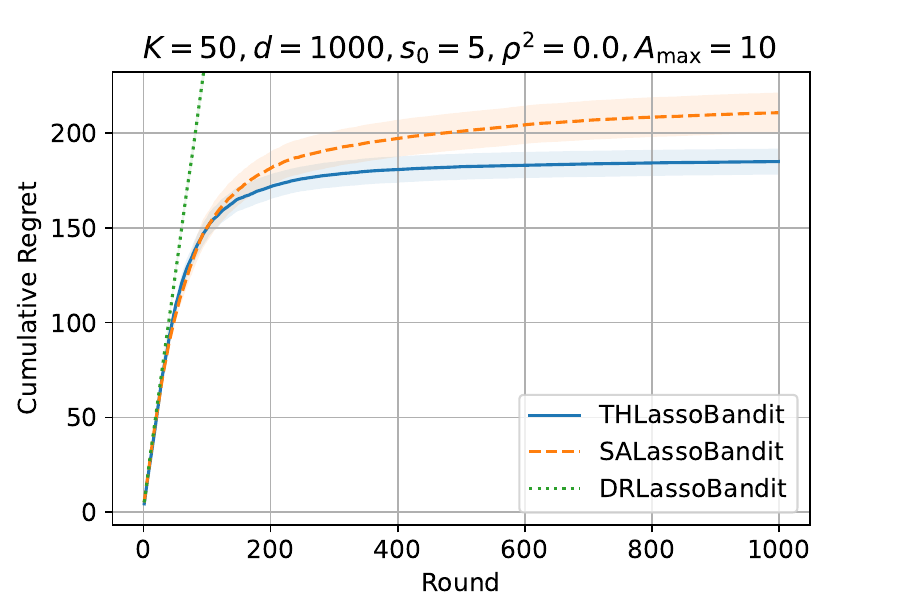}
    \end{minipage}
    \begin{minipage}[t]{0.24\columnwidth}
        \centering
        \includegraphics[width=1.1\textwidth]{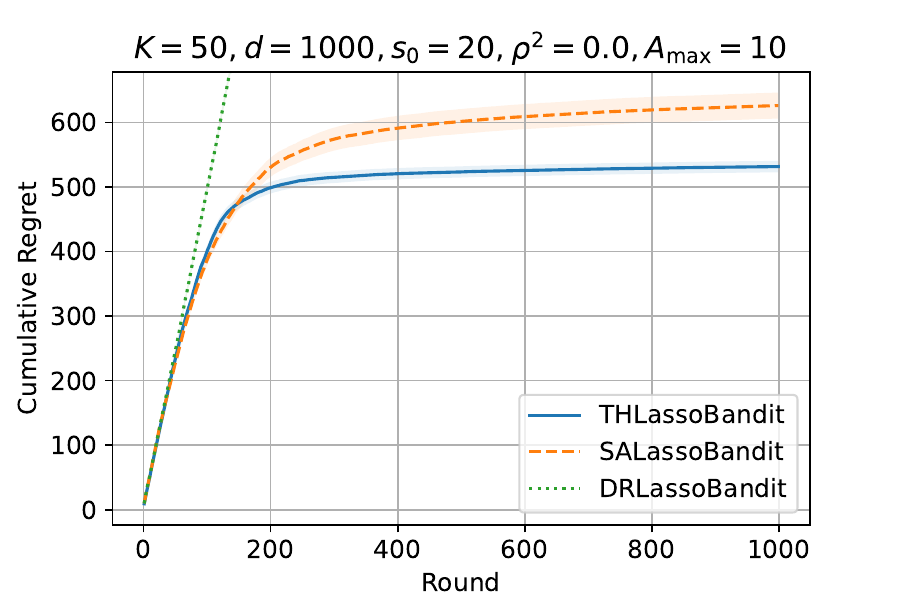}
    \end{minipage} \\
    \begin{minipage}[t]{0.24\columnwidth}
        \centering
        \includegraphics[width=1.1\textwidth]{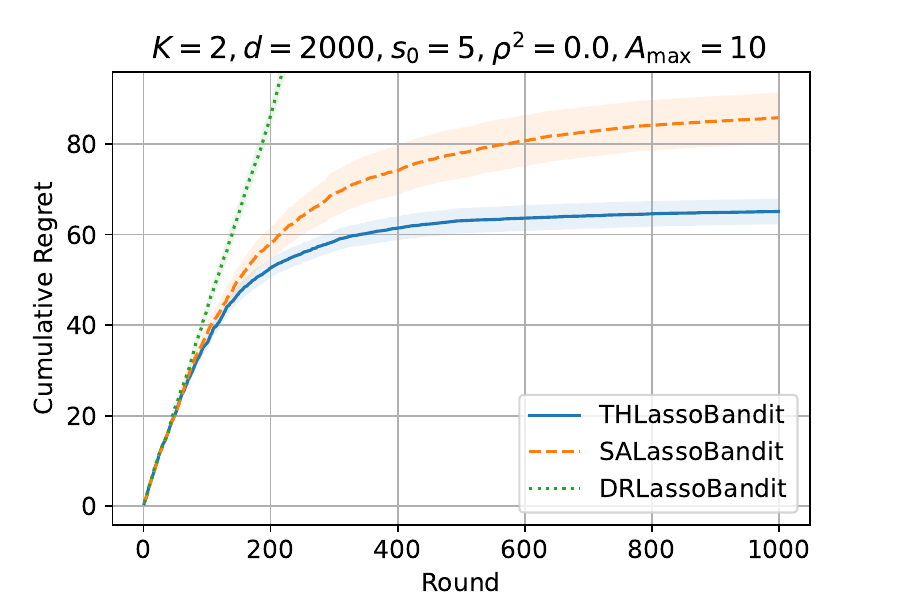}
    \end{minipage}
    \begin{minipage}[t]{0.24\columnwidth}
        \centering
        \includegraphics[width=1.1\textwidth]{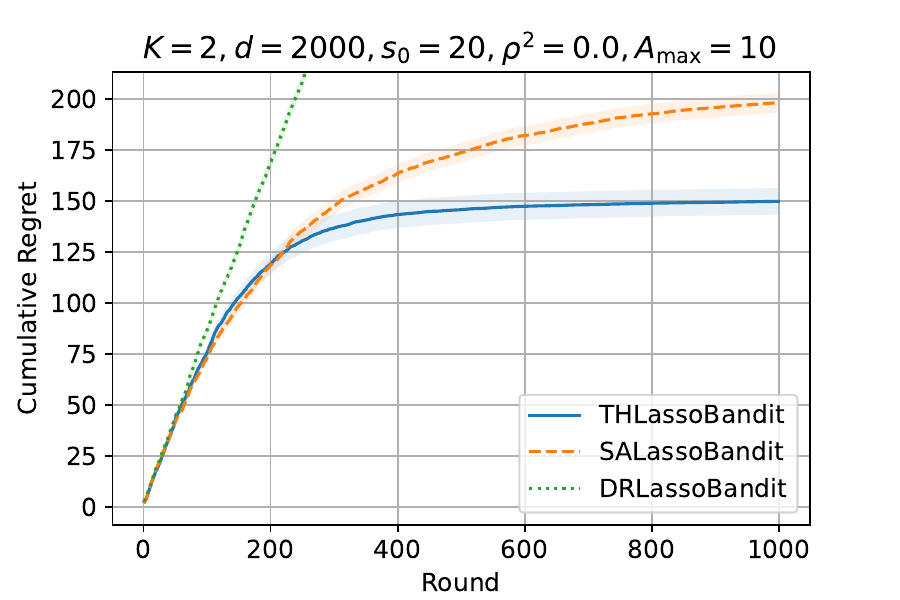}
    \end{minipage}
    \begin{minipage}[t]{0.24\columnwidth}
        \centering
        \includegraphics[width=1.1\textwidth]{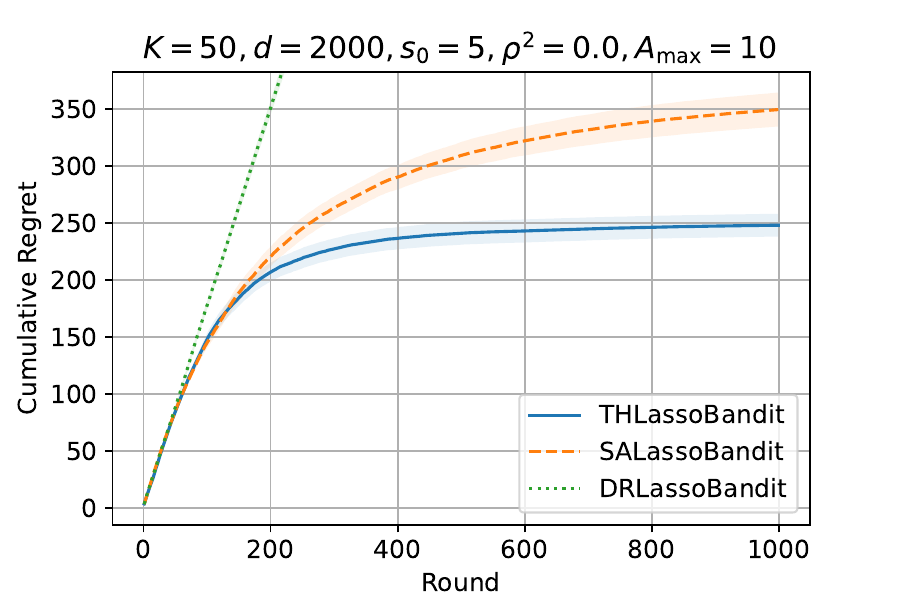}
    \end{minipage}
    \begin{minipage}[t]{0.24\columnwidth}
        \centering
        \includegraphics[width=1.1\textwidth]{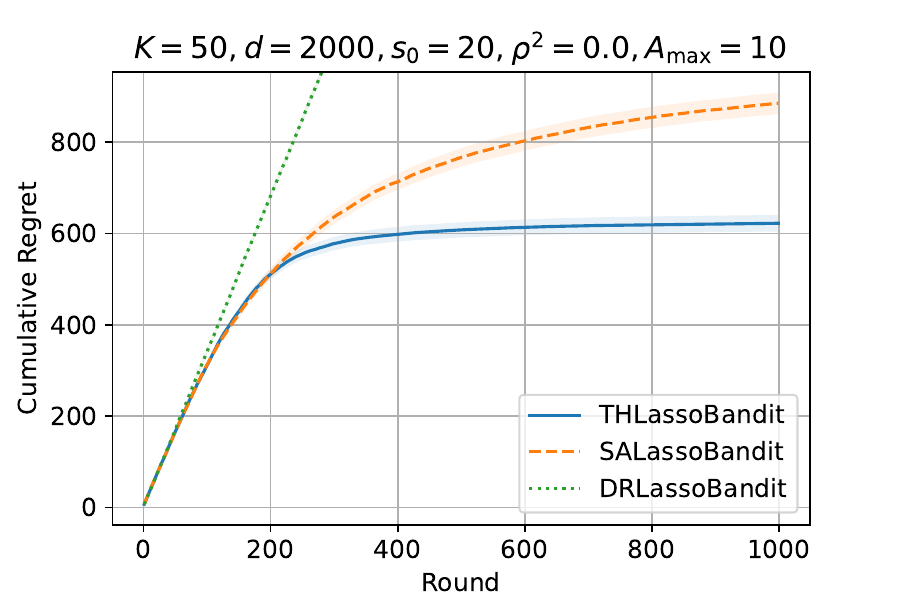}
    \end{minipage} \\
    \begin{minipage}[t]{0.24\columnwidth}
        \centering
        \includegraphics[width=1.1\textwidth]{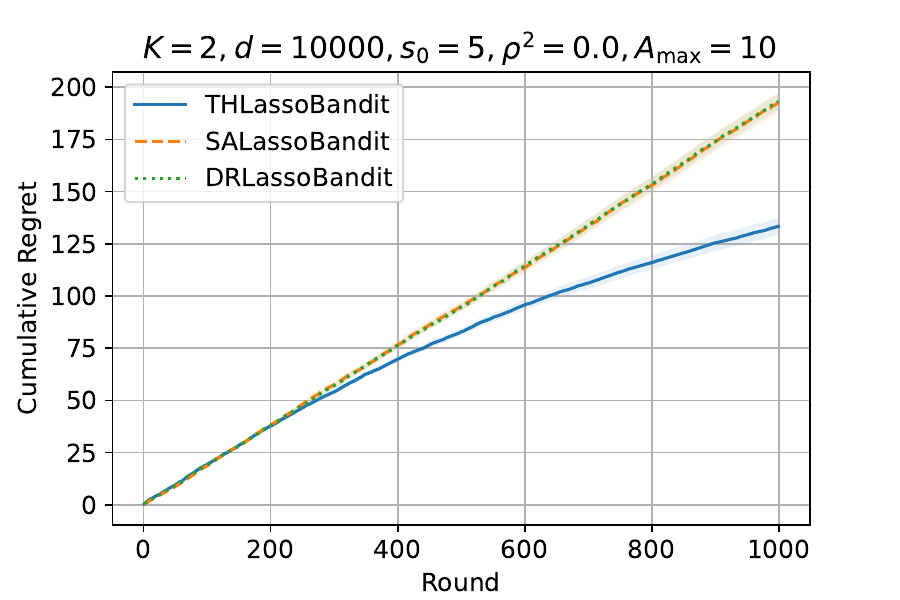}
    \end{minipage}
    \begin{minipage}[t]{0.24\columnwidth}
        \centering
        \includegraphics[width=1.1\textwidth]{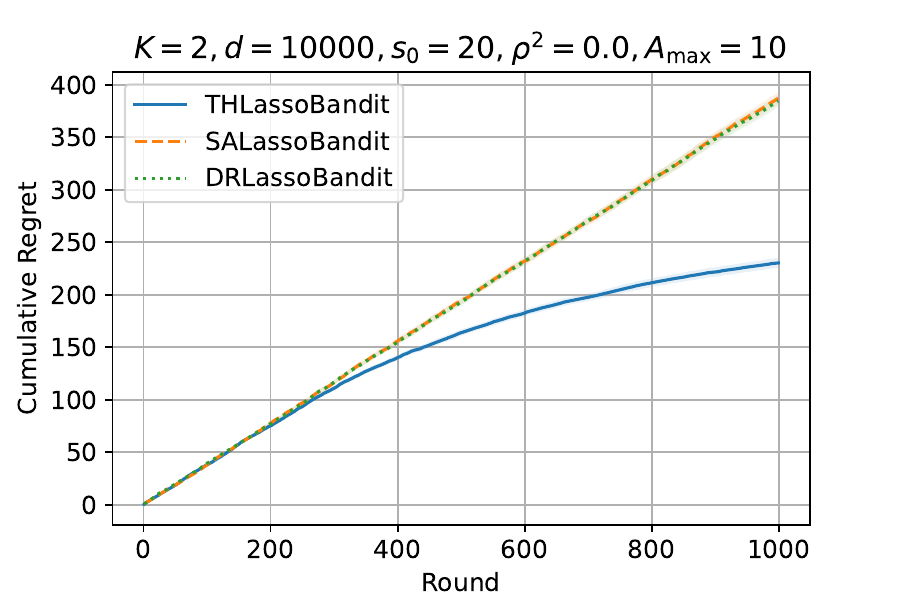}
    \end{minipage}
    \begin{minipage}[t]{0.24\columnwidth}
        \centering
        \includegraphics[width=1.1\textwidth]{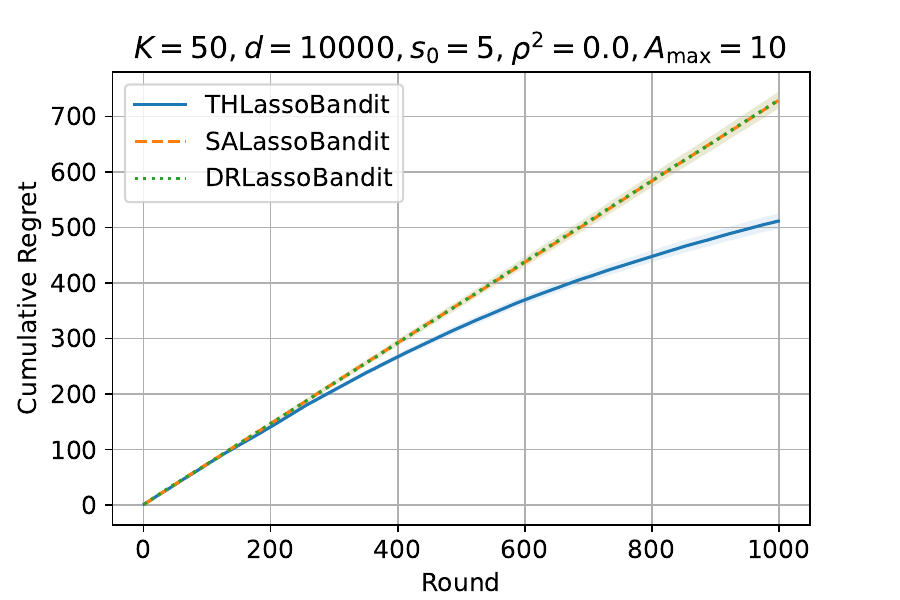}
    \end{minipage}
    \begin{minipage}[t]{0.24\columnwidth}
        \centering
        \includegraphics[width=1.1\textwidth]{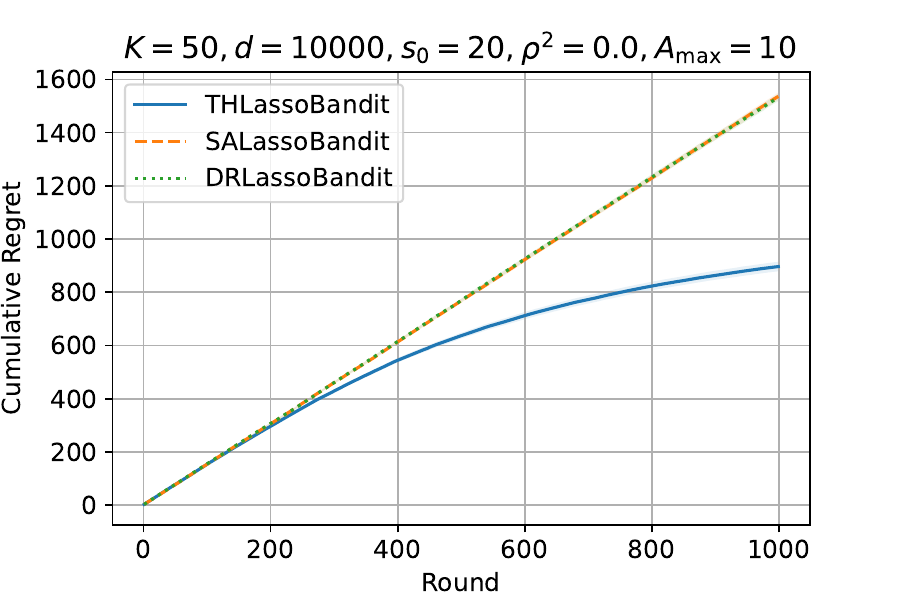}
    \end{minipage}
    \caption{ \small Cumulative regret of the three algorithms with $\rho^2=0.0$, $A_{\max}=10$, $K\in\{2, 50\}$, $d\in \{100, 1000, 2000, 10000\}$, and $s_0\in \{5, 20\}$.
    The shaded area represents the standard errors.}
    \label{fig:synthetic_regrets_rho0.0}
\end{figure}

\begin{figure}[!ht]
    \centering
    \begin{minipage}[t]{0.24\columnwidth}
        \centering
        \includegraphics[width=1.1\textwidth]{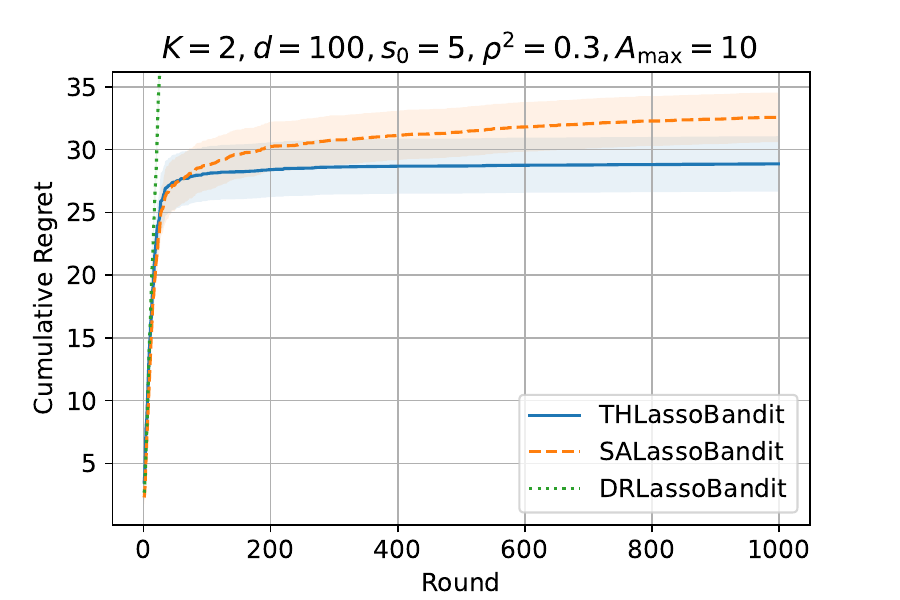}
    \end{minipage}
    \begin{minipage}[t]{0.24\columnwidth}
        \centering
        \includegraphics[width=1.1\textwidth]{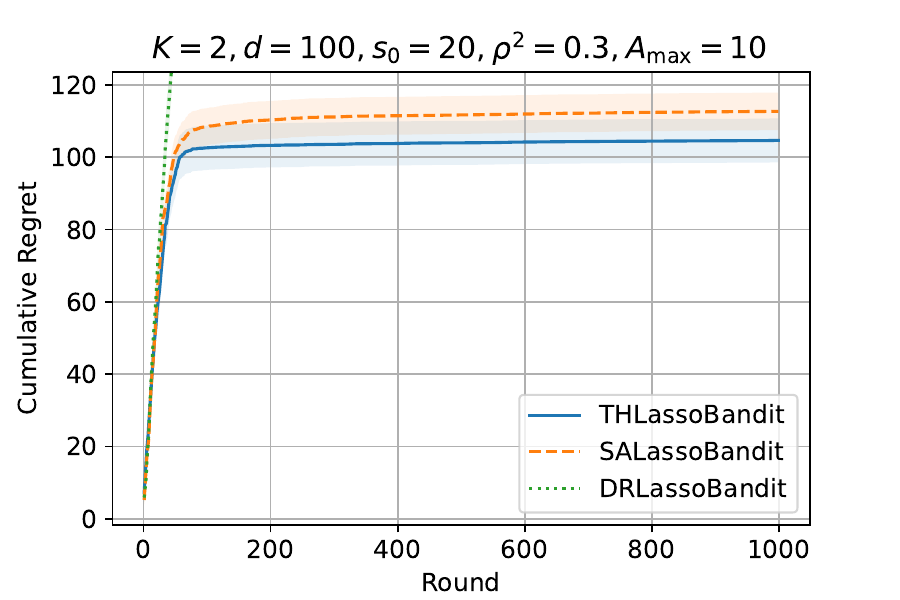}
    \end{minipage}
    \begin{minipage}[t]{0.24\columnwidth}
        \centering
        \includegraphics[width=1.1\textwidth]{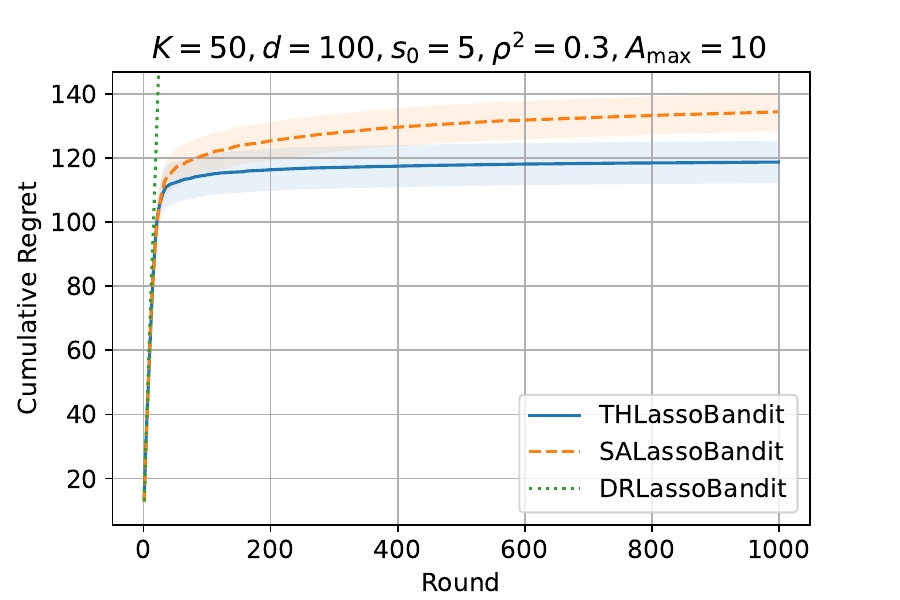}
    \end{minipage}
    \begin{minipage}[t]{0.24\columnwidth}
        \centering
        \includegraphics[width=1.1\textwidth]{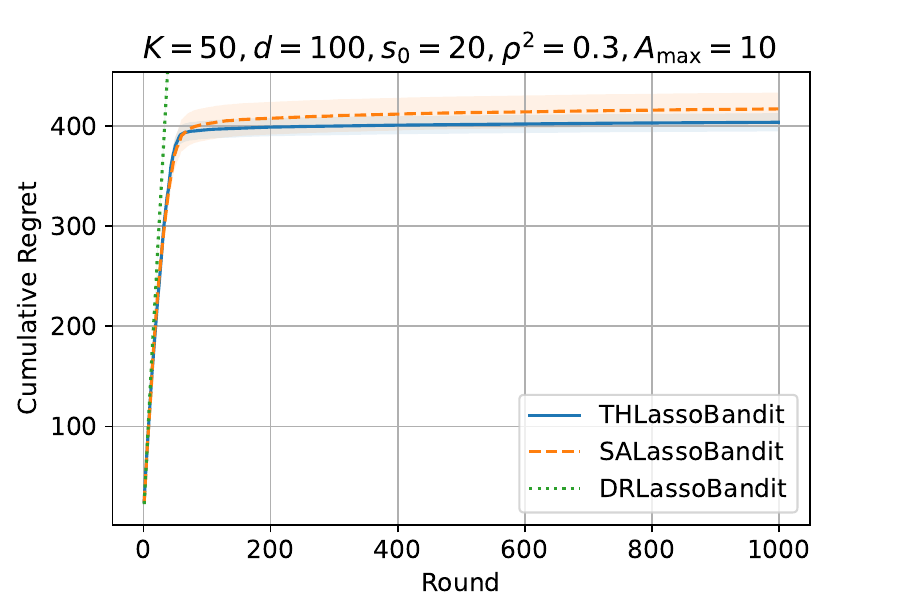}
    \end{minipage} \\
    \begin{minipage}[t]{0.24\columnwidth}
        \centering
        \includegraphics[width=1.1\textwidth]{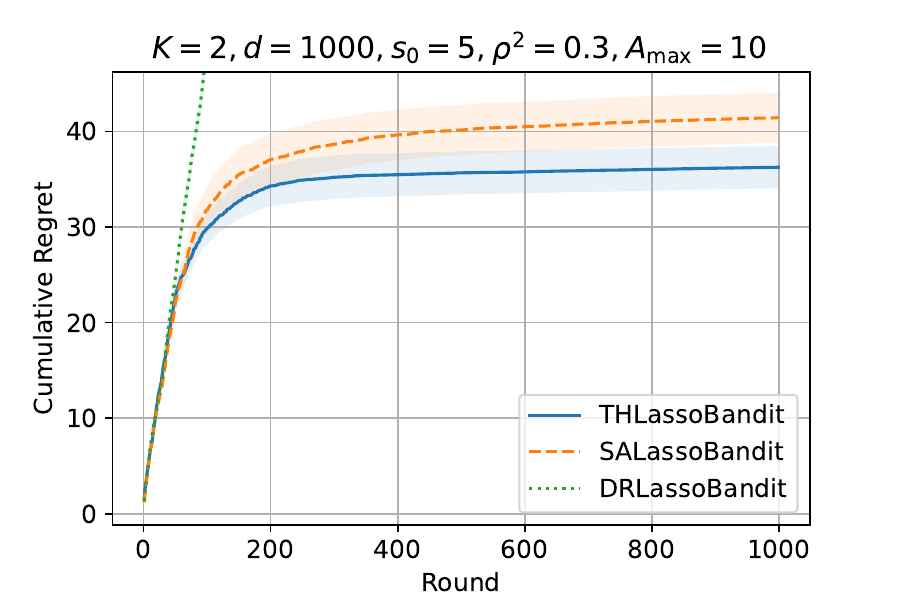}
    \end{minipage}
    \begin{minipage}[t]{0.24\columnwidth}
        \centering
        \includegraphics[width=1.1\textwidth]{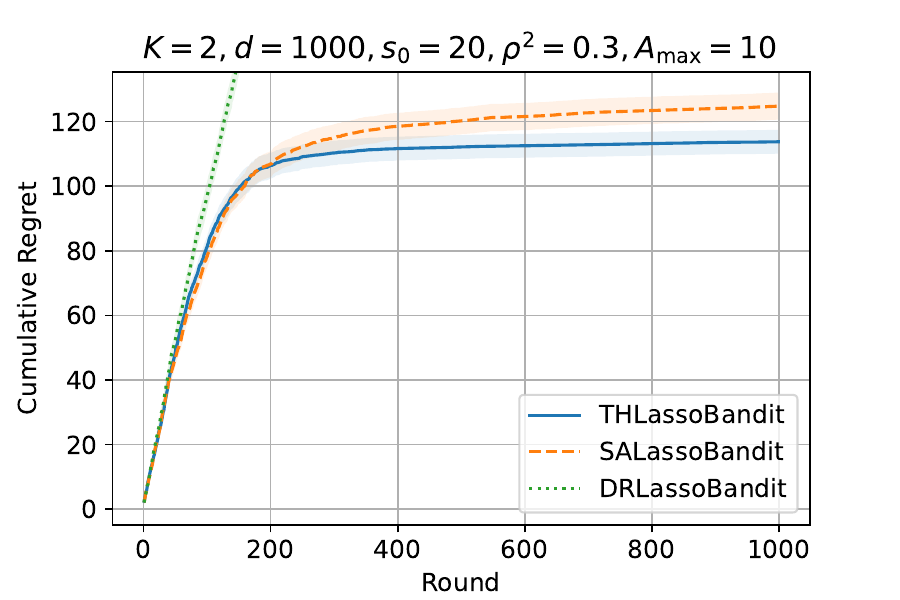}
    \end{minipage}
    \begin{minipage}[t]{0.24\columnwidth}
        \centering
        \includegraphics[width=1.1\textwidth]{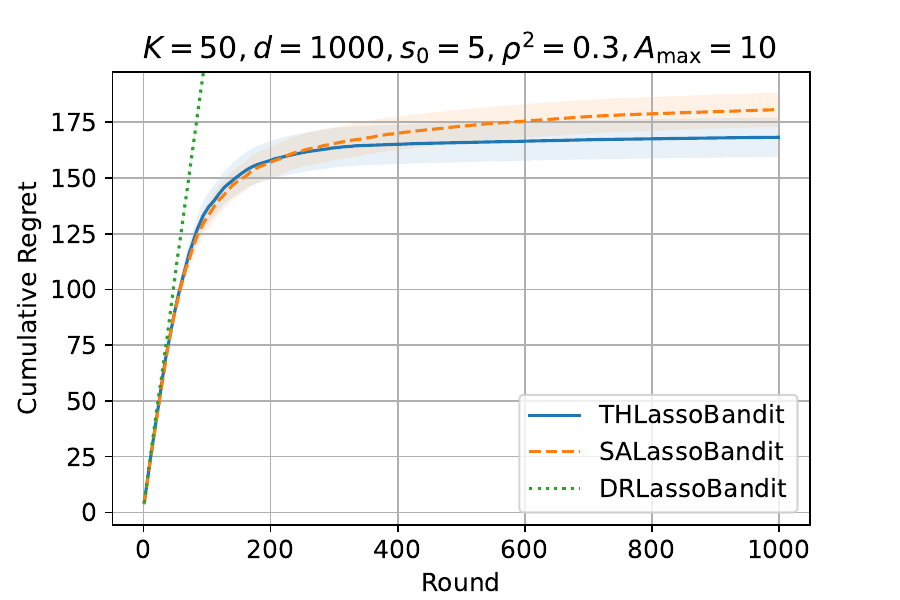}
    \end{minipage}
    \begin{minipage}[t]{0.24\columnwidth}
        \centering
        \includegraphics[width=1.1\textwidth]{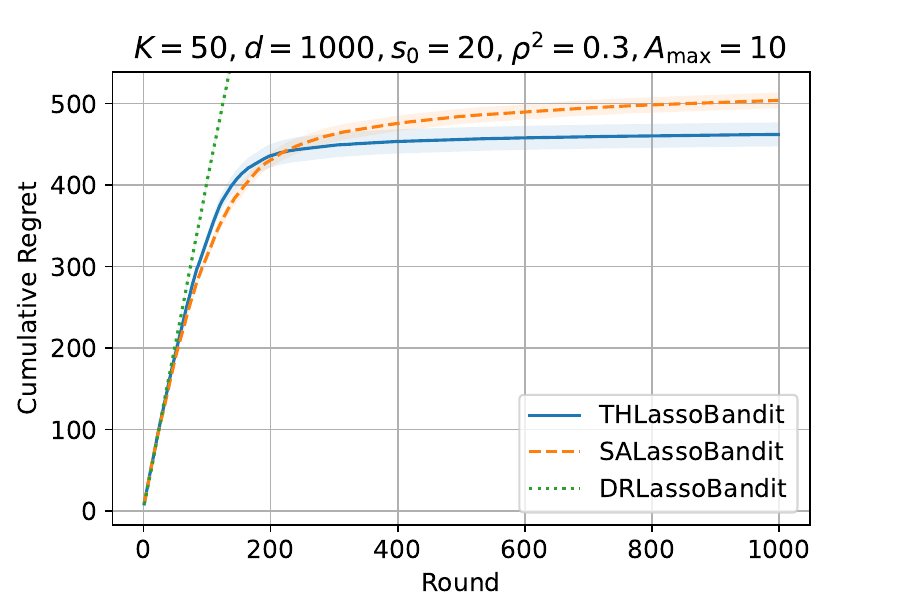}
    \end{minipage} \\
    \begin{minipage}[t]{0.24\columnwidth}
        \centering
        \includegraphics[width=1.1\textwidth]{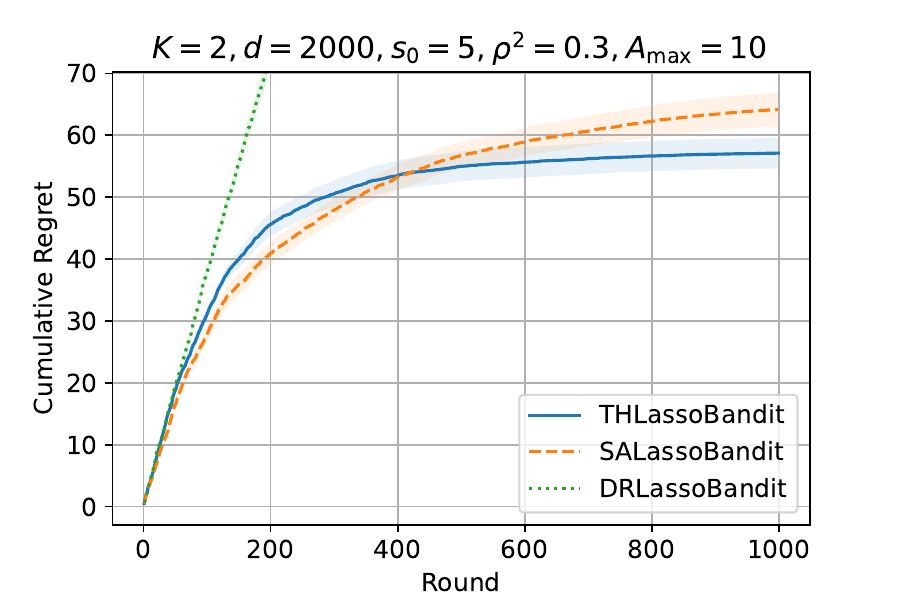}
    \end{minipage}
    \begin{minipage}[t]{0.24\columnwidth}
        \centering
        \includegraphics[width=1.1\textwidth]{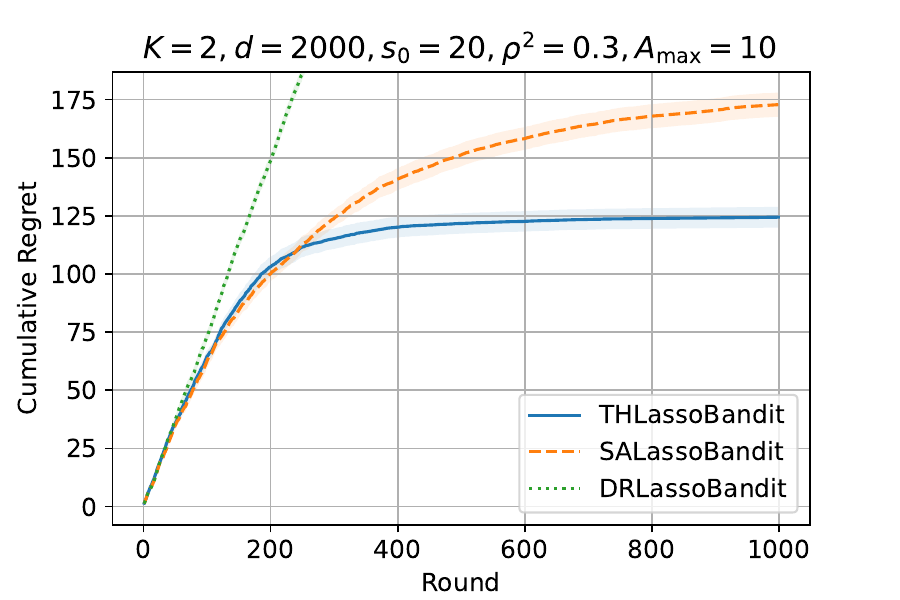}
    \end{minipage}
    \begin{minipage}[t]{0.24\columnwidth}
        \centering
        \includegraphics[width=1.1\textwidth]{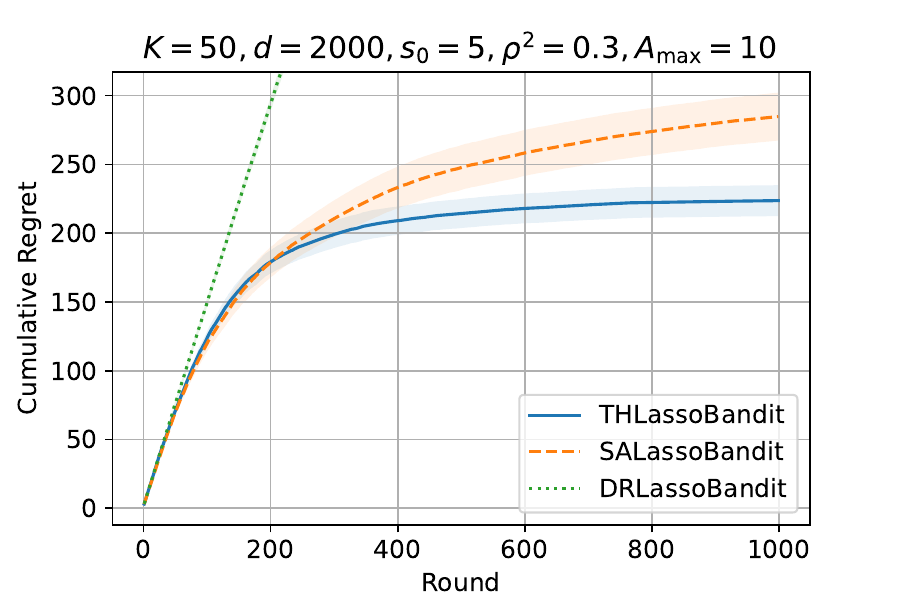}
    \end{minipage}
    \begin{minipage}[t]{0.24\columnwidth}
        \centering
        \includegraphics[width=1.1\textwidth]{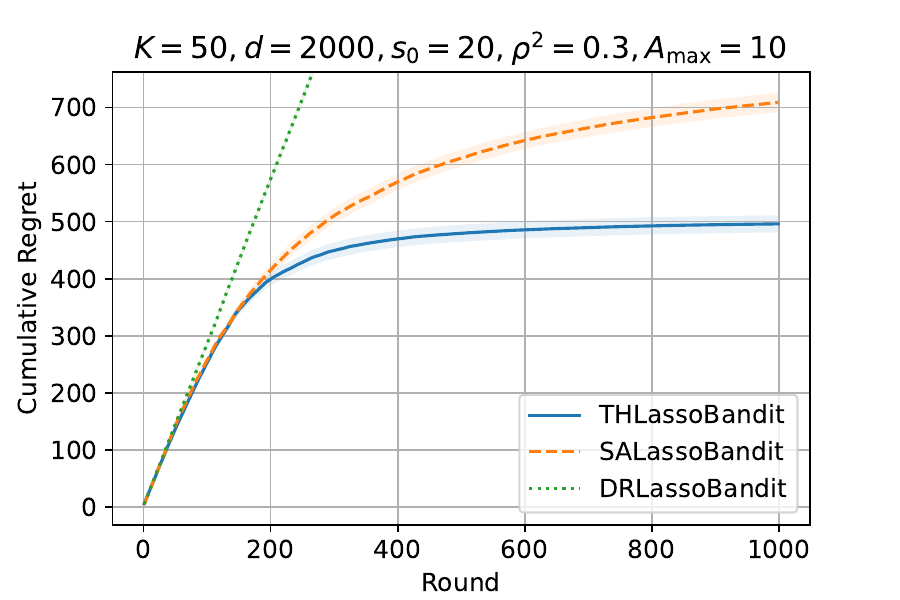}
    \end{minipage} \\
    \begin{minipage}[t]{0.24\columnwidth}
        \centering
        \includegraphics[width=1.1\textwidth]{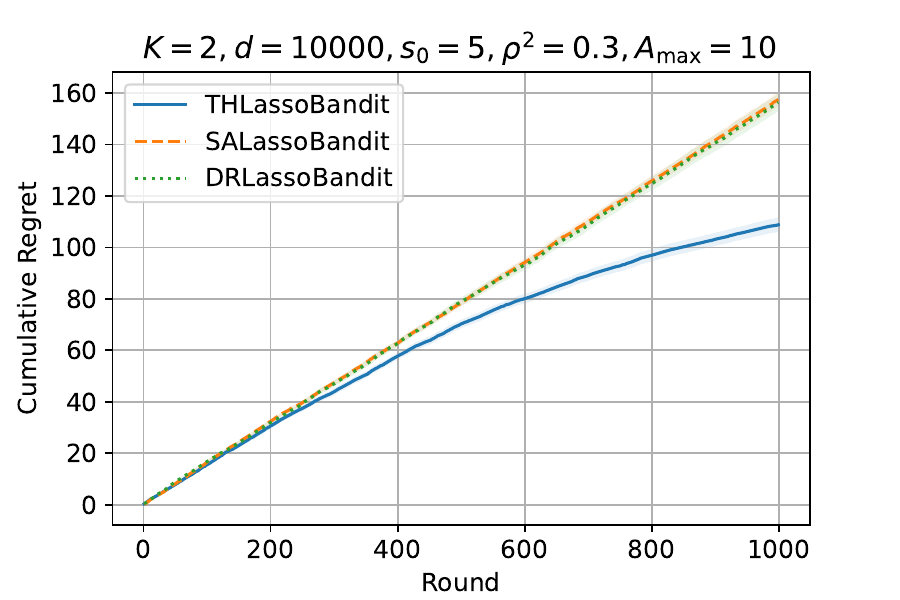}
    \end{minipage}
    \begin{minipage}[t]{0.24\columnwidth}
        \centering
        \includegraphics[width=1.1\textwidth]{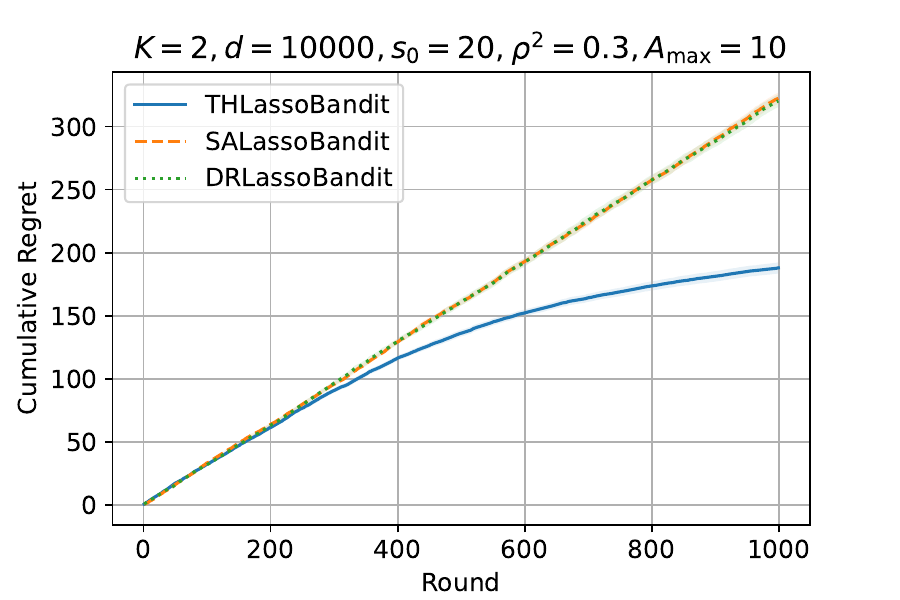}
    \end{minipage}
    \begin{minipage}[t]{0.24\columnwidth}
        \centering
        \includegraphics[width=1.1\textwidth]{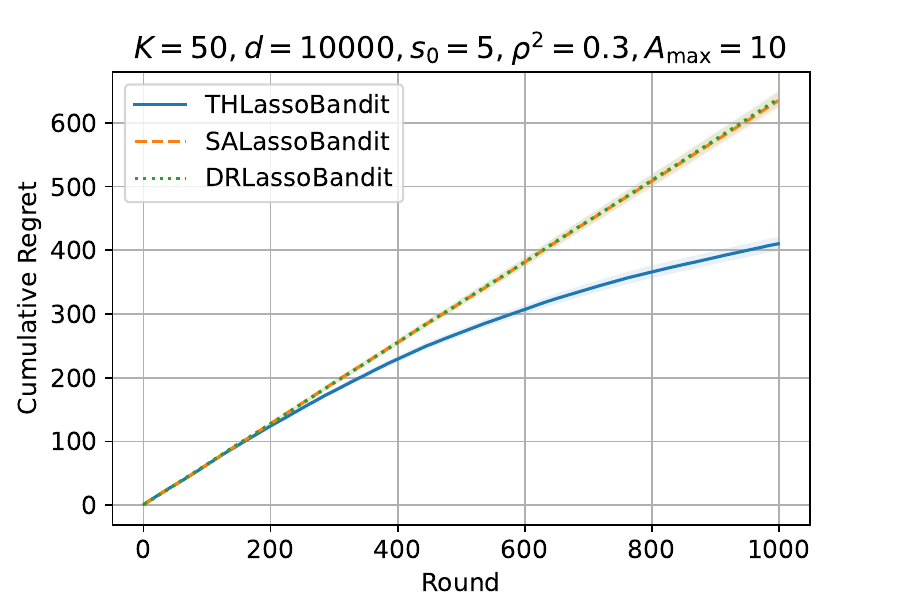}
    \end{minipage}
    \begin{minipage}[t]{0.24\columnwidth}
        \centering
        \includegraphics[width=1.1\textwidth]{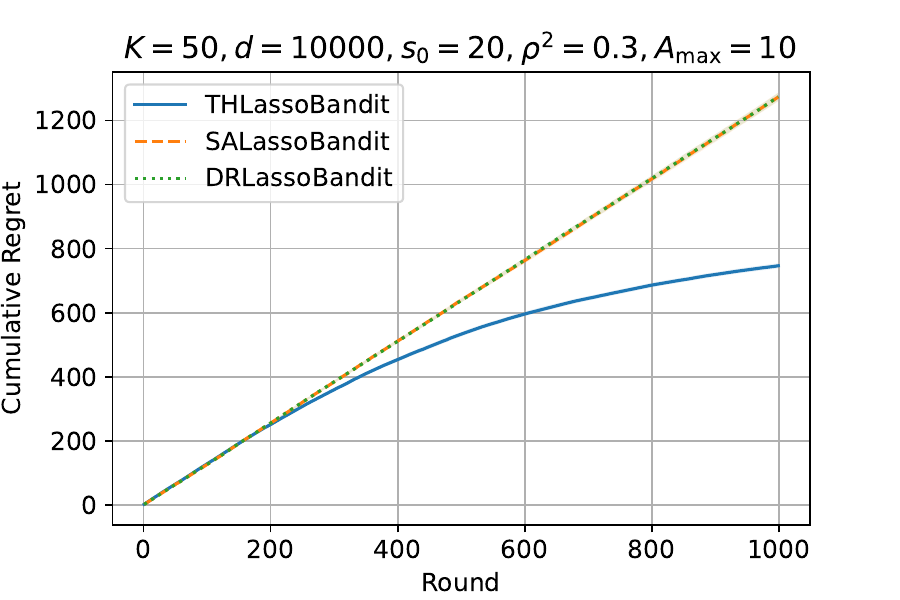}
    \end{minipage}
    \caption{\small Cumulative regret of the three algorithms with $\rho^2=0.3$, $A_{\max}=10$, $K\in\{2, 50\}$, $d\in \{100, 1000, 2000, 10000\}$, and $s_0\in \{5, 20\}$.
    The shaded area represents the standard errors.}
    \label{fig:synthetic_regrets_rho0.3}
\end{figure}

\begin{figure}[!ht]
    \centering
    \begin{minipage}[t]{0.24\columnwidth}
        \centering
        \includegraphics[width=1.1\textwidth]{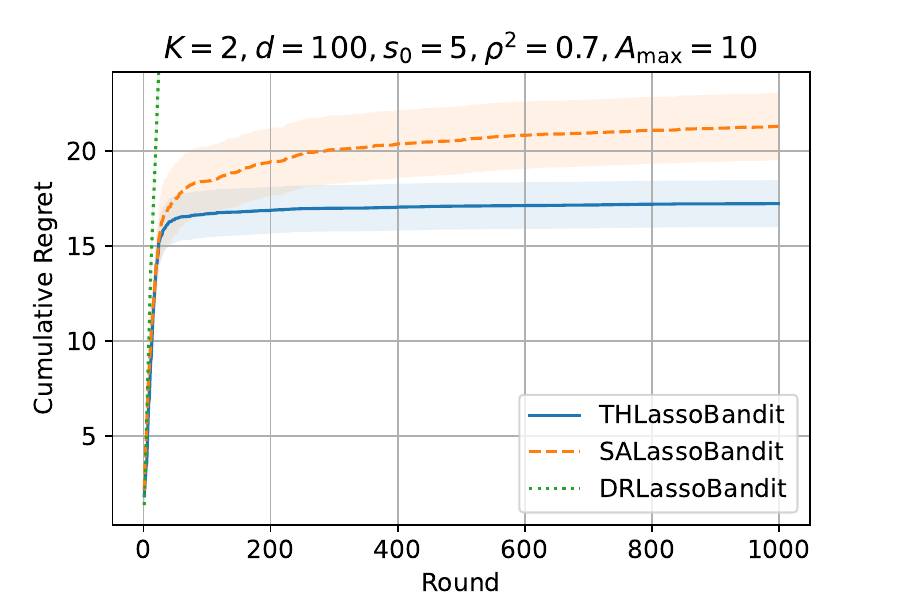}
    \end{minipage}
    \begin{minipage}[t]{0.24\columnwidth}
        \centering
        \includegraphics[width=1.1\textwidth]{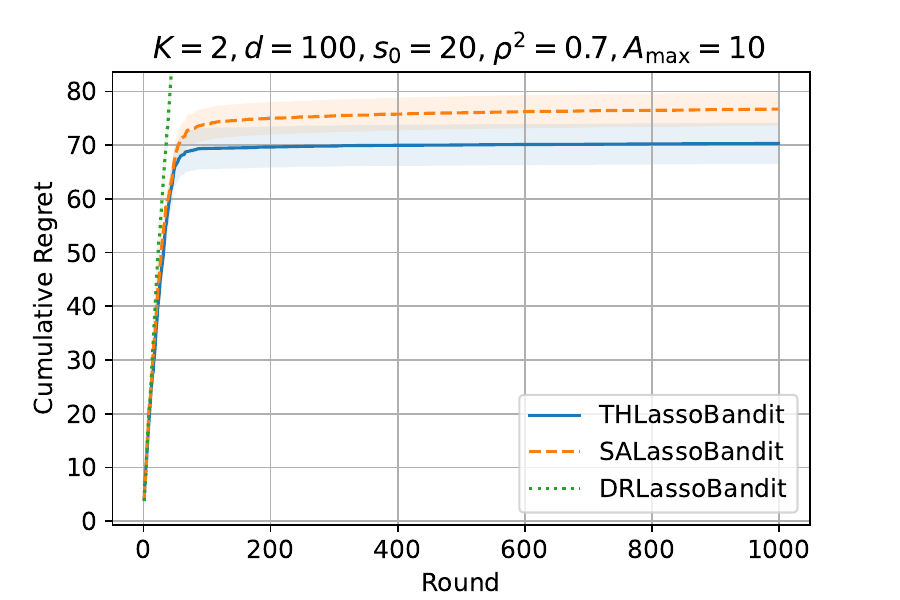}
    \end{minipage}
    \begin{minipage}[t]{0.24\columnwidth}
        \centering
        \includegraphics[width=1.1\textwidth]{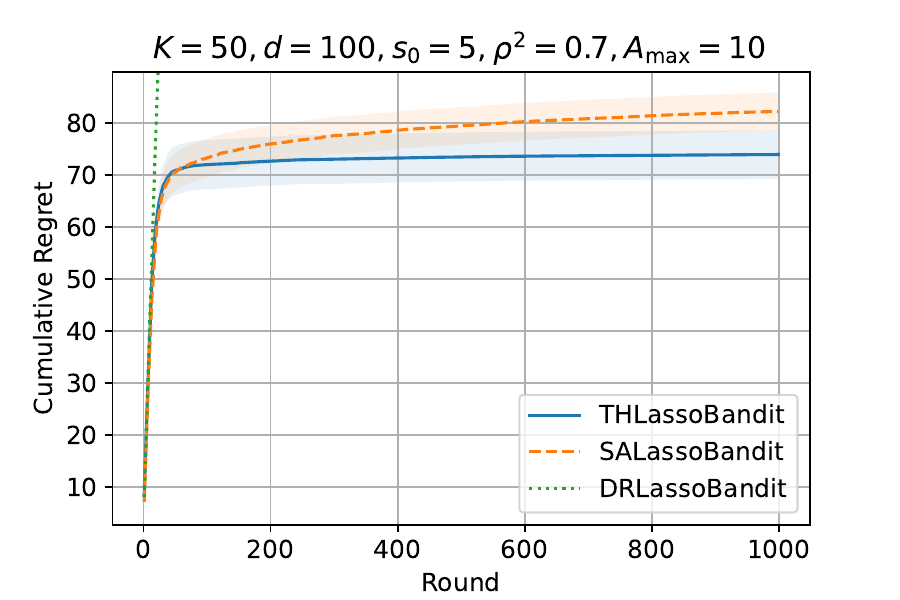}
    \end{minipage}
    \begin{minipage}[t]{0.24\columnwidth}
        \centering
        \includegraphics[width=1.1\textwidth]{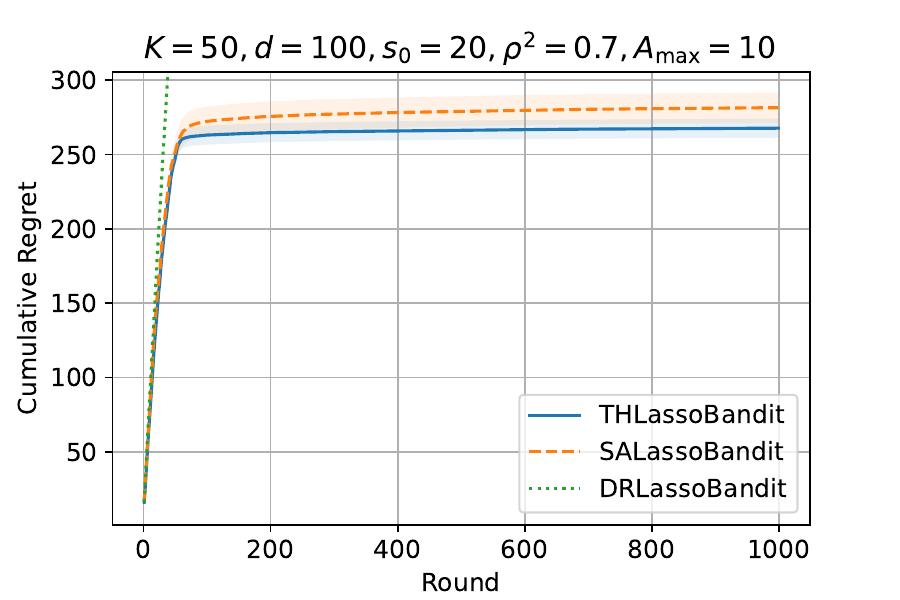}
    \end{minipage} \\
    \begin{minipage}[t]{0.24\columnwidth}
        \centering
        \includegraphics[width=1.1\textwidth]{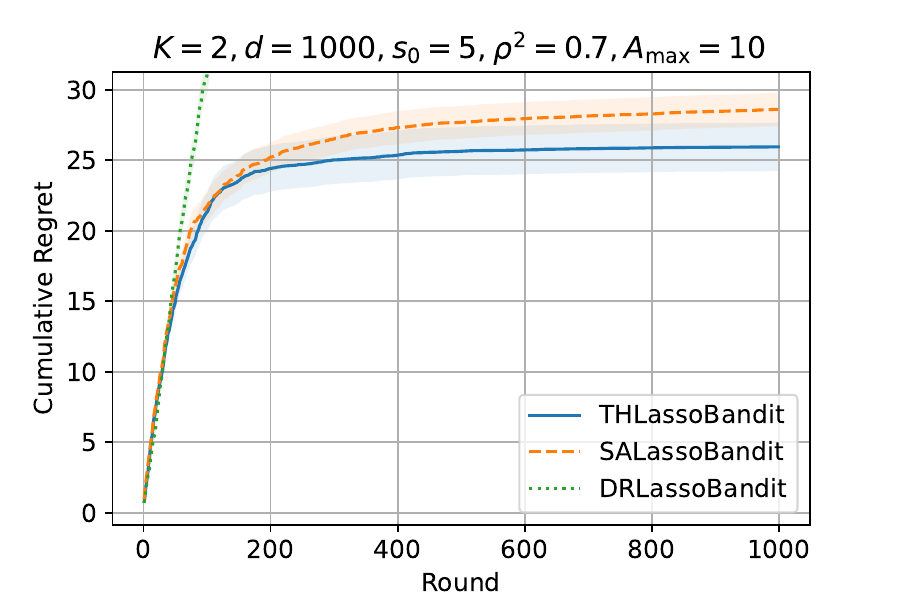}
    \end{minipage}
    \begin{minipage}[t]{0.24\columnwidth}
        \centering
        \includegraphics[width=1.1\textwidth]{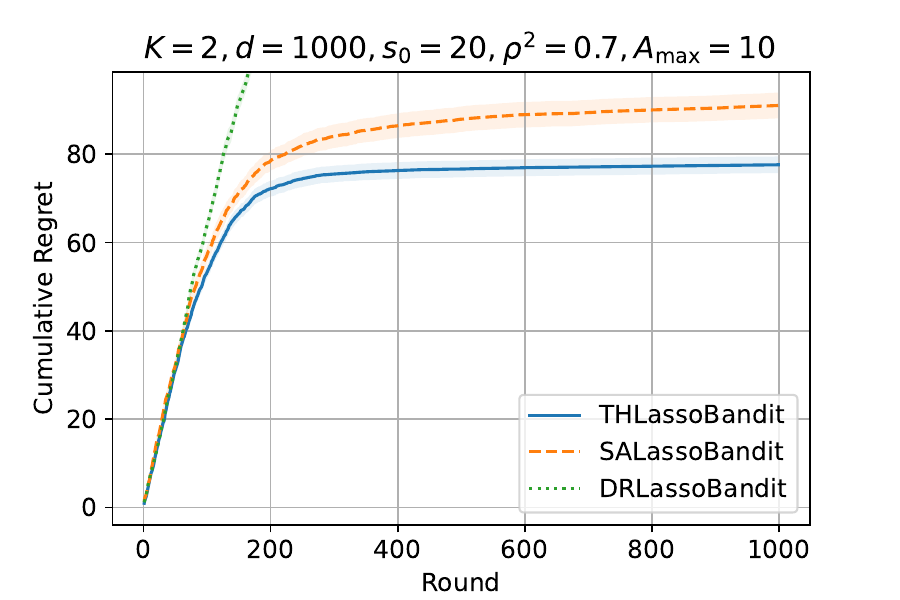}
    \end{minipage}
    \begin{minipage}[t]{0.24\columnwidth}
        \centering
        \includegraphics[width=1.1\textwidth]{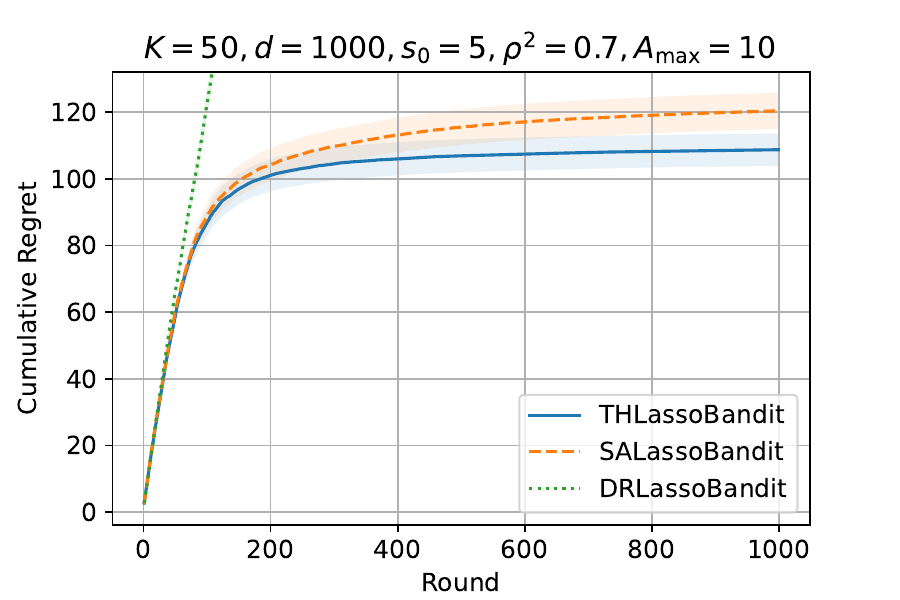}
    \end{minipage}
    \begin{minipage}[t]{0.24\columnwidth}
        \centering
        \includegraphics[width=1.1\textwidth]{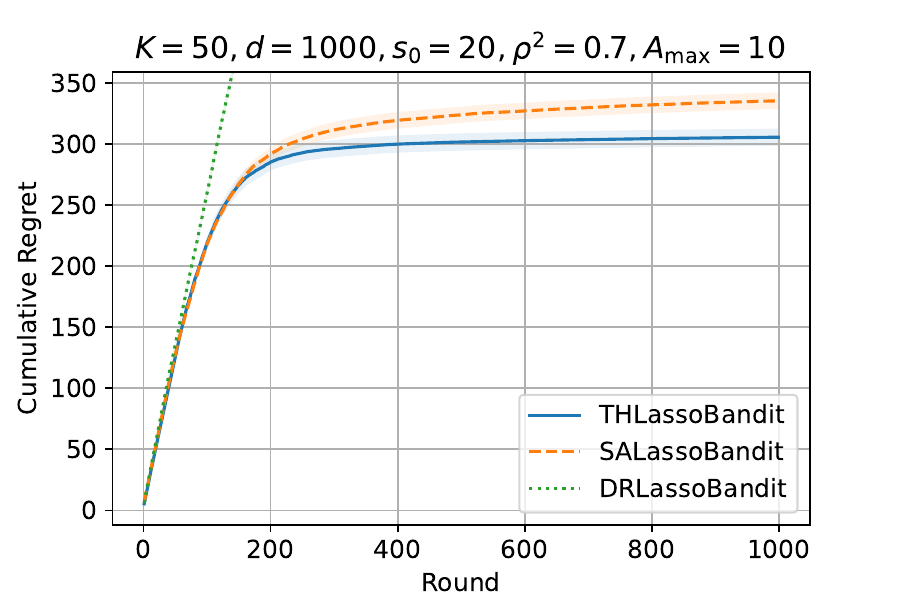}
    \end{minipage} \\
    \begin{minipage}[t]{0.24\columnwidth}
        \centering
        \includegraphics[width=1.1\textwidth]{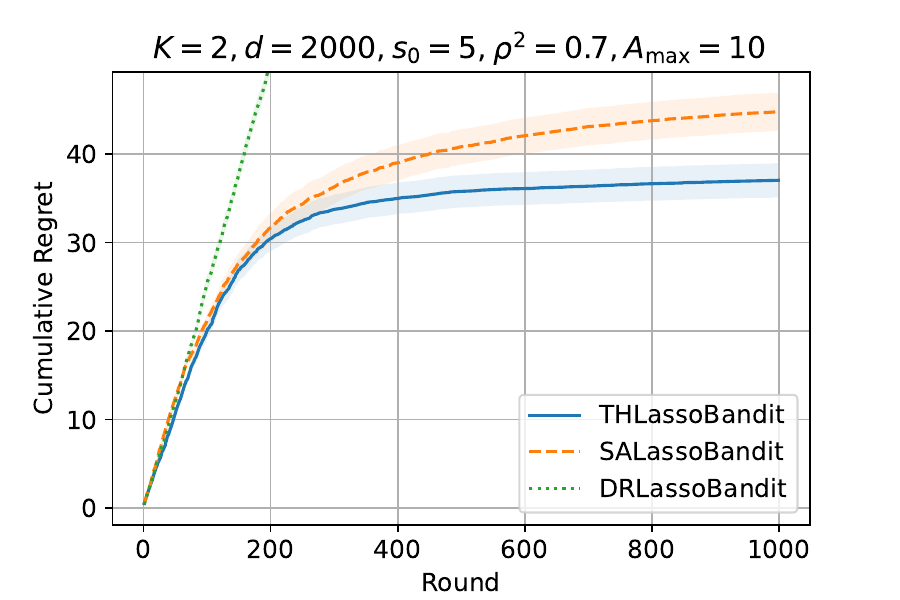}
    \end{minipage}
    \begin{minipage}[t]{0.24\columnwidth}
        \centering
        \includegraphics[width=1.1\textwidth]{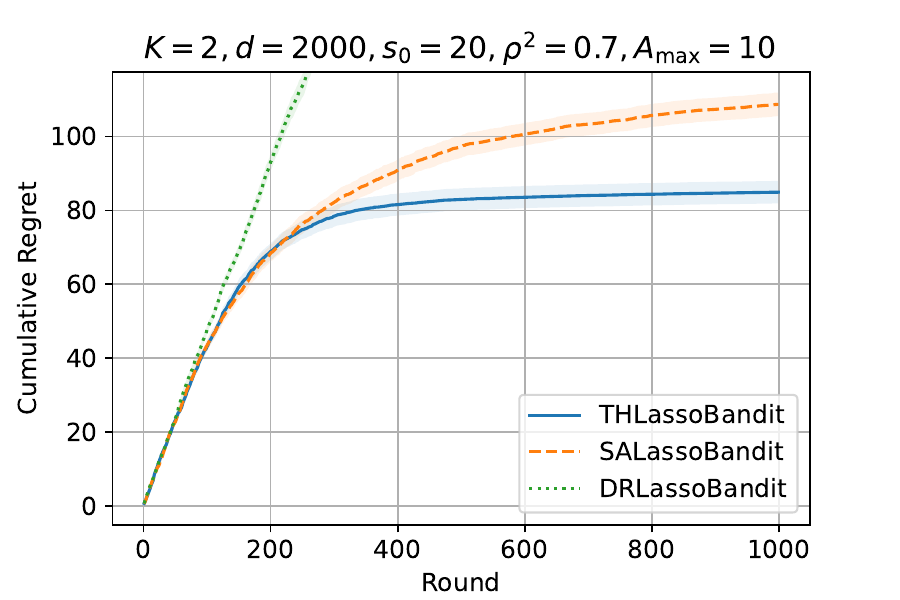}
    \end{minipage}
    \begin{minipage}[t]{0.24\columnwidth}
        \centering
        \includegraphics[width=1.1\textwidth]{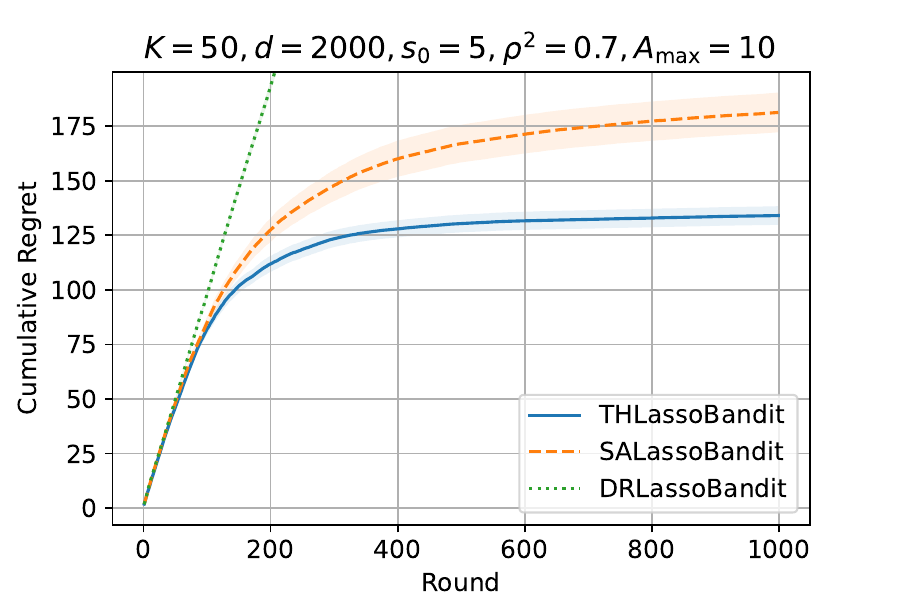}
    \end{minipage}
    \begin{minipage}[t]{0.24\columnwidth}
        \centering
        \includegraphics[width=1.1\textwidth]{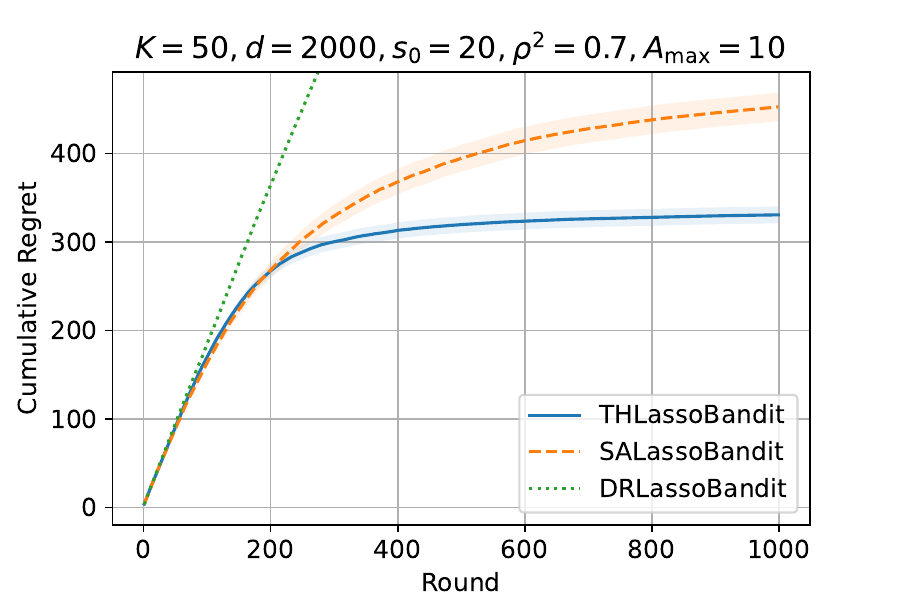}
    \end{minipage} \\
    \begin{minipage}[t]{0.24\columnwidth}
        \centering
        \includegraphics[width=1.1\textwidth]{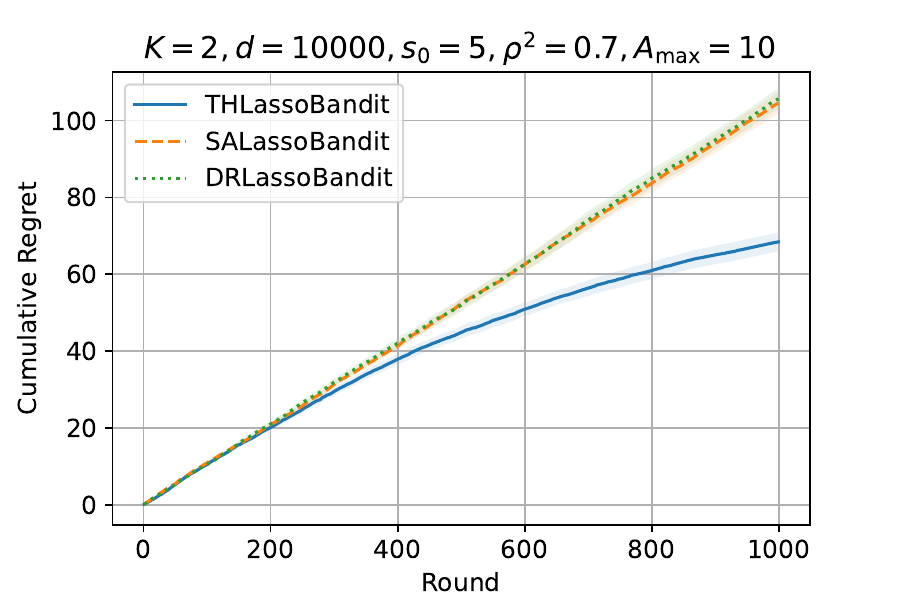}
    \end{minipage}
    \begin{minipage}[t]{0.24\columnwidth}
        \centering
        \includegraphics[width=1.1\textwidth]{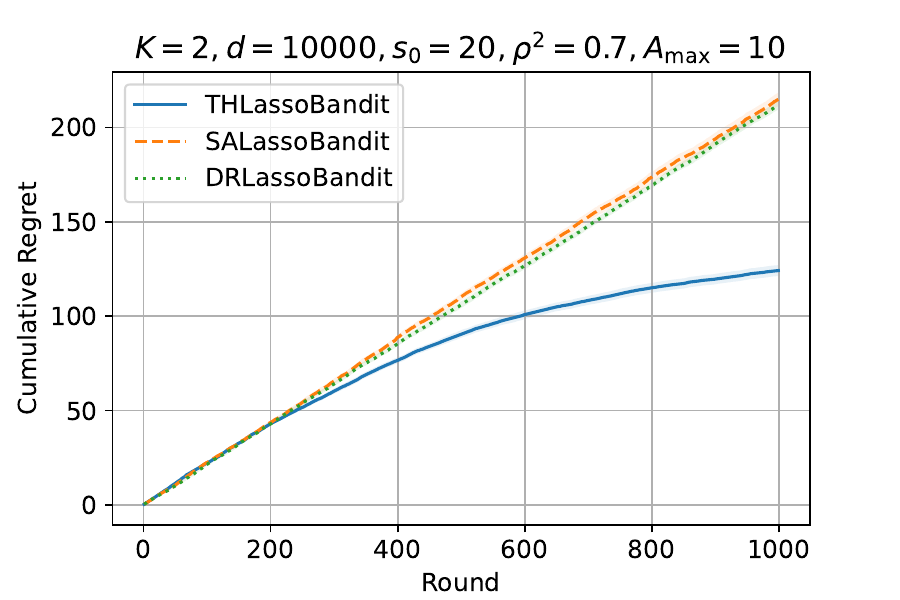}
    \end{minipage}
    \begin{minipage}[t]{0.24\columnwidth}
        \centering
        \includegraphics[width=1.1\textwidth]{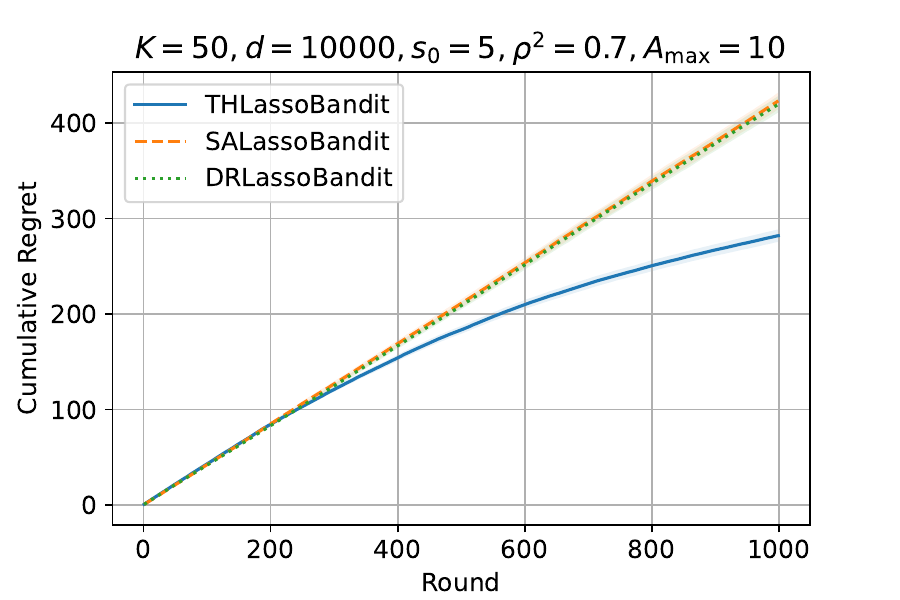}
    \end{minipage}
    \begin{minipage}[t]{0.24\columnwidth}
        \centering
        \includegraphics[width=1.1\textwidth]{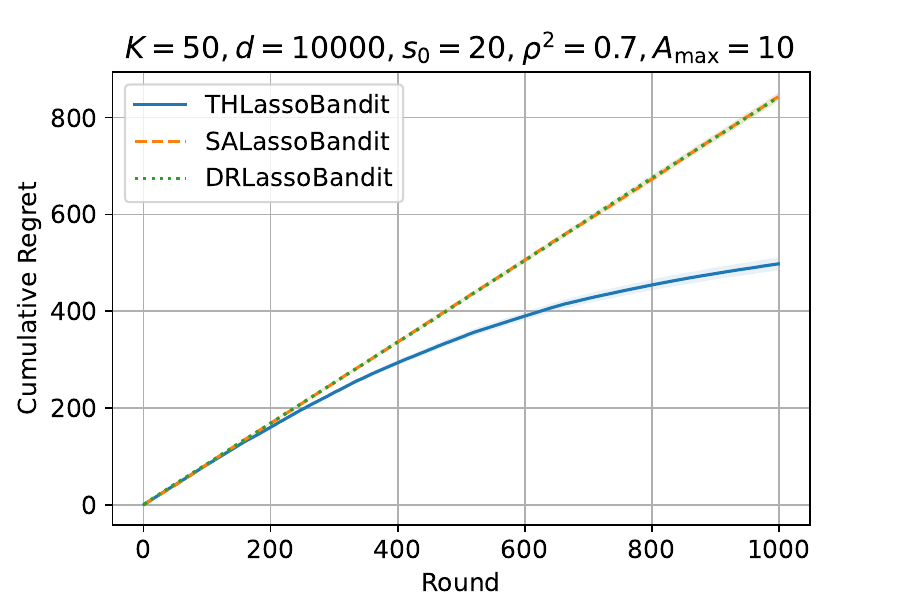}
    \end{minipage}
    \caption{\small Cumulative regret of the three algorithms with $\rho^2=0.7$, $A_{\max}=10$, $K\in\{2, 50\}$, $d\in \{100, 1000, 2000, 10000\}$, and $s_0\in \{5, 20\}$.
    The shaded area represents the standard errors.}
    \label{fig:synthetic_regrets_rho0.7}
\end{figure}

Figures \ref{fig:synthetic_regrets_rho0.0}-\ref{fig:synthetic_regrets_rho0.7} show the numerical results with correlation levels between two arms $\rho^2\in \{0.0, 0.3, 0.7\}$ and dimension $d\in \{100, 1000, 2000, 10000\}$, respectively.
We find that TH Lasso bandit exhibits lower regret than SA Lasso bandit and DR Lasso bandit in all scenarios. In particular, the difference between TH Lasso and SA Lasso becomes more apparent as the dimension $d$ increases, just as the theorem shows.

\noindent {\small \bf Case 1: $\rho^2=0.0$}

\clearpage
\noindent {\small \bf Case 2: $\rho^2=0.3$}

\clearpage
\noindent {\small \bf Case 3: $\rho^2=0.7$}

\clearpage
\subsection{Additional Results with Various $A_{\max}$ for $K$-Armed Bandits}
\label{subsec:additional_Amax}
\begin{figure}[h!]
\centering
\includegraphics[width=0.5\columnwidth]{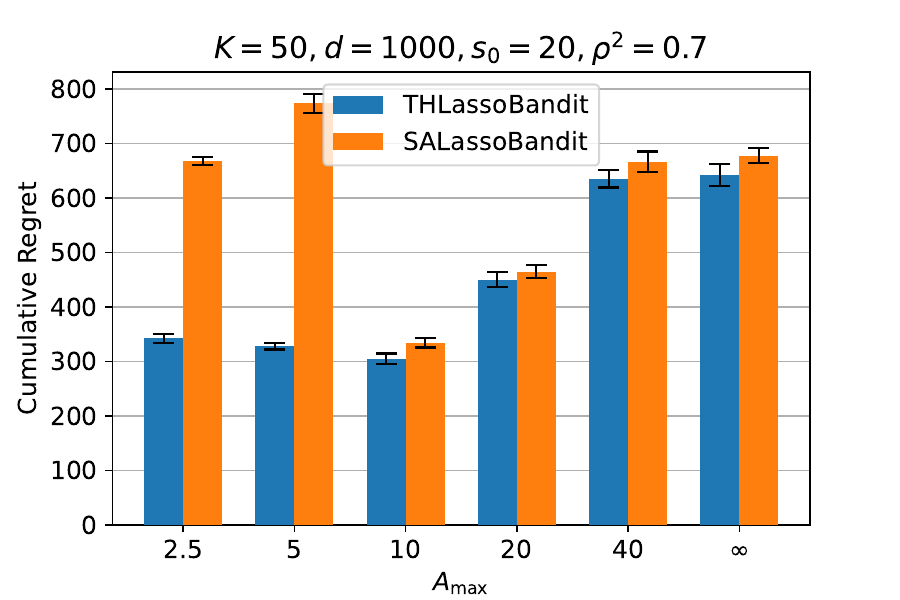}
\caption{Cumulative regret at round $t=1000$ of TH Lasso bandit and SA Lasso bandit with $\rho^2=0.7$, $K=50$, $d=1000$, $s_0=20$, and varying $A_{\max}\in \{2.5, 5, 10, 20, 40, \infty\}$.
The error bars represent the standard errors.}
\label{fig:experiment3_regrets}
\end{figure}

We also present the experimental results with varying $A_{\max}\in \{2.5, 5, 10, 20, 40, \infty\}$ and a different parameter setting.
We set $K=50, d=1000$, and $s_0=20$.
Figure~\ref{fig:experiment3_regrets} shows the average cumulative regret at $t=1000$ of TH Lasso bandit and SA Lasso bandit for each $A_{\max}$.
We observe that TH Lasso bandit outperforms SA Lasso bandit for all $A_{\max}$.

\newpage
\subsection{Additional Results with Non-Gaussian Distributions}
\label{sec:exp_non_gaussian}
\begin{figure}[!ht]
    \centering
    \begin{minipage}[t]{0.32\columnwidth}
        \centering
        \includegraphics[width=1.1\textwidth]{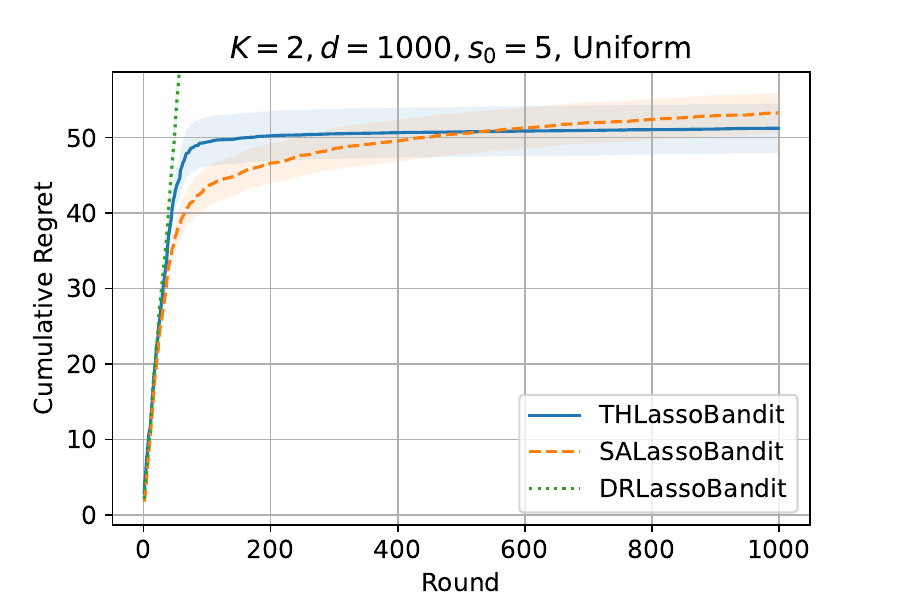}
    \end{minipage}
    \begin{minipage}[t]{0.32\columnwidth}
        \centering
        \includegraphics[width=1.1\textwidth]{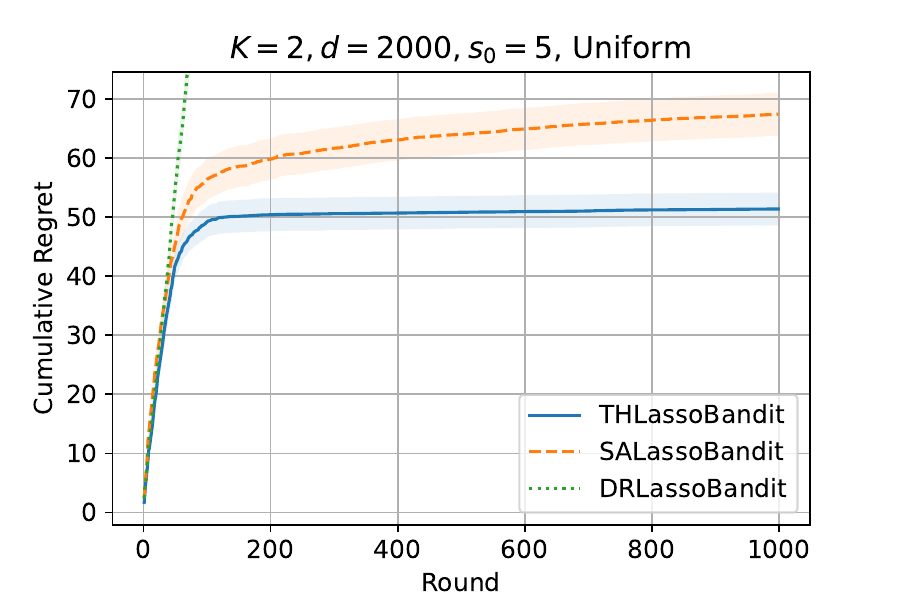}
    \end{minipage}
    \begin{minipage}[t]{0.32\columnwidth}
        \centering
        \includegraphics[width=1.1\textwidth]{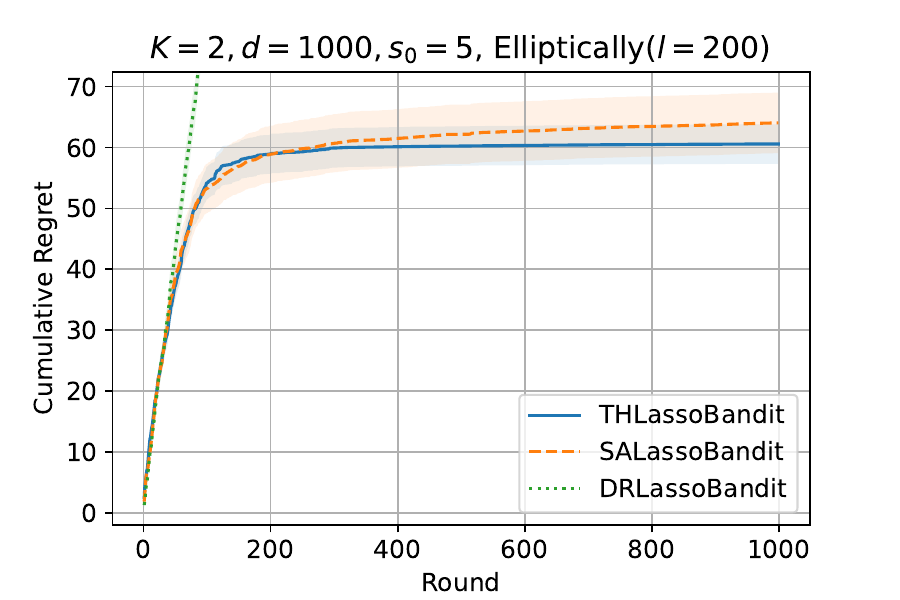}
    \end{minipage} \\
    \begin{minipage}[t]{0.32\columnwidth}
        \centering
        \includegraphics[width=1.1\textwidth]{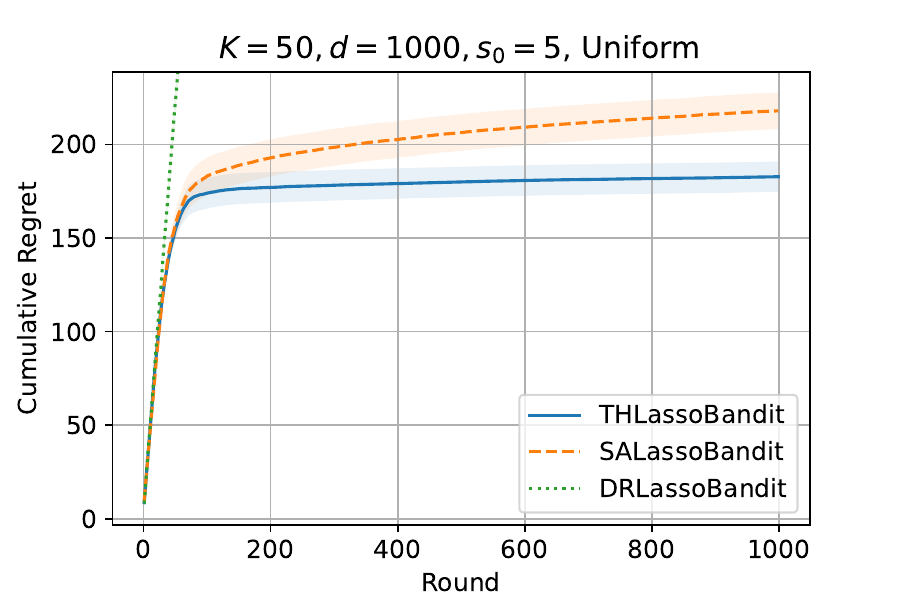}
    \end{minipage}
    \begin{minipage}[t]{0.32\columnwidth}
        \centering
        \includegraphics[width=1.1\textwidth]{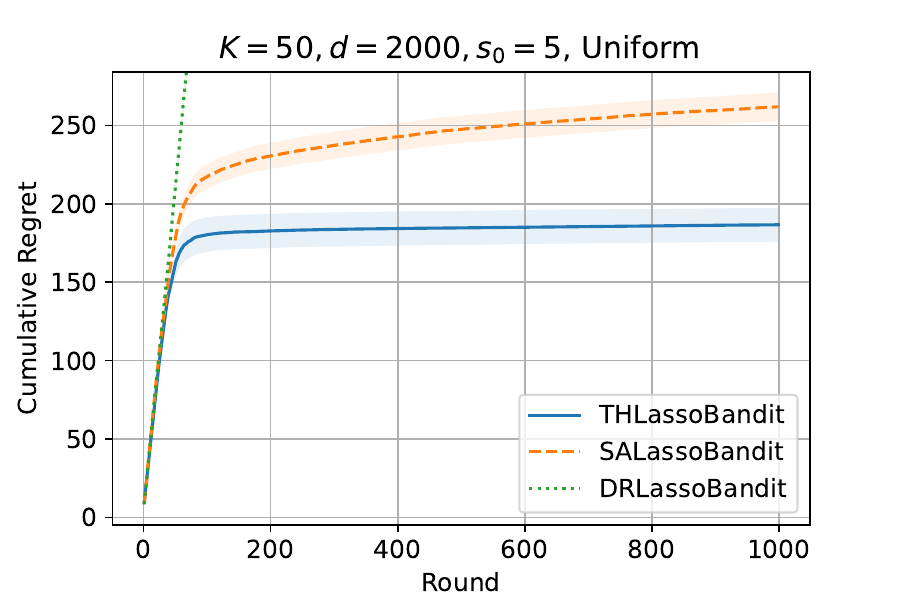}
    \end{minipage}
    \begin{minipage}[t]{0.32\columnwidth}
        \centering
        \includegraphics[width=1.1\textwidth]{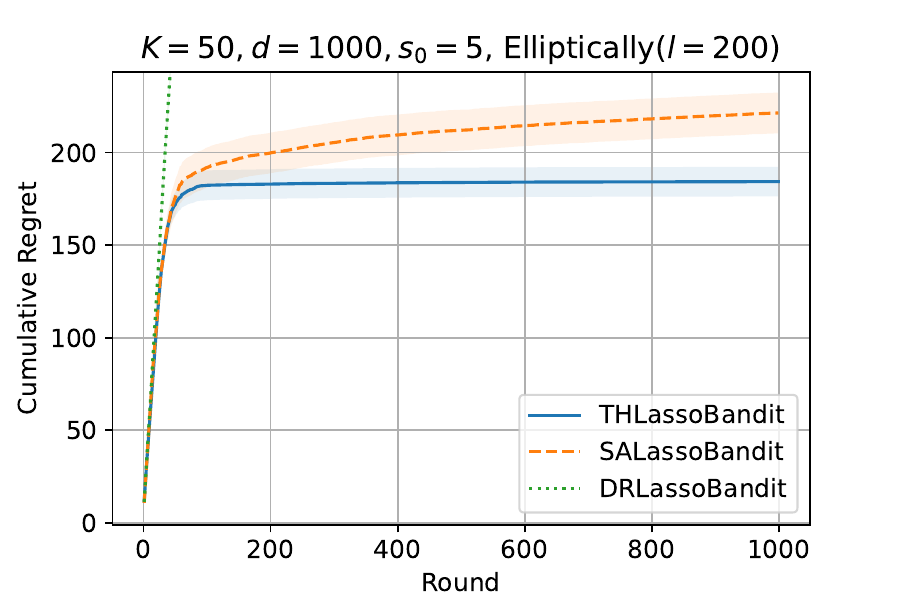}
    \end{minipage}
    \caption{Cumulative regret of the three algorithms with non-Gaussian distributions.
    The figures in left and center columns show the experimental results with the uniform distribution.
    The figures in right column shows the experimental results with the elliptical distribution.
    The shaded area represents the standard errors.}
    \label{fig:synthetic_regrets_non-gaussian}
\end{figure}

Figure \ref{fig:synthetic_regrets_non-gaussian} shows the numerical results with the uniform distribution and non-Gaussian elliptical distributions.
For experiments with the uniform distribution, in each round $t$, we sample each feature vector $A_{t,k}$ independently from a $d$-dimensional hypercube $[-1, 1]^d$.
For experiments with the elliptical distribution, we construct each feature vector $A_{t,k}$ by the following equation:
\begin{align*}
    A_{t,k} = \mu + RAU^{(l)}
\end{align*}
where $\mu\in \mathbb{R}^{d}$ is a mean vector, $U^{(l)} \in \mathbb{R}^{l}$ is uniformly distributed on the unit sphere in $\mathbb{R}^{(l)}$, $R\in \mathbb{R}$ is a random variable independent of $U^{(l)}$, and $A$ is a $d\times l$-dimensional matrix with rank $l$.
We sample $R$ from Gaussian distribution $\mathcal{N}(0, 1)$, and sample each element of $A$ in an i.i.d manner using the uniform distribution on $[-1, 1]$.
We set $\mu=\mathbf{0}_{d}$ and set $l=200$.
As in the previous experiments, we generate each non-zero components of $\theta$ in an i.i.d manner using the uniform distribution on $[1, 2]$.
The noise process is Gaussian, i.i.d. over rounds: $\epsilon_t \sim \mathcal{N}(0, 1)$.
We find that TH Lasso bandit exhibits lower regret than SA Lasso bandit and DR Lasso bandit in experiments with both distributions.

\clearpage

\subsection{Additional Results in Hard Instances}\label{subsec:numerical_hard_instances}
\begin{figure}[!ht]
    \centering
    \begin{minipage}[t]{0.32\columnwidth}
        \centering
        \includegraphics[width=1.1\textwidth]{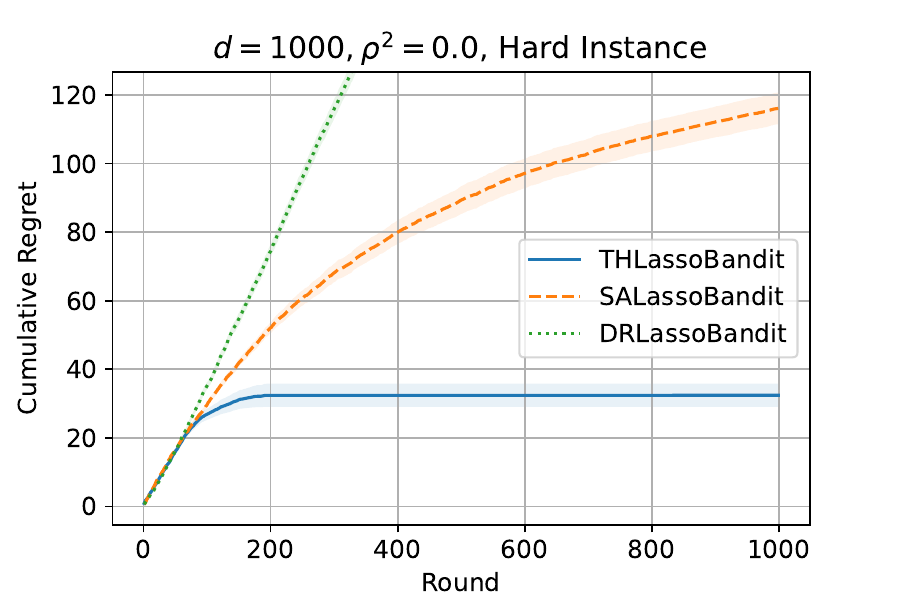}
    \end{minipage}
    \begin{minipage}[t]{0.32\columnwidth}
        \centering
        \includegraphics[width=1.1\textwidth]{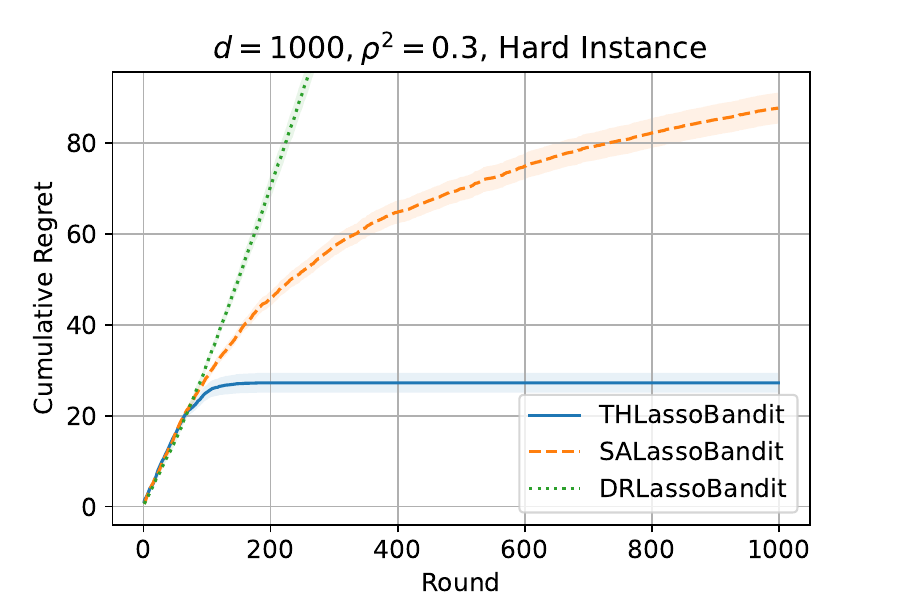}
    \end{minipage}
    \begin{minipage}[t]{0.32\columnwidth}
        \centering
        \includegraphics[width=1.1\textwidth]{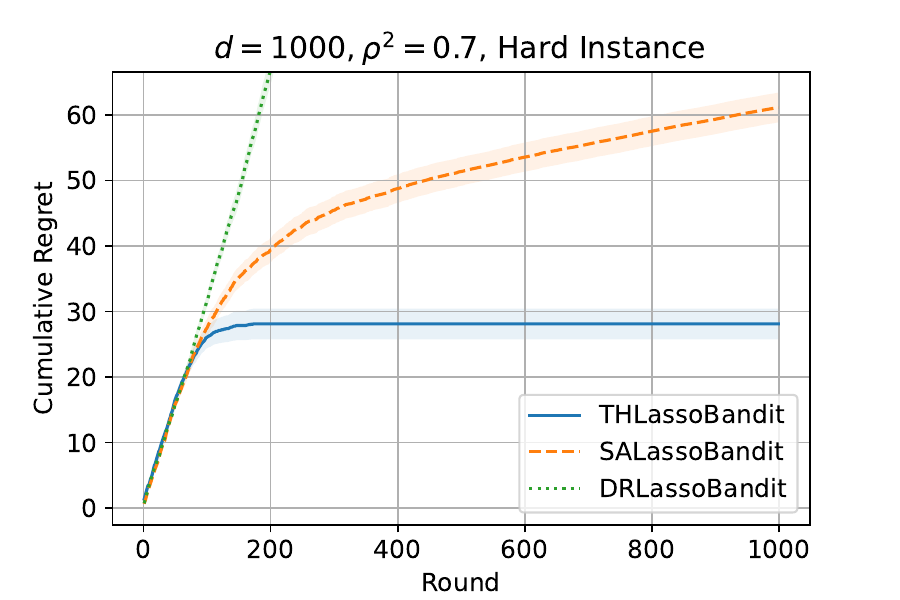}
    \end{minipage} \\
    \begin{minipage}[t]{0.32\columnwidth}
        \centering
        \includegraphics[width=1.1\textwidth]{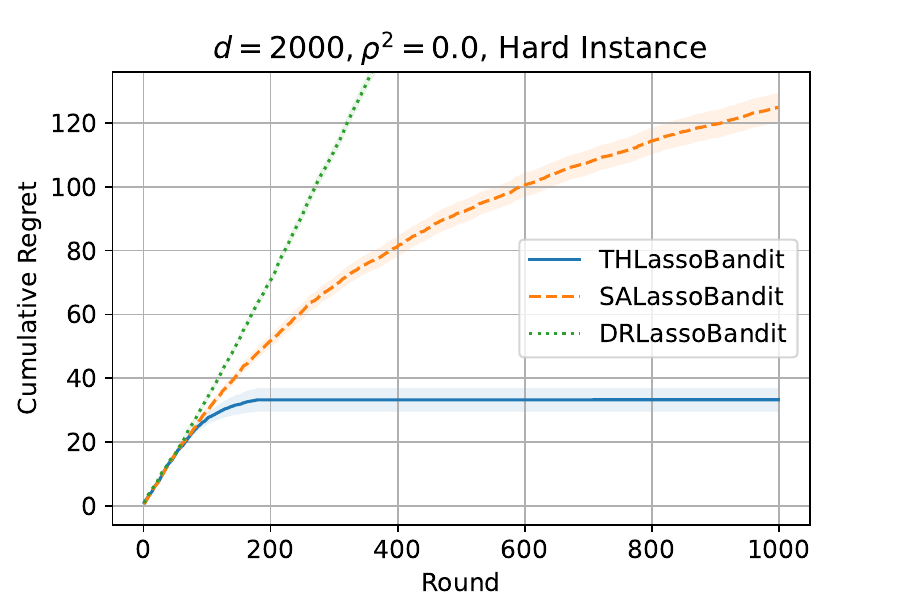}
    \end{minipage}
    \begin{minipage}[t]{0.32\columnwidth}
        \centering
        \includegraphics[width=1.1\textwidth]{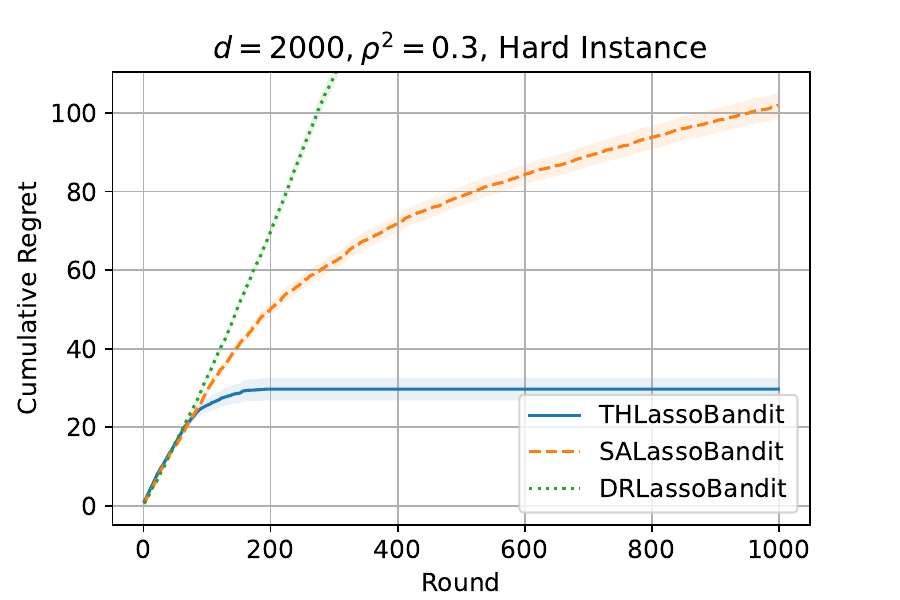}
    \end{minipage}
    \begin{minipage}[t]{0.32\columnwidth}
        \centering
        \includegraphics[width=1.1\textwidth]{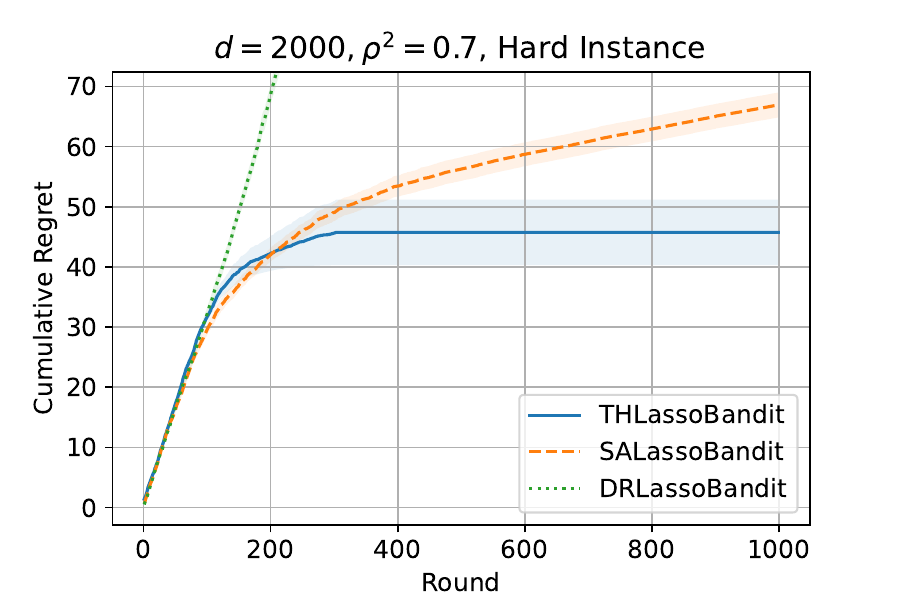}
    \end{minipage}
    \caption{Cumulative regret of the three algorithms in hard instances.
    The shaded area represents the standard errors.}
    \label{fig:synthetic_regrets_non_diversity}
\end{figure}

We further investigate the performance of TH Lasso bandit in hard instances where the feature vectors of the best arms do not cover the support of $\theta$ (situations where the covariate diversity condition \cite{bastani2021mostly} does not hold).
In this experiment, we set $K=3$ and $\theta = (1, 0.1, 1, 0, 0, \cdots, 0)^\top$ so that $S=\{1, 2, 3\}$.
We generate the arm set by generating feature vectors separately on the support $S$ and the non-support $S^c=[d]\setminus S$.
First, in each round $t$, we generate the feature vectors on $S$ as the following procedure: in each round $t$, the set of feature vectors $\mathcal{A}_t^S=\{A_{t,1}^S, A_{t,2}^S, A_{t,3}^S\}$ is set to $\mathcal{A}^1$ with probability $0.3$ and is set to $\mathcal{A}^2$ with probability $0.7$.
We set $\mathcal{A}^1=\{(1, 0, 0)^\top, (0, 1, 0)^\top, (0.9, 0.5, 0)^\top\}$ and $\mathcal{A}^1=\{(0, 1, 0)^\top, (0, 0, 1)^\top, (0.0, 0.5, 0.9)^\top\}$.
Second, for each component $i\in S^c$, we sample $((A_{t,1}^{S^c})_i, (A_{t,2}^{S^c})_i, (A_{t,3}^{S^c})_i)^\top \in \mathbb{R}^3$ from a Gaussian distribution $\mathcal{N}(\mathbf{0}_3, V)$ where $V_{j,j}=1$ for all $j\in[3]$ and $V_{j,k}=\rho^2$ for all $j\neq k\in [3]$.
We then define $A_{t,k}^{S^c}=((A_{t,k}^{S^c})_1, \cdots, (A_{t,k}^{S^c})_{|S^c|})^\top \in \mathbb{R}^{|S^c|}$.
Finally, by concatenating the feature vectors on $S$ and $S^c$, we construct the feature vector $A_{t,k} = ((A_{t,k}^S)^\top, (A_{t,k}^{S^c})^\top)^{\top}\in \mathbb{R}^d$.
Note that $(1, 0, 0)^\top\in \mathcal{A}^1$ and $(0, 0, 1)^\top\in \mathcal{A}^2$ are always included in the best arms, and they do not span $\mathbb{R}^3$.
Figure \ref{fig:synthetic_regrets_non_diversity} shows the numerical results with correlation levels between two arms $\rho^2\in \{0.0, 0.3, 0.7\}$ and dimension $d\in \{1000, 2000\}$.
We find that TH Lasso bandit exhibits lower regret than SA Lasso bandit and DR Lasso bandit in all scenarios.

\subsection{Additional Results with Real-world Dataset}
\label{subsec:real_data_experiments}
In this section, we empirically evaluate the TH Lasso bandit algorithm on a real-world dataset.
We compare its performance to those of the (uniformly) random policy and LinUCB \cite{li2010contextual} with the exploration parameter $\alpha=0.1$.

\begin{figure}[h!]
\centering
\includegraphics[width=0.5\columnwidth]{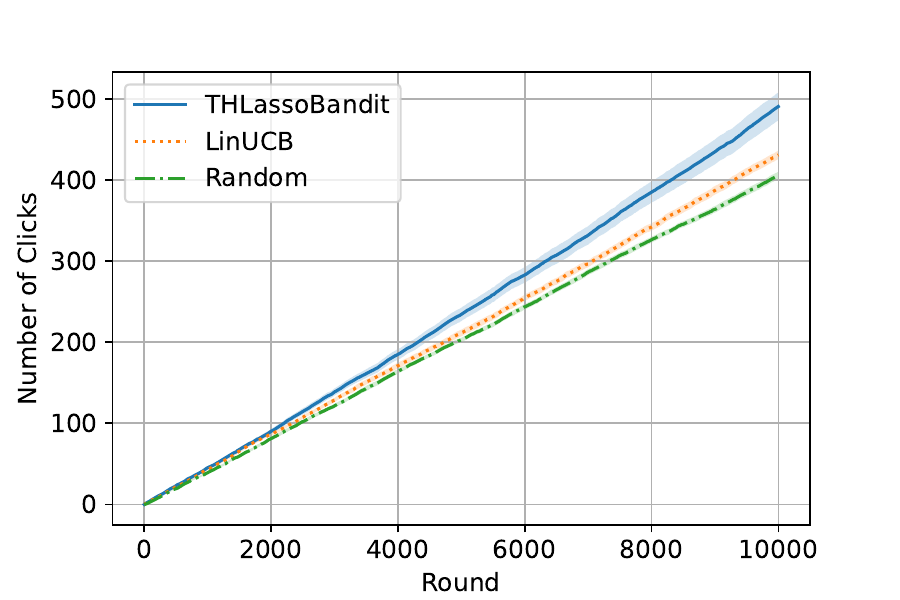}
\caption{Average number of clicks for the three algorithms on a real-world dataset.
The error bars represent the standard errors.}
\label{fig:real_data_experiments}
\end{figure}

We use the R6A dataset\footnote{\url{https://webscope.sandbox.yahoo.com}} that contains a part of the user view/click log for articles displayed on the Yahoo!'s Today Module.
Specifically, we use the dataset corresponding to May 1st, 2009.
We evaluate each algorithm using the replay method \cite{li2011unbiased} with $T=10,000$ valid events.
To evaluate the expected value and variance, we subsampled the data so that each event is used with probability $0.9$ for each instance.
For a fair comparison between LinUCB and TH Lasso bandit, we modify the TH Lasso bandit algorithm so that the unknown vector $\theta$ varies across arms.
To emulate a high-dimensional setting, we generate each context vector $A_{t,k}$ by concatenating the original $12$-dimensional feature vector with the random dummy vector sampled from the uniform distribution on $[0,1]^{88}$.
We present the average number of clicks from users for $20$ instances in Figure \ref{fig:real_data_experiments}. 
We observe that TH Lasso bandit outperforms the random policy and LinUCB.

\end{document}